\documentclass[10pt,twocolumn,letterpaper]{article}

\usepackage{cvpr}              % To produce the CAMERA-READY version
\usepackage[dvipsnames]{xcolor}

\definecolor{cvprblue}{rgb}{0.21,0.49,0.74}
\usepackage[pagebackref,breaklinks,colorlinks,citecolor=cvprblue]{hyperref}
\usepackage{pifont}% 
\usepackage{multirow} 

\usepackage{caption} 
\usepackage{etoolbox}
\AtEndEnvironment{table}{\vspace{-5mm}}
\AtEndEnvironment{figure}{\vspace{-2mm}}
\usepackage{chngcntr}
\usepackage{cvpr}  

\usepackage[utf8]{inputenc} % allow utf-8 input
\usepackage[T1]{fontenc}    % use 8-bit T1 fonts
\usepackage{url}            % simple URL typesetting
\usepackage{booktabs}       % professional-quality tables
\usepackage{amsfonts}       % blackboard math symbols
\usepackage{nicefrac}       % compact symbols for 1/2, etc.
\usepackage{microtype}      % microtypography
\usepackage{xcolor}         % colors

\usepackage{graphicx}
\usepackage{multirow}
\usepackage{caption}
\usepackage{amsmath}
\usepackage{amssymb}
\usepackage{xspace}
\usepackage{algorithm,algpseudocode}
\usepackage{color}
\usepackage{subcaption}
\definecolor{cvprblue}{rgb}{0.21,0.49,0.74}
\usepackage[pagebackref,breaklinks,colorlinks,citecolor=cvprblue]{hyperref}
\newcommand{\rankfirst}[1]{\textbf{#1}}
\newcommand{\ranksecond}[1]{\underline{#1}}
\usepackage{pifont}% 
\usepackage{colortbl}

\newcommand*{\ourmodel}{MASA\@\xspace}

\definecolor{codeblue}{rgb}{0.25,0.5,0.5}

\algnewcommand\algorithmicinput{\textbf{Input:}}
\algnewcommand\INPUT{\item[\algorithmicinput]}
\algnewcommand\algorithmicinputt{\textbf{Track:}}
\algnewcommand\TRACK{\item[\algorithmicinputt]}
\newcommand{\LineComment}[1]{\hfill \textcolor{codeblue}{\# #1}}

\def\paperID{310} % *** Enter the Paper ID here
\def\confName{CVPR}
\def\confYear{2024}

\title{Matching Anything by Segmenting Anything}

\author{
Siyuan Li\textsuperscript{1} \quad Lei Ke\textsuperscript{1} \quad Martin Danelljan\textsuperscript{1} \quad Luigi Piccinelli \textsuperscript{1}\\[0.1cm]
Mattia Segu\textsuperscript{1} \quad Luc Van Gool\textsuperscript{1,2} \quad Fisher Yu\textsuperscript{1}\\[0.4cm]
$^1$ETH Z\"urich \quad $^2$INSAIT \\
  \url{https://matchinganything.github.io} }

\begin{document}
\maketitle
\begin{abstract}
The robust association of the same objects across video frames in complex scenes is crucial for many applications, especially Multiple Object Tracking (MOT). Current methods predominantly rely on labeled domain-specific video datasets, which limits the cross-domain generalization of learned similarity embeddings.
We propose MASA, a novel method for robust instance association learning, capable of matching any objects within videos across diverse domains without tracking labels. Leveraging the rich object segmentation from the Segment Anything Model (SAM), MASA learns instance-level correspondence through exhaustive data transformations. We treat the SAM outputs as dense object region proposals and learn to match those regions from a vast image collection.
We further design a universal MASA adapter which can work in tandem with foundational segmentation or detection models and enable them to track any detected objects. Those combinations present strong zero-shot tracking ability in complex domains.
Extensive tests on multiple challenging MOT and MOTS benchmarks indicate that the proposed method, using only unlabeled static images, achieves even better performance than state-of-the-art methods trained with fully annotated in-domain video sequences, in zero-shot association. Our code is available at \href{https://github.com/siyuanliii/masa}{github.com/siyuanliii/masa}.

\end{abstract}    
\section{Introduction}

\begin{figure}[!t]
  \centering
\includegraphics[width=1\linewidth]{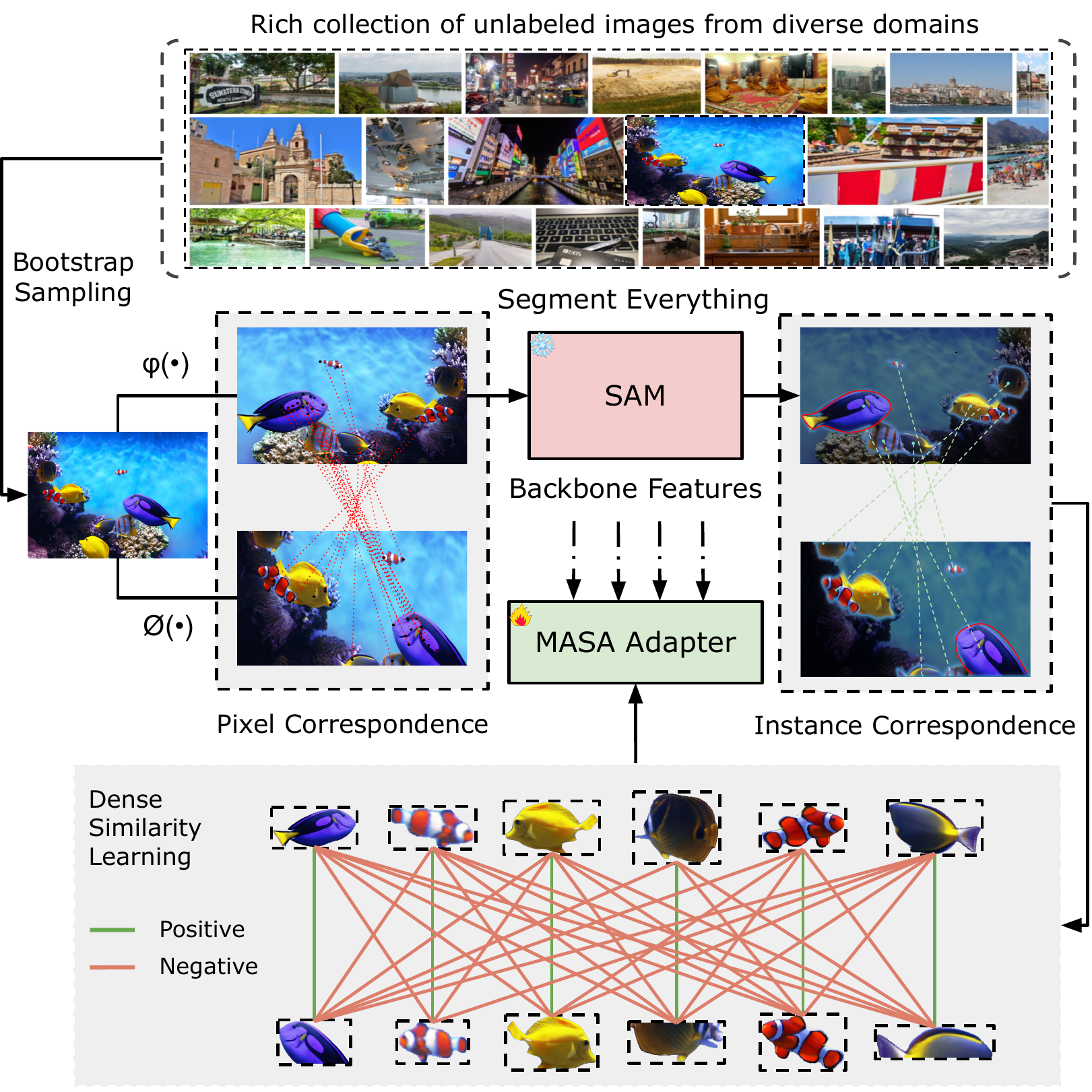}
  \caption{ Given an unlabeled image from any domain, we apply strong augmentations, $\varphi(\cdot)$ and $\phi(\cdot)$, to the image, generating two different views with automatically established pixel correspondences. Then, we leverage the rich object-level information encoded by the foundation segmentation model SAM to transfer the pixel-level to dense instance-level correspondence. Such correspondences enable us to utilize a diverse collection of unlabeled images to train a universal tracking adapter atop any segmentation or detection foundation models e.g. SAM. This adapter empowers the foundational models to track any objects they have detected, and shows strong zero-shot tracking ability in complex domains.} 
  \label{fig-teaser}
\end{figure}

Multiple Object Tracking (MOT) is one of the fundamental problems in computer vision. It plays a pivotal role in numerous robotics systems such as autonomous driving. 
Tracking requires both detecting the objects of interest in videos and associating them across frames.
While recent advancements in vision foundation models~\cite{SAM, Detic, GroundingDINO, Cascade-DETR, sam_hq, UniDepth} have demonstrated an exceptional ability to detect, segment, and perceive depth for any objects, associating those objects in videos remains challenging.
Recent successful multiple object tracking approaches~\cite{TETer,UNINEXT} have emphasized the importance of learning discriminative instance embeddings for accurate association. Some~\cite{QDTrack} even argued that it is the only necessary tracking component besides detection. \par
However, learning effective object association usually requires a significant amount of annotated data. 
While collecting detection labels on a diverse set of images is laborious, obtaining tracking labels on videos is even more challenging.
Consequently, current MOT datasets mostly focus on objects from a specific domain with a small number of fixed categories or a limited number of labeled frames. 

Training on those datasets limits the generalizability of tracking models to different domains and novel concepts. 
Although recent studies~\cite{GroundingDINO, Detic, SAM} have made successful attempts to address the model generalization issue for object detection and segmentation, the path to learning a universal association model for tracking any objects is still unclear. \par
Our goal is to develop a method capable of matching any objects or regions. We aim to integrate this generalizable tracking capability with any detection and segmentation methods to help them track any object they have detected. A primary challenge is acquiring matching supervision for general objects across diverse domains, without incurring substantial labelling costs. \par
To this end, we propose the \textbf{M}atching \textbf{A}nything by \textbf{S}egmenting \textbf{A}nything (MASA) pipeline to learn object-level associations from unlabeled images of any domain. \figureautorefname~\ref{fig-teaser} presents an overview of our MASA pipeline.
We leverage the rich object appearance and shape information encoded by the foundation segmentation SAM, combined with extensive data transformation, to establish strong instance correspondence. 

Applying different geometric transformations to the same image gives automatic pixel-level correspondence in two views from the same image. SAM's segmentation ability allows for the automatic grouping of pixels from the same instance, facilitating the conversion of pixel-level to instance-level correspondence. This process creates a self-supervision signal for learning discriminative object representation, utilizing dense similarity learning between view pairs. 
Our training strategy enables us to use a rich collection of raw images from diverse domains, demonstrating that such automatic self-training on diverse raw images provides excellent zero-shot multiple object tracking performance, even surpassing models reliant on in-domain video annotations for association learning. \par 

Beyond the self-training pipeline, we further build a universal tracking adapter \textemdash \ MASA adapter, to empower any existing open-world segmentation and detection foundation models such as SAM~\cite{SAM}, Detic~\cite{Detic} and Grounding-DINO~\cite{GroundingDINO} for tracking any objects they have detected. To preserve their original segmentation and detection ability, we freeze their original backbone and add the MASA adapter on the top.

Moreover, we propose a multi-task training pipeline that jointly performs the distillation of SAM's detection knowledge and instance similarity learning. This approach allows us to learn the object's location, shape and appearance prior of SAM, and simulate real detection proposals during contrastive similarity learning. This pipeline further improves the generalization capabilities of our tracking features. Additionally, our learned detection head speeds up the original SAM dense uniform point proposals for segmenting everything by over tenfold, crucial for tracking applications. 

We evaluate \ourmodel on multiple challenging benchmarks, including TAO MOT~\cite{TAO}, Open-vocabulary MOT~\cite{OVTrack},  MOT and MOTS on BDD100K~\cite{BDD100K}, and UVO~\cite{UVO}. 
Extensive experiments indicate that compared with state-of-the-art object tracking approaches trained on thoroughly in-domain labeled videos, our method achieves on-par or even better association performance, using a single model with the same model parameters and testing in zero-shot association settings.

\section{Related Work}
\label{sec:relatedwork}

\subsection{Learning Instance-level Association}

Learning robust instance-level correspondence is crucial to object tracking. Existing approaches can be divided into self-supervised~\cite{uniTrack} and supervised~\cite {QDTrack, IDOL, UNINEXT, TETer, MOTR, MOTRv2,sushi, Trackformer, JDE, unicorn} strategies. Specifically, as a representative self-supervised method, UniTrack~\cite{uniTrack} attempts to directly use off-the-shelf self-supervised representations~\cite{MoCov2, VFS} for association. Despite competitive results on some benchmarks~\cite{MOT17}, these methods cannot fully exploit instance-level training data, limiting their performance in challenging scenarios.
In contrast, supervised methods train discriminative instance embeddings on frame pairs, by contrastive learning. Although achieving superior performance on challenging benchmarks~\cite{BDD100K, TAO, Youtube, TAO-OW, OVTrack}, these methods rely on tremendous in-domain labeled video data. Several methods~\cite{HODOR,solotrack, CenterTrack, OVTrack, GTR} learn tracking signals from static images but still require substantial fine-grained instance annotations in specific domains or post-hoc test-time adaptation~\cite{segu2023darth}, limiting their ability for cross-domain generalization. To tackle these problems, we exploit the exhaustive object shape, and appearance information encoded by SAM to learn universal instance matching, purely from unlabeled images. Our learned representation shows exceptional zero-shot association ability across diverse domains. 

\subsection{Segment and Track Anything Models}
Deva~\cite{deva}, TAM~\cite{yang2023track} and SAM-Track~\cite{SAMTrack} integrate SAM~\cite{SAM} with video object segmentation (VOS) approaches (such as XMem~\cite{XMem} and DeAOT~\cite{DeAOT}) to enable an interactive pipeline for tracking any object, where SAM is mainly used for mask initialization/correction and XMem/DeAOT handle the tracking and prediction. 
SAM-PT~\cite{sampt} combines SAM with point-tracking methods such as ~\cite{CoTracker,PIPS, OmniMotion} to perform tracking.
However, all those approaches face limitations, such as poor mask propagation quality due to domain gaps and the inability to handle multiple diverse objects or rapid objects entry and exit, common in scenarios like autonomous driving.
Our work focuses on a different direction. Instead of building an interactive tracking pipeline or using off-the-shelf VOS or point-based trackers, we focus on learning universal association modules by leveraging SAM's rich instance segmentation knowledge. 

\section{Method}
\label{sec:method}

\subsection{Preliminaries: SAM}

SAM~\cite{SAM} is composed of three modules: \textbf{(a)} Image encoder: A heavy ViT-based backbone for feature extraction. \textbf{(b)} Prompt encoder: Modeling the positional information from the interactive points, box, or mask prompts. \textbf{(c)} Mask decoder: A transformer-based decoder takes both the extracted image embedding with the concatenated output and prompt tokens for final mask prediction. 
To generate all potential mask proposals, SAM adopts densely sampled regular grids as point anchors and generates mask predictions for each point prompt.
The complete pipeline includes patch cropping with greedy box-based NMS, three-step filtering, and heavy post-processing on masks. For more details on SAM's everything mode, we refer readers to~\cite{SAM}.

\subsection{Matching Anything by Segmenting Anything}
\begin{figure*}[!t]
\centering
\includegraphics[width=1.0\linewidth]{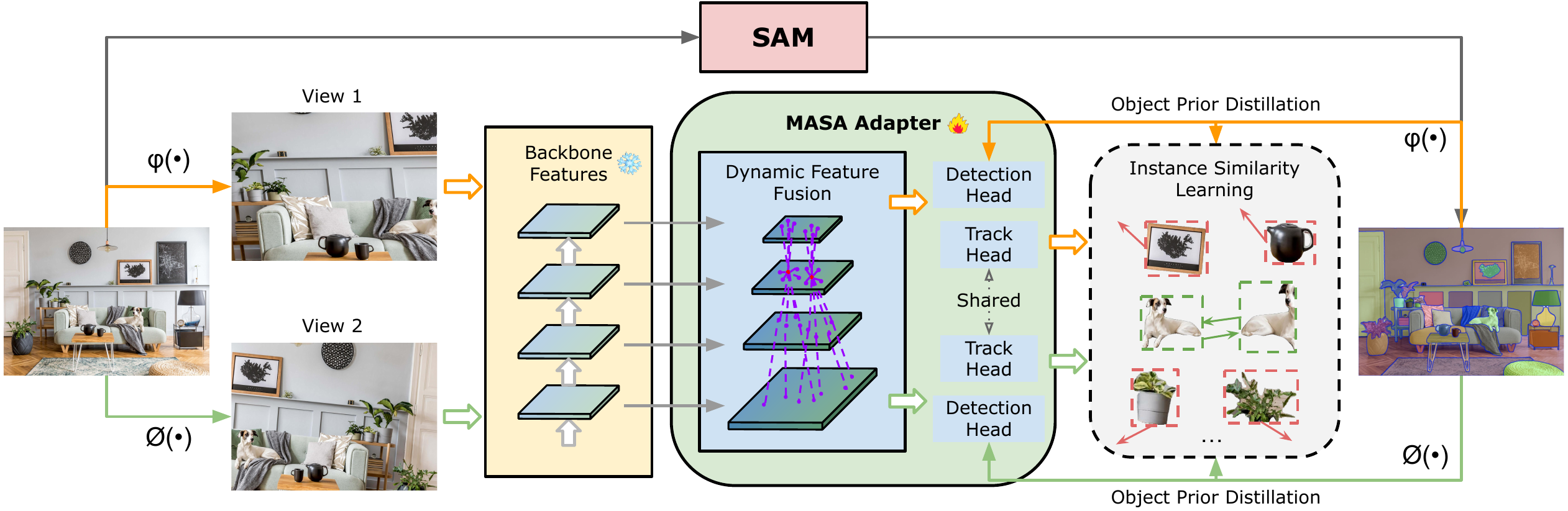}
  \caption{MASA training pipeline. Given an unlabeled image from any domain, SAM automatically generates exhaustive instance masks for it. Then we apply strong augmentations, $\phi(\cdot)$ and $\varphi(\cdot)$, to the original image and exhaustive instance segmentation, obtaining two different views as the inputs of our model. We train our MASA adapter by joint distillation of SAM's detection knowledge and instance similarity learning. Better view in color with zoom-in.} 
  \label{fig-data-pipeline}
  \vspace{-2mm}
\end{figure*}

Our method consists of two key components. First, based on SAM, we develop a new pipeline: MASA (Section~\ref{sec:masa_pipeline}). With this pipeline, we construct exhaustive supervision for dense instance-level correspondence from a rich collection of unlabeled images.
It enables us to learn strong discriminative instance representations to track any objects, without requiring any video annotations.
Second, we introduce a universal MASA adapter~(Section~\ref{sec:masa_adapter}) to effectively transform the features from a frozen detection or segmentation backbone for learning generalizable instance appearance representations. As a byproduct, the distillation branch of the MASA adapter can also significantly improve the efficiency of segmenting everything. 
Besides, we also construct a unified model to jointly detect / segment and track anything (Section~\ref{sec:unified}).
Our complete training pipeline is shown in ~\figureautorefname~\ref{fig-data-pipeline}. 

\subsubsection{MASA Pipeline} \label{sec:masa_pipeline}
To learn instance-level correspondence, previous works~\cite{QDTrack,unicorn,TETer,MOTRv2,UNINEXT} heavily relied on manually labeled in-domain video data. However, current video datasets~\cite{MOT17, nuscenes, BDD100K} contain only a limited range of fixed categories. This limited diversity in datasets leads to learning appearance embeddings that are tailored to specific domains, posing challenges in their universal generalization. \par
UniTrack~\cite{uniTrack} demonstrates that universal appearance features can be learned through contrastive self-supervised learning techniques~\cite{VFS, MoCov2, DINO-SSL} from raw images or videos. These representations, harnessing the diversity of a large volume of unlabeled images, can generalize across different tracking domains. However, they often depend on clean, object-centered images, such as those in ImageNet~\cite{ImageNet}, or videos like DAVIS17~\cite{DAVIS17}, and focus on frame-level similarities. This focus causes them to fail in fully leveraging instance information, leading to difficulties in learning discriminative instance representations in complex domains with multiple instances, as demonstrated in Table~\ref{tab:self-supervised}.

To address these issues, we propose the MASA training pipeline. Our core idea is to increase diversity from two perspectives: training \textit{image diversity} and \textit{instance diversity}. As shown in ~\figureautorefname~\ref{fig-teaser}, we first construct a rich collection of raw images from diverse domains to prevent learning domain-specific features. These images also contain a rich number of instances in complex environments to enhance instance diversity. Given an image $I$, we simulate appearance changes in videos by adopting two different augmentations on the same image. By applying strong data augmentations $\varphi(I)$ and $\phi(I)$, we construct two different views $V_1$ and $V_2$ of $I$, thereby automatically obtaining pixel-level correspondence. 
 
If the image is clean and contains only one instance, such as those in ImageNet, frame-level similarity can be applied as in~\cite{MoCov2, DINO, VFS}. However, with multiple instances, we need to further mine the instance information contained in such raw images. The foundational segmentation model SAM~\cite{SAM} offers us this capability. SAM automatically groups pixels belonging to the same instances and also provides the shape and boundary information of detected instances, valuable for learning discriminative features. 

Since we construct the dataset by selecting images with multiple instances, SAM's exhaustive segmentation of the entire images automatically yields a dense and diverse collection of instance proposals $Q$. With pixel-level correspondences established, applying the same $\phi(\cdot)$ and $\varphi(\cdot)$ to $Q$ transfers pixel-level correspondence to dense instance-level correspondence. This self-supervision signal enables us to use the contrastive learning formula from~\cite{SupCon, QDTrack, TETer} to learn a discriminative contrastive embedding space:

\begin{align*}
\mathcal{L_{C}} = -\sum_{q \in Q}{\mathrm{log}\frac{e^{\frac{\mathrm{sim}(q, q^+)}{\tau}}}{e^{\frac{\mathrm{sim}(q, q^+)}{\tau}} + \sum_{q^- \in Q^-}e^{\frac{\mathrm{sim}(q, q^-)}{\tau}}}},
\end{align*}

Here, $q^+$ and $q^-$ denote the positive and negative samples to $q$, respectively. Positive samples are the same instance proposals being applied different $\phi(\cdot)$ and $\varphi(\cdot)$. Negative samples are from different instances. Furthermore, $\mathrm{sim(\cdot)}$ denotes the cosine similarity and $\tau$ is a temperature parameter, set to 0.07 in our experiments. 

This contrastive learning formula pushes object embeddings belonging to the same instance closer while distancing embeddings from different instances. As demonstrated by existing works~\cite{SimCLR,QDTrack}, negative samples are crucial for learning discriminative representations. Under the contrastive learning paradigm, the dense proposals generated by SAM naturally provide more negative samples, thus enhancing learning better instance representation for association.

\subsubsection{MASA Adapter} \label{sec:masa_adapter}
We introduce the MASA adapter, designed to extend the open-world segmentation and detection models (such as SAM~\cite{SAM}, Detic~\cite{Detic}, and Grounding-DINO~\cite{GroundingDINO}) to track any detected objects. The MASA adapter operates in conjunction with frozen backbone features from these foundational models, ensuring their original detection and segmentation capabilities are preserved. However, as not all pre-trained features are inherently discriminative for tracking, we first transform these frozen backbone features into new features more suitable for tracking.

\begin{figure}[!t]
\centering%
\includegraphics[width=1.0\linewidth]{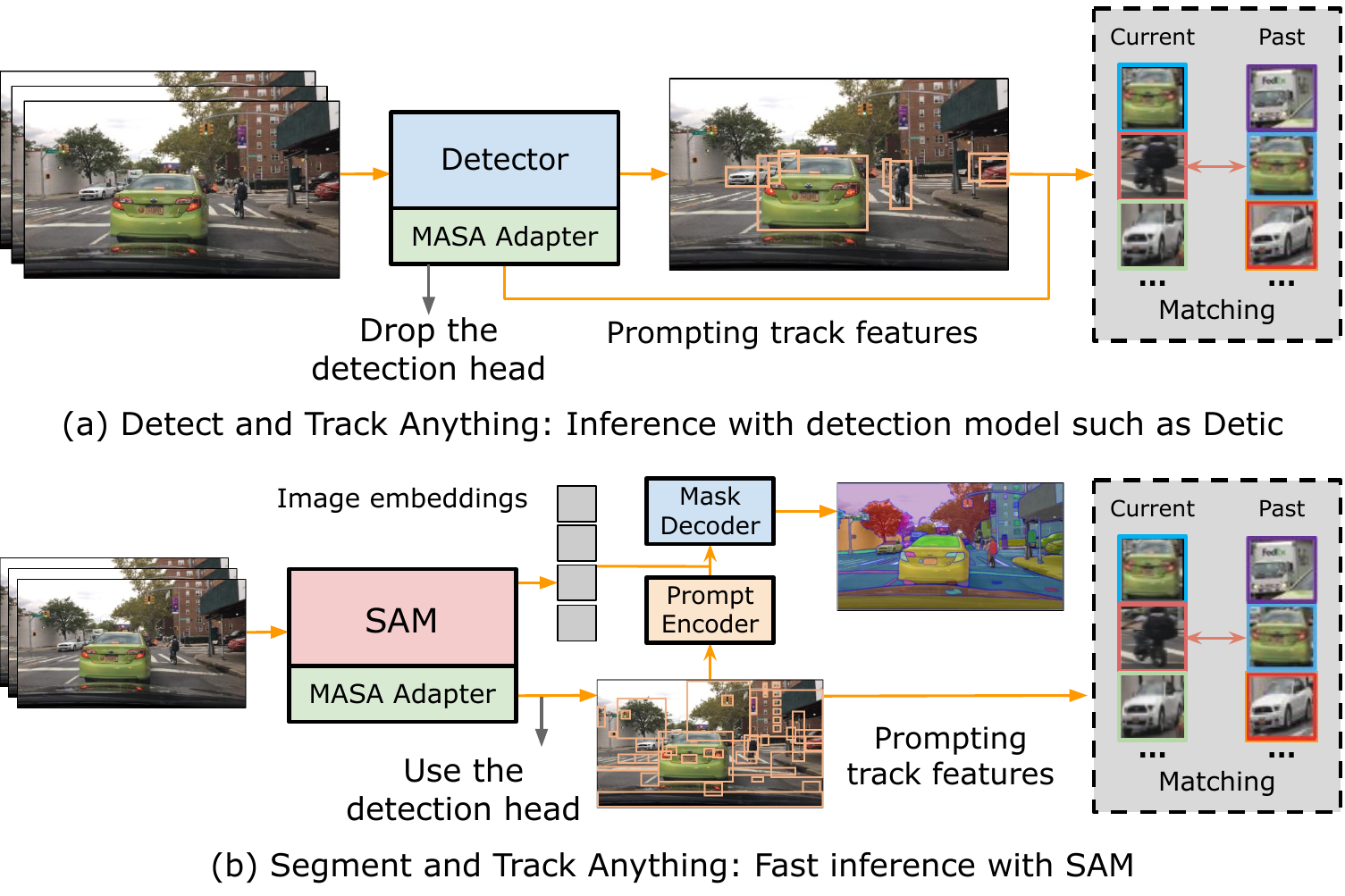}%
  \caption{The inference pipeline of our unified methods. }%
  \label{fig-inference}
  \vspace{3mm}
\end{figure}

Given the diversity in shapes and sizes of objects, we construct a multi-scale feature pyramid. For hierarchical backbones like the Swin Transformer~\cite{swin} in Detic and Grounding DINO, we directly employ FPN~\cite{FPN}. For SAM, which utilizes a plain ViT~\cite{ViT} backbone, we use Transpose Convolution and MaxPooling to upsample and downsample the single-scale features of stride $16\times$ to produce hierarchical features with scale ratios of ${\frac{1}{4}, \frac{1}{8}, \frac{1}{16}, \frac{1}{32}}$.
To effectively learn discriminative features for different instances, it's essential that objects in one location are aware of the appearances of instances in other locations. Hence, we use deformable convolution to generate dynamic offsets and aggregate information across spatial locations and feature levels as~\cite{DyHead}:
\begin{align}
\begin{split}
F(p) = \frac{1}{L} \sum_{j=1}^{L} \sum_{k=1}^{K} &w_k \cdot F^j(p + p_k + \Delta p_k^j) \cdot \Delta m_k^j,
\end{split}
\end{align}
where $L$ represents the feature level, $K$ is the number of sampling locations for a convolutional kernel, $w_k$ and $p_k$ are the weight and predefined offset for the $k$-th location, respectively, and $\Delta p_k^j$ and $\Delta m_k^j$ are the learnable offset and modulation factor for the $k$-th location at the $j$-th feature level. For SAM-based models, we additionally use task-aware attention and scale-aware attention from Dyhead~\cite{DyHead}, since the detection performance is important for accurate auto mask generation as in Figure~\ref{fig-inference} (b). 
After acquiring the transformed feature map, we extract instance-level features by applying RoI-Align~\cite{MaskRCNN} to the visual features $F$, followed by processing with a lightweight track head comprising $4$ convolutional layers and $1$ fully connected layer to generate instance embeddings. \par 
Additionally, we introduce an object prior distillation branch as an auxiliary task during training. This branch employs a standard RCNN~\cite{FasterRCNN} detection head to learn bounding boxes that tightly encompass SAM's mask predictions for each instance. It effectively learns exhaustive object location and shape knowledge from SAM and distils this information into the transformed feature representations. This design not only strengthens the features of the MASA adapter, resulting in improved association performance but also accelerates SAM's everything mode by directly providing the predicted box prompts. \par
The MASA adapter is optimized using a combination of detection and contrastive losses as defined in Section~\ref{sec:masa_pipeline}: $\mathcal{L} = \mathcal{L}_\text{det} + \mathcal{L}_C$. The detection loss is identical to that in~\cite{FasterRCNN}.

\subsubsection{Inference}
~\figureautorefname~\ref{fig-inference} shows the test pipeline with our unified models.
\label{sec:unified}

\noindent\textbf{Detect and Track Anything} When we integrate the MASA adapter with object detectors, we remove the MASA detection head that was learned during training. The MASA adapter then solely serves as a tracker. The detectors predict the bounding boxes, and then they are utilized to prompt the MASA adapter, which retrieves corresponding tracking features for instance matching. We use a simple bi-softmax nearest neighbor search for accurate instance matching, as illustrated in Section~\ref{sec:supp-infer-details} of the Appendix.

\noindent{\textbf{Segment and Track Anything}} With SAM, we keep the detection head. We use it to predict all potential objects within a scene, forwarding box predictions as prompts to both the SAM mask decoder and the MASA adapter for segmenting and tracking everything. The predicted box prompts omit the need for the heavy post-processing illustrated in the original SAM's everything mode, therefore, significantly speeding up the auto mask generation of SAM.

\noindent{\textbf{Testing with Given Observations}}
When detections are obtained from sources other than the one the MASA adapter is built upon, our MASA adapter serves as a tracking feature provider. We directly utilize the provided bounding boxes as prompts to extract tracking features from our MASA adapter through the ROI-Align~\cite{MaskRCNN} operation.
\section{Experiments}
\label{sec:exp}

We perform experiments on multiple challenging MOT/MOTS benchmarks with diverse domains. 
\begin{table}[t]
    \caption{State-of-the-art comparison on TAO TETA benchmark~\cite{TETer}. $^{\dagger}$ indicates using the same detection observations. Our zero-shot models can achieve better performance compared with the state-of-the-art fully supervised methods using the same detection observations. The performance can be even better when using original detections from our unified model.}%
    \centering%
    \resizebox{.8\linewidth}{!}{\resizebox{\linewidth}{!}{
\begin{tabular}{lccccccc}
    \toprule 
    Method & TETA & LocA & AssocA & ClsA \\
    \hline%\hline
    \multicolumn{5}{l}{\underline{\textit{Fully-supervised, in-domain}}}  \\
    SORT~\cite{SORT} & 24.9 & 48.1 & 14.3 & 12.1 \\
    Tracktor~\cite{Tracktor} & 24.2 & 47.4 & 13.0 & 12.1 \\
    Tracktor++~\cite{Tracktor} & 28.0 & 49.0 & 22.8 & 12.1 \\
    DeepSORT~\cite{DeepSORT} & 26.0 & 48.4 & 17.5 & 12.1 \\
    AOA~\cite{AOA} & 25.3 & 23.4 & 30.6 & \textbf{21.9} \\
    QDTrack~\cite{QDTrack} & 30.0 & 50.5 & 27.4 & 12.1 \\
    TETer~\cite{TETer}$^{\dagger}$ & {34.6} & \rankfirst{52.1} & \ranksecond{36.7} & 15.0 \\
    \midrule
    \multicolumn{5}{l}{\underline{\textit{Self-supervised, zero-shot}}}  \\
    \textbf{Ours-Detic}$^{\dagger}$ & \ranksecond{34.7} & \ranksecond{51.9} & 36.4 & \ranksecond{15.8} \\
    \textbf{Ours-Grounding-DINO}$^{\dagger}$ & \rankfirst{34.9} & 51.8 & \rankfirst{37.6} & {15.4} \\
    \textbf{Ours-SAM-B}$^{\dagger}$ & 34.5 & {51.8} & 36.6 & 15.1 \\
    \textbf{Ours-SAM-H}$^{\dagger}$ & 34.5 & 51.8 & 36.4 & 15.4 \\ \midrule
    \textbf{Ours-Detic}   & \textbf{46.3} & \textbf{65.8} & \textbf{44.1} & \textbf{28.9} \\ %\bottomrule
    \bottomrule
\end{tabular}
}

}%
    \label{tab:tao}%
    %\vspace{5mm}%
\end{table}

\begin{table}[t]
\vspace{4mm}
    \caption{State-of-the-art comparison on open-vocabulary MOT benchmark~\cite{OVTrack}. All methods are trained with base annotations.}%
    \centering%
    \resizebox{1\linewidth}{!}{
\begin{tabular}{l|cccc|cccc}
\toprule
\multirow{2}{*}{Method} & \multicolumn{4}{c|}{Base} & \multicolumn{4}{c}{Novel}\\ 
& TETA & LocA & AssocA & ClsA & TETA & LocA & AssocA & ClsA  \\ \midrule
DeepSORT ~\cite{DeepSORT} & 28.4 &	52.5 &	15.6 &	17.0 &24.5 &	49.2 & 15.3  & 9.0 \\
Tracktor++ ~\cite{Tracktor} & 29.6 & 52.4 & 19.6 & 16.9 & 25.7 & 50.1 & 18.9 & 8.1 \\
Bytetrack~\cite{bytetrack} & 29.5 & 51.7 & 19.7 & 17.2 & 25.4 & 49.4 & 18.1 & 8.7 \\
OC-SORT~\cite{OC-SORT} & 30.0 & 53.3 & 23.5 & 13.3 & 26.7 & 51.5 & 21.6 & 7.1 \\
OVTrack~\cite{OVTrack} & 36.3 & 53.9 & 36.3 & 18.7 & 32.0 & 51.4 & 33.2 & 11.4 \\ \midrule
\textbf{Ours-Detic} & \textbf{47.0} & \textbf{66.0} & \textbf{44.5} & \textbf{30.5} & \textbf{40.8} & \textbf{64.4} & \textbf{41.2} & \textbf{17.0} \\
\bottomrule
\end{tabular}
}%
    \label{tab:ovtrack_tao}%
    \vspace{5mm}
\end{table}

\begin{table}[t]
    \caption{State-of-the-art comparison on TAO Track mAP benchmark.$^{\dagger}$ indicates using same detections with GTR. Note that GTR is an offline tracking method while ours is online.}%
    \centering%
    \resizebox{.95\linewidth}{!}{\resizebox{\linewidth}{!}{
\begin{tabular}{lccc}
\toprule
Method        & Track mAP50 & Track mAP75 & Track mAP \\ \midrule
\multicolumn{4}{l}{\underline{\textit{Fully-supervised, in-domain}}}\\
SORT-TAO~\cite{TAO} & 13.2       & -          & -        \\
QDTrack~\cite{QDTrack}   & 15.9       & 5.0          & 10.6     \\
TAC~\cite{tac}     & 17.7      & 5.8       & 7.3     \\
BIV~\cite{biv}  & 21.6      & 10.4       & 16.1    \\
{GTR}$\dagger$~\cite{GTR}  & {22.5}      & -          & -     \\ \midrule 
\multicolumn{4}{l}{\underline{\textit{Self-supervised, zero-shot}}}\\
\textbf{Ours-Detic}$\dagger$ & {22.0}      & {12.2}      & {17.1}    \\  
\textbf{Ours-Grounding-DINO}$\dagger$ & {22.8}      & \ranksecond{12.3}      & \ranksecond{17.6}    \\  
\textbf{Ours-SAM-B}$\dagger$ & \textbf{23.9}      & \textbf{13.0}      & \textbf{18.4}    \\  
\textbf{Ours-SAM-H}$\dagger$ & \ranksecond{22.9}      & {12.1}      & {17.5}  \\  \midrule
\textbf{Ours-Detic} & \textbf{30.9}      & \textbf{18.0}      & \textbf{24.4} \\
\bottomrule
\end{tabular}
}}%
    \label{tab:tao_trackmAP}%
    \vspace{5mm}
\end{table}

\begin{table}[t]
    % \tablefontsize
    \caption{State-of-the-art comparison on BDD MOTS benchmark (validation set). $^{\dagger}$ represents that we provide the same detection observations. AssocA, mIDF1, and IDF1 mainly evaluate the association quality. MASA achieves the best results on all metrics.}
    \centering
    {\resizebox{1\linewidth}{!}{
    \begin{tabular}{lccccc}
        \toprule
        Method & mIDF1$\uparrow$ & AssocA$\uparrow$ & TETA$\uparrow$ & mMOTSA$\uparrow$ & mHOTA$\uparrow$ \\
        \midrule
        \underline{\textit{Fully-supervised, in-domain}} & & & & &  \\
        MaskTrackRCNN~\cite{VIS} & 26.2 & - & - & 12.3 & - \\
        STEm-Seg~\cite{stemseg} & 25.4 & - & - & 12.2 & - \\
        QDTrack-mots~\cite{QDTrack} & 40.8 & - & - & 22.5 & - \\
        PCAN~\cite{PCAN} & 45.1 & 46.7 & 46.8 & 27.4 & 35.9 \\
        VMT~\cite{VMT} & 45.7 & 47.3 & 47.1 & 28.7 & 36.6 \\
        Unicorn~\cite{unicorn} & 44.2 & - & - & 29.6 & - \\
        UNINEXT-H~\cite{UNINEXT}$^{\dagger}$ & 48.5 & 53.2 & {53.6} & {35.7} & {40.6} \\
        \midrule
        \underline{\textit{Self-supervised, zero-shot}} & & & & &  \\
        \textbf{Ours-Detic}$^{\dagger}$ & \ranksecond{49.5} & {53.5} & {54.4} & \textbf{36.4} & 40.2 \\
        \textbf{Ours-Grounding-DINO}$^{\dagger}$ & {48.6} & {52.3} & {54.0} & \ranksecond{36.1} & {40.0} \\
        \textbf{Ours-SAM-B}$^{\dagger}$ & {49.2} & \ranksecond{53.9} & \rankfirst{54.8} & 35.2 & \ranksecond{40.7} \\
        \textbf{Ours-SAM-H}$^{\dagger}$ & \rankfirst{49.7} & \rankfirst{54.5} & \ranksecond{54.7} & {35.8} & \rankfirst{40.8} \\
        \bottomrule
    \end{tabular}
}

}
    \label{tab:mots}
    \vspace{5mm}
\end{table}

\begin{table}[t]
    \caption{State-of-the-art comparison on BDD MOT benchmark (validation set). $^{\dagger}$ represents that we provide the same detection observations.}
    \centering
    {\resizebox{1\linewidth}{!}{
    \begin{tabular}{lcccccc}
        \toprule
        Method & mIDF1$\uparrow$ & IDF1$\uparrow$ & TETA$\uparrow$ & AssocA$\uparrow$ & mMOTA$\uparrow$ \\
        \midrule
        \underline{\textit{Fully-supervised, in-domain}} & & & & &  \\
        QDTrack~\cite{QDTrack} & 50.8 & 71.5 & 47.8 & 48.5 & 36.6 \\
        TETer~\cite{TETer} & 53.3 & 71.1 & 50.8 & \rankfirst{52.9} & 39.1 \\
        MOTR~\cite{MOTR} & 54.0& 65.8 & -&-&32.3 \\
        Unicorn~\cite{unicorn} & 54.0 & 71.3 & - & - & 41.2 \\
        UNINEXT-H~\cite{UNINEXT} & \rankfirst{56.7} & 69.9 & - & - & 44.2 \\
        ByteTrack~\cite{bytetrack}$^{\dagger}$ & 54.8 & 70.4 & \rankfirst{55.7} & 51.5 & \rankfirst{45.5} \\
        \midrule
        \underline{\textit{Self-supervised, zero-shot}} & & & & &  \\
        \textbf{Ours-Detic}$^{\dagger}$ & \ranksecond{55.8} & {71.3} & 54.4 & \textbf{52.9} & \ranksecond{44.6} \\
        \textbf{Ours-Grounding-DINO}$^{\dagger}$ & {55.6} & \rankfirst{71.7} & \ranksecond{54.5} & \ranksecond{52.7} & {44.5} \\
        \textbf{Ours-SAM-B}$^{\dagger}$ & {55.6} & \ranksecond{71.6} & 54.0 & {52.6} & 44.1 \\
        \textbf{Ours-SAM-H}$^{\dagger}$ & {55.3} & \rankfirst{71.7} & {54.2} & {51.9} & {44.5} \\
        \bottomrule
    \end{tabular}
}

}
    \label{tab:bdd}
\end{table}

\subsection{Experimental Setup}
\noindent\textbf{TAO MOT} TAO dataset~\cite{TAO} is designed to track a diverse range of objects, encompassing over 800 categories, making it the most diverse MOT dataset with the largest class collection to date. It contains 500, 988, and 1,419 videos annotated at 1 FPS in the train, validation, and test sets, respectively. We report performances on the validation set. TAO comprises several benchmarks, each highlighting different characteristics and requirements. The TAO TETA benchmark~\cite{TETer} emphasizes association by rewarding trackers that produce clean trajectories with no overlaps. Conversely, the TAO Track mAP benchmark~\cite{TAO} values particularly the classification of trajectories, and does not heavily penalize overlapping trajectories. The open-vocabulary MOT benchmark~\cite{OVTrack} requires trackers to avoid training with annotations from novel classes, focusing on the generalization ability to track novel categories. 

\noindent\textbf{BDD100K MOT~\cite{BDD100K}} requires trackers to track common objects in autonomous driving scenarios. The dataset is annotated at 5 FPS with 200 videos in the validation set.

\noindent\textbf{BDD100K MOTS} Different from BDD100K MOT, BDD100K MOTS~\cite{BDD100K} requires trackers to track and segment objects simultaneously, evaluating tracking performance on masks. There are 154 videos for training, 32 videos for validation, and 37 videos for testing.

\noindent\textbf{UVO~\cite{UVO}} is a challenging benchmark for open-world instance segmentation in videos. Compared with previous video-level object segmentation datasets~\cite{VIS}, it annotates much more diverse instances. UVO has two evaluation tracks, an image track, and a video track. We evaluate all methods on the UVOv0.5 validation set.

\noindent\textbf{Evaluation Metrics} As analyzed in previous works~\cite{TETer}, traditional tracking metrics like mMOTA~\cite{BDD100K}, and track mAP~\cite{TAO} can be misleading, particularly in long-tail scenarios, due to their high sensitivity to classification. To address this issue, \cite{TETer} introduced TETA, a new tracking metric that decomposes into three separate components: AssocA, LocA, and ClsA, reflecting the accuracy of association, localization, and classification, respectively. In standard MOT benchmarks, to ensure a fair comparison of trackers' association abilities, we adopt the same detection observations used by leading state-of-the-art trackers. Therefore, our focus is primarily on \textbf{association-related} metrics like \textbf{AssocA}, \textbf{mIDF1}, and \textbf{IDF1}. Additionally, when evaluating our unified models, we consider the full spectrum of metrics to capture their comprehensive capabilities.
Particularly for open-world segmentation on UVO, our emphasis is on AR100 and Track AR100 metrics in the image and video levels. This is due to the fact that SAM often segments every part of an object, whereas UVO lacks such detailed annotations, making traditional AP evaluations less accurate.

\noindent\textbf{Training Data} SA-1B~\cite{SAM} consists of 11M diverse, high-resolution images, containing diverse scenarios with multiple object interactions in complex environments. We sub-sample the SA-1B raw images to construct a training set of 500K images, SA-1B-500K. 

\noindent\textbf{Implementation Details} For our models, we utilize the official weights of SAM~\cite{SAM}, Detic, and Grounding-DINO, ensuring that all components of these models remain frozen during the training phase. Specifically, we employ SAM with both ViT-Base and ViT-Huge backbones, and Detic and Grounding-DINO are used with the SwinB backbone. We train the models with bootstrapping sampling for 200,000 images per epoch, with a batch size of 128. We use SGD with an initial learning rate of 0.04, coupled with a step policy for learning rate decay. Momentum and weight decay parameters are set to 0.9 and 1e-4. Our training spans 12 epochs, with the learning rate being reduced at the 8th and 11th epochs. For data augmentation, we use random affine, MixUp~\cite{mixup}, and Large-scale Jittering~\cite{copy-paste}, in addition to standard practices like flipping, color jittering, and random cropping. More details are provided in Section~\ref{sec:supp-implementation_detail} of the Appendix.

\subsection{State-of-the-Art Comparison}

We evaluate our methods in two ways. Firstly, to accurately assess the association ability of our method, we always provide the same detection observations as current state-of-the-art methods in standard MOT benchmarks. Secondly, to evaluate the integrated abilities of our unified models, we follow this protocol: for SAM-based models, we evaluate on the open-world video segmentation dataset UVO. For the detectors-based models, we evaluated on the open-vocabulary MOT benchmark~\cite{OVTrack}. We also report the scores on TAO TETA and TAO TrackmAP benchmarks. Note that we perform zero-shot association tests for all our variants, and use the same weights across all benchmarks.

\noindent\textbf{TAO TETA} We use the same observations as TETer-SwinT~\cite{TETer}.  As shown in Table~\ref{tab:tao}, our method with Grounding-DINO's backbone performs the best, in the zero-shot setting, without training on any in-domain labeled videos, on both AssocA and TETA. We also test our unified Detic model which jointly outputs the detection and tracking results. It outperforms all other methods significantly and achieves the new state-of-the-art. It demonstrates our method can couple well with current detection foundation models and transfer their strong detection ability into tracking.

\noindent\textbf{Open-vocabulary MOT}
Similar to the open-vocabulary object detection task~\cite{VILD}, open-vocabulary MOT~\cite{OVTrack} stipulates that methods should only use the frequent and common classes annotations from LVIS~\cite{LVIS} for training, treating the rare classes as novel. We evaluated our unified 'detect and track anything' model Detic, which was trained exclusively with base class annotations. ~\tableautorefname~\ref{tab:ovtrack_tao} shows our unified Detic model outperforms existing models on all metrics across both base and novel splits, and it achieves this significant lead despite our tracker being trained solely with out-of-domain, unlabeled images.

\noindent\textbf{TAO Track mAP} We use the same observations as GTR~\cite{GTR}.  As shown in Table~\ref{tab:tao_trackmAP}, our method with SAM-B performs the best (Track mAP50 of 23.9) given the same detections. Most of our models outperform the current state-of-the-art 
GTR, which is an offline method that utilizes future information for association. In contrast, our methods conduct tracking in an online fashion and test in a zero-shot setting. Our unified Detic model again, achieves the new state-of-the-art by outperforming GTR by a large margin.

\noindent\textbf{BDD100K MOTS} We use the same observations as the state-of-the-art method, UNINEXT-H~\cite{UNINEXT} and perform zero-shot association test on BDD100K MOTS benchmark. As shown in Table~\ref{tab:mots}, our method achieves the best association performance (mIDF1 of 49.7 and AssocA of 54.5) among all approaches. This demonstrates the superiority of the instance embeddings learned by our method.

\noindent\textbf{BDD100K MOT} As shown in Table~\ref{tab:bdd}, given the same observations as ByteTrack ~\cite{bytetrack}, our method achieves the best IDF1 of 71.7 and AssocA 52.9. Compared with state-of-the-art ByteTrack~\cite{bytetrack}, our method also achieves better association performance, being about 1.4\% higher on both IDF1 and AssocA, without using any BDD images for training. ByteTrack additionally selects low-confidence boxes and adds them to the tracklets, resulting in a better mMOTA score which prioritizes detection performance ~\cite{HOTA}.

\begin{figure}[t]
\centering
\small
    \begin{minipage}{0.42\linewidth} 
    \centering%
    {%
        \resizebox{\linewidth}{!}{    \resizebox{1\linewidth}{!}{%
        \begin{tabular}{llc}
        \multicolumn{3}{l}{{(a) Quantitative results on UVO~\cite{UVO} dataset.}} \\

            \toprule
             Track & Method & AR100 \\
            \midrule
            \multirow{3}{*}{image} & Mask R-CNN~\cite{MaskRCNN} & 41.3 \\
            & \textbf{Ours-SAM-B} & \ranksecond{43.7} \\
            & \textbf{Ours-SAM-H} & \rankfirst{50.8} \\
            \midrule
            &\multicolumn{2}{l}{\underline{\textit{Fully-supervised, in-domain}}}  \\
            \multirow{6}{*}{video} & MaskTrack R-CNN~\cite{VIS}& 17.2\\
            & QDTrack~\cite{QDTrack} & 20.9 \\
            & SeqFormer~\cite{SeqFormer} & 21.3 \\
            &\multicolumn{2}{l}{\underline{\textit{Zero-shot test}}}  \\
            & \textbf{Ours-SAM-B} & \ranksecond{24.9} \\
            & \textbf{Ours-SAM-H} & \rankfirst{28.4} \\
            \bottomrule
        \end{tabular}%
        }}%
        % \vspace{-1mm}%
    }
    \end{minipage}
    \quad
    \begin{minipage}{0.53\linewidth} 
    \centering%
    {%
        \includegraphics[width=\linewidth]{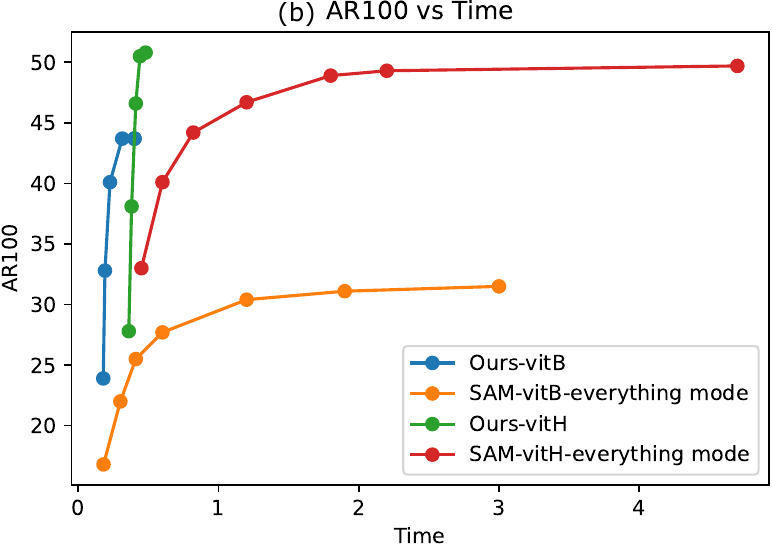}%
        %\vspace{-4mm}%
    }
    \end{minipage}
\\
\caption{Comparison on the UVO~\cite{UVO} dataset. (a) We evaluate class-agnostic object detection and video object tracking results with our \ourmodel. Both object localization and association achieve promising performance compared with previous in-domain training methods. 
(b) We compare the inference time (s) with the original SAM by sampling different numbers of prompt points. Our detection head learns to localize all the potential objects effectively.}%
\label{tab:uvo}
\end{figure}

\noindent\textbf{UVO VIS} We perform zero-shot tests for our unified 'segment and track anything' model based on SAM. We directly use the box prompts from the MASA detection head for faster segmenting everything. As shown in Figure~\ref{tab:uvo}a, our method achieves the best performance on both image and video tracks, outperforming its counterparts by a large margin. 
 Besides, we also compare our method with SAM's default auto mask segmentation. As shown in Figure~\ref{tab:uvo}b, as the inference time increases, AR100 of our method grows much faster than SAM due to the distillate detection branch. The upper bound AR100 of our method with ViT-Base backbone even surpasses SAM by 10\%. Besides, when achieving the same AR100, our method is about $10\times$ faster than SAM. This stems from the fact that our method learns a strong object prior to capturing potential objects with a small number of sparse proposals. However, to segment everything, SAM has to sample about $1k$ points evenly, which is inflexible and inefficient, while also relying on hand-crafted complex post-processing methods.

\noindent\textbf{Compare with VOS Methods}
We evaluated the VOS-based method Deva~\cite{deva}, which integrates XMem~\cite{XMem} for tracking multiple objects and SAM-PT~\cite{sampt}, which uses point-tracking. To ensure a fair comparison, we provide the same observations on BDD MOTS, TAO TETA  and UVO benchmarks. For UVO, we use SAM's auto-mask generation to generate masks first, then we resolve the overlapping masks following the heuristic in Deva~\cite{deva} and use Deva to generate per-frame observations.  

Table~\ref{tab:vos_compare} shows that our method outperforms Deva across all benchmarks. Notably, on the autonomous driving BDD100K benchmark, where objects frequently enter and exit the scene, VOS-based methods like Deva are prone to a significant increase in false positives. This is reflected in the TETA scores, where such errors are heavily penalized. Additionally, Deva struggles with overlapping predictions, a common issue with current detection models. We provide a more in-depth analysis in Section~\ref{sec:supp-vos} of the Appendix.

\noindent\textbf{Compare with Self-supervised Methods}
We further compare our approach with self-supervised methods aimed at learning universal appearance features from raw images or videos. To ensure a fair comparison, we train all methods using a mix of BDD and COCO raw images. Specifically, for VFS, we utilize raw videos from BDD. We employ a ResNet-50 model for VFS~\cite{VFS} and MoCov2~\cite{MoCov2}, and a ViT-B model for DINO~\cite{DINO-SSL}, following the association tracking strategy outlined in UniTrack~\cite{uniTrack}. Additionally, we ensure that detection observations are identical across all models. Table~\ref{tab:self-supervised} demonstrates that our methods significantly outperform other self-supervised approaches. This advantage stems from the fact that traditional self-supervised learning primarily focuses on frame-level similarities, which limits their effectiveness in leveraging instance information and causes struggles when training with images containing multiple objects. Further analysis of this is provided in Section~\ref{sec:supp-ssl-compare} of the Appendix.

\begin{table}[t]
\vspace{3mm}
    \caption{Compare with VOS methods. $^{\dagger}$ represents that we provide the same detection observation as inputs.}%
    \centering%
    \resizebox{.6\linewidth}{!}{
    \resizebox{0.8\linewidth}{!}{%
\begin{tabular}{@{}l|cc|c|c@{}}
\toprule
\multirow{2}{*}{Method} & \multicolumn{2}{c|}{BDD MOTS} & \multicolumn{1}{c|}{TAO}& \multicolumn{1}{c}{UVO} \\
& AssocA & TETA & AssocA & AR100 \\
\midrule
SAM-PT~\cite{sampt} &-&-&-&31.8\\
Deva$^{\dagger}$~\cite{deva} & 46.8 & 32.1 & 22.4 & 36 \\
\textbf{Ours-SAM-H$^{\dagger}$ }& \textbf{54.5} & \textbf{54.7} & \textbf{36.4} & \textbf{37.5} \\
\bottomrule
\end{tabular}
}}%
    \label{tab:vos_compare}%
\end{table}

\begin{table}[t]
    \caption{Compare with self-supervised based methods. All methods use the same BDD and COCO raw images for training and the same detections for testing.}%
    \centering%
    \resizebox{.95\linewidth}{!}{\begin{tabular}{@{}l|c|cc|c|cc@{}}
\toprule
\multirow{2}{*}{Method} &\multirow{2}{*}{Video}& \multicolumn{2}{c|}{BDD MOT} & \multicolumn{1}{c|}{TAO} & \multicolumn{2}{c}{BDD MOTS} \\
& & AssocA &mIDF1 & AssocA & AssocA &mIDF1 \\
\midrule
\underline{\textit{Train on BDD \& COCO}} &&&&&&\\
VFS~\cite{VFS} & \ding{51} & 29.2 & 35.0 & 19.1 & 30.7 & 30.1 \\
MoCov2~\cite{MoCov2} &\ding{55} & 42.7 & 46.7 & 30.7 & 51 & 45.3 \\
DINO~\cite{DINO-SSL} &\ding{55} & 23.1 & 16.8 & 12.9 & 20.2 & 22.2 \\
\textbf{Ours-SAM-B} &\ding{55} & \textbf{51.9} & \textbf{54.9} & \textbf{35.8} & \textbf{53.7} & \textbf{49.1} \\
\bottomrule
\end{tabular}
}%
    \label{tab:self-supervised}%
   \vspace{5mm}
\end{table}

\subsection{Ablation Study and Analysis}
To reduce the training costs, we bootstrap fewer raw images (40K) for training for the ablation experiments. Unless specified we train the model with an image collection containing 70k raw images from \cite{BDD100K} and 110k images from \cite{COCO} training set respectively. We employ the Ours-SAM-B model and test on BDD MOT and TAO TETA benchmarks.

\noindent\textbf{Effect of Training Strategies and Model Architectures} Table~\ref{tab:ablation_model} illustrates that directly using the off-the-shelf SAM features (row 1) for association yields poor results. The primary reason is that SAM's original features are optimized for segmentation, not for instance-level discrimination. However, integrating our MASA training approach and adding a lightweight track head significantly enhances performance, yielding improvements of $15.6\%$ in AssocA and $14.4\%$ in mIDF1 on BDD MOT. This underscores the efficacy of our training strategy. Incorporating a dynamic feature fusion block further enhances performance by $1.6\%$. Additionally, joint training with the object prior distillation branch leads to an increase of $1.8\%$ in AssocA and $1.6\%$ in mIDF1, showing the effect of these architectural designs. 

\begin{table}[t]
    %\vspace{-1.5mm}
    \caption{Effect of training strategies and model architectures. The performance is evaluated on BDD MOT~\cite{BDD100K} dataset.}%
    \centering%
            \resizebox{1\linewidth}{!}{
        \begin{tabular}{c|c|c|c|c}
            \toprule
              MASA training  & Dynamic feature fusion & Object prior distillation &AssocA & mIDF1 \\
            \midrule
               &  & &  32.9 &   37.3  \\
             \ding{51}& & & 48.5 &51.7 \\
             \ding{51}&\ding{51}& & 50.1 &53.3 \\
             \ding{51}&\ding{51}&\ding{51}& 51.9 &54.9 \\
            \bottomrule
        \end{tabular}
    }%
    \label{tab:ablation_model}%
    %\vspace{4mm}
\end{table}

\begin{table}[t]
\small
\vspace{4mm}
\caption{Ablation study on different augmentations strategies, proposal quality and quantity.}
\centering
    \begin{minipage}[t]{0.47\linewidth} 
    \centering 
    {
     \vspace{-10.8mm}
         \resizebox{\linewidth}{!}{
    \begin{tabular}{l|c|c}
    \multicolumn{3}{c}{(a) The effect of proposal quality.} \\
        \toprule
         & AssocA & AssocA  \\ 
        \midrule
          Mask2Former & 46.4 & 29.8 \\ 
          SAM  & 50.9 & 34.1 \\ 
        \bottomrule
    \end{tabular}
}
        
    }
    
    \end{minipage}
    \begin{minipage}[t]{0.47\linewidth} 
    \centering 
    {
            \resizebox{.95\linewidth}{!}{%
\begin{tabular}{@{}c|c|cc@{}}
\multicolumn{4}{c}{(b) The effect of proposal number.}\\
\toprule
\multirow{2}{*}{Instance Number} & \multicolumn{1}{c|}{BDD} & \multicolumn{1}{c}{TAO} \\
& AssocA  & AssocA \\
\midrule
64 & 47.1& 31.5 \\
128 & 50.9& 34.6 \\
256 & 51.9& 35.8 \\
\bottomrule
\end{tabular}
}
    }
    \end{minipage}
    \quad
    \vspace{1mm}
\\

    \begin{minipage}[t]{0.6\linewidth}
    \centering 
    {
            \resizebox{\linewidth}{!}{
        \begin{tabular}{lccc|cc|c}
            \multicolumn{6}{c}{(c) The effect of data augmentation.}\\
            \toprule
            \multirow{2}{*}{\#}  & \multirow{2}{*}{Affine} & \multirow{2}{*}{Mixup} & \multirow{2}{*}{LSJ} & \multicolumn{2}{c|}{BDD MOT}  & TAO\\
            &&&&mIDF1 & AssocA  & AssocA \\
            \midrule
            1&&&&48.2&43.9&28.5\\
            2&\ding{51}&&&53.0&49.1&33.3\\
            3&&\ding{51}&&52.8&49.9&31.5\\
            4&&&\ding{51}&52.9&49.3&32.3\\
            5&\ding{51}&\ding{51}&\ding{51}&54.9&51.9&35.8\\
            \bottomrule
        \end{tabular}
        }

    }
    \end{minipage}
\\%
\label{tab:ablation}%
\end{table}

\noindent\textbf{Effect of Proposal Diversity}
We evaluate different proposal generation mechanisms in association learning. We use only raw images from the training set of the BDD detection task for training. By substituting SAM in our MASA pipeline with Mask2former-SwinL~\cite{Mask2Former}, pre-trained on COCO. As shown in Table~\ref{tab:ablation}a, we found that the model trained with SAM's proposals significantly enhanced both in-domain performance on BDD and zero-shot tracking on TAO. This underscores the importance of SAM's dense, diverse object proposals for superior contrastive similarity learning.

\noindent\textbf{Effect of Proposal Quantity}
Investigating the impact of SAM's proposal quantity on learning, we experimented with different upper bounds of 64, 128, and 256 proposals per batch. Table~\ref{tab:ablation}b shows consistent improvements in AssocA on BDD and TAO with increasing proposal numbers, indicating that a rich collection of instances fosters more discriminative tracking features.

\noindent\textbf{Effect of Data Augmentations} As shown in Table~\ref{tab:ablation}c, the combination of random affine, Mixup~\cite{mixup} and LSJ~\cite{copy-paste} gives the best performance. Method 1 represents basic data augmentation including flipping, resizing, color jitter and random cropping. If there is no strong augmentation (method 1), its mIDF1 on BDD MOT drops by 6.7\%, being much worse than that with method 5. 
These results illustrate the necessity of strong augmentations in training only on static images.

\noindent\textbf{Qualitative Results}
In Figure~\ref{fig-qualitative}, we present the qualitative results of our unified methods, Grounding-DINO and SAM-H. Our methods accurately detect, segment, and track multiple objects and even their parts across diverse domains. This includes animated movie scenes featuring many similar-looking characters and driving scenes within complex environments.

\begin{figure}[!t]
  \centering%
\includegraphics[width=1.0\linewidth]{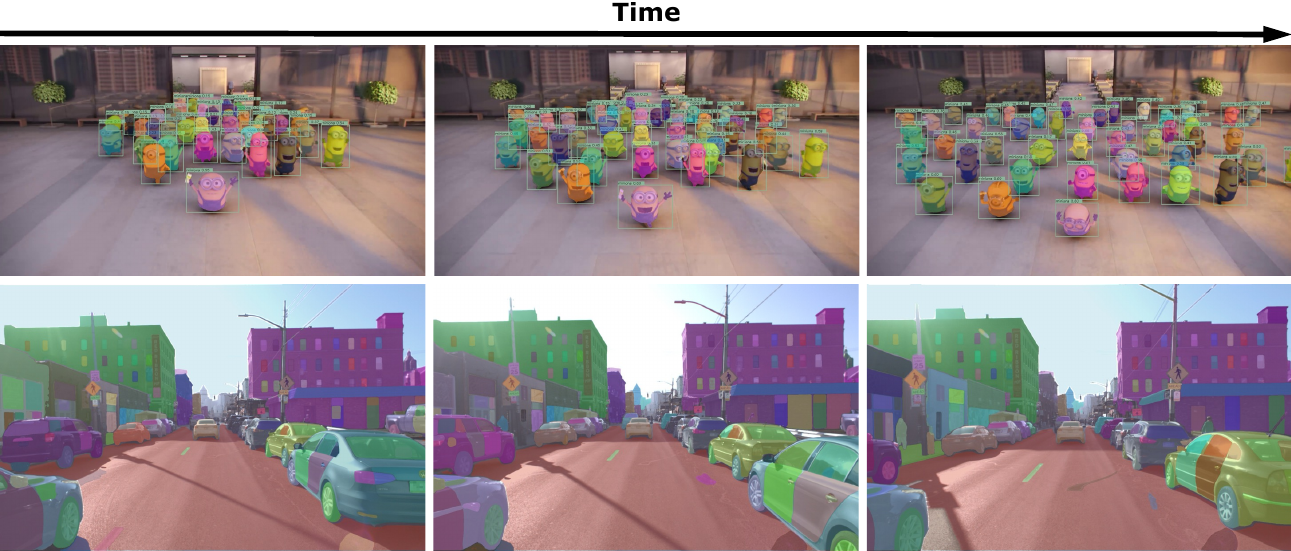}%
  \caption{Qualitative results of our unified models using Ours-Grounding-DINO (top) and Ours-SAM-H (bottom). We use SAM-H to generate masks given the detected boxes.}%
  \label{fig-qualitative}%
\end{figure}

\section{Conclusion}
\label{sec:conclusion}
We present \ourmodel, a novel method that exploits the extensive instance-level shape and appearance information from SAM to learn generalizable instance associations from unlabeled images. \ourmodel demonstrates exceptional zero-shot association performance across various benchmarks, eliminating the need for expensive domain-specific labels. Moreover, our universal MASA adapter can be added to any existing detection and segmentation models, enabling them to efficiently track any objects across diverse domains. 

\section{Acknowledgments}
We would like to express our gratitude to Ruolan Xiang for her help with editing, and to Bin Yan for both editing and the valuable discussions.

{
\small
\bibliographystyle{ieeenat_fullname}
\bibliography{main}
}
\newpage
% \newpage
\renewcommand{\thepart}{}
\renewcommand{\partname}{}
\counterwithin*{section}{part}

\part{Appendix}
\renewcommand{\thesection}{\Alph{section}}
In this supplementary material, we provide additional ablation studies and qualitative results of our fast proposal generation and of our association. We also elaborate on our experimental setup, method details, and training and inference hyper-parameters. 

\noindent The supplementary material is structured as follows:
\begin{itemize}
    \item Section~\ref{sec:supp-backbones}: Effectiveness on other backbones.
    \item Section~\ref{sec:supp-youtubeVIS}: Zero-shot evaluation on YoutubeVIS.
    \item Section~\ref{sec:supp-vis-instance-embeddings}: Visualization of instance embeddings.
    \item Section~\ref{sec:supp-domain-gap}: Domain gap and adaptation.
    \item Section~\ref{sec:supp-pho-aug}: Impact of additional photometric augmentation.
    \item Section~\ref{sec:supp-prop-diverse}: More detailed comparison of proposal diversity.
    \item Section~\ref{sec:supp-ssl-compare}: Compare with self-supervised methods.
    \item Section~\ref{sec:supp-vos}: Comparison with VOS-based methods.
    \item Section~\ref{sec:qua-res}: More qualitative results.
    \item Section~\ref{sec:supp-implementation_detail}: More implementation details.
    \item Section~\ref{sec:supp-limit}: Limitations.

\end{itemize}

\section{Effectiveness on Other Backbones}
\label{sec:supp-backbones}
In our main paper, we introduced four method variants, each building upon foundational detection and segmentation models: SAM-ViT-B, SAM-ViT-H, Grounding-DINO, and Detic. Notably, the latter two variants leverage the Swin-B backbone. Our MASA training pipeline and adapter have shown great adaptability to a range of variables, including variations in backbone structures, pre-training methods (such as detection or segmentation), and the diverse datasets employed in training these foundational models.

A critical observation from our study is the reliance of these variants on large, complex backbones and their pre-training on extensive datasets. This reliance poses an important question about scalability and efficiency: Can our method sustain its effectiveness when applied to smaller, more streamlined backbones, like the ResNet-50, especially with standard ImageNet pre-training? To explore this, we devised a new variant, "\textbf{Ours-R50}," which integrates the ResNet-50 backbone pre-trained on the ImageNet classification task (IN-Sup R50). This new variant maintains the MASA adapter architecture from our main research, adhering to the identical training protocol established by our initial four variants.

We have assessed the performance of Ours-R50 across various benchmarks, including BDD MOTS, BDD MOT, and TAO TETA. The quantitative results, detailed in Tables~\ref{tab:mots_sup},~\ref{tab:bdd_sup}, and~\ref{tab:tao_sup}, demonstrate the efficacy of Ours-R50. These findings are significant as they suggest that our approach can be effectively adapted to smaller backbones, offering the potential for more efficient and scalable solutions in detection and segmentation tasks.

\noindent\textbf{BDD MOTS}: For BDD MOTS (Table~\ref{tab:mots_sup}), Ours-R50, equipped with the ResNet-50 backbone and our MASA training approach, not only outperforms the UNINEXT-H model with a +0.2 mIDF1 and +0.4 AssocA but also shows minimal performance drop compared to the strongest variant, Ours-SAM-H (-1 mIDF1 and -0.9 AssocA). This highlights our method's ability to yield competitive instance embeddings, even without the advanced features provided by larger, specialized models.

\noindent\textbf{BDD MOT}: In the BDD MOT benchmark (Table~\ref{tab:bdd_sup}), Ours-R50 surpasses ByteTrack in terms of IDF1 (+0.9) and AssocA (+0.2) scores. Its performance is on par with our other variants, showing only a slight decrease compared to Ours-Detic (-1 mIDF1 and -1.2 AssocA). These results reaffirm the adaptability of our method across various backbone architectures and pre-training environments.

\noindent\textbf{TAO TETA}: Evaluating on TAO TETA (Table~\ref{tab:tao_sup}), Ours-R50, with its standard ResNet-50 backbone, continues to perform robustly. It closely matches the fully supervised TETer model, with only a slight decrease in AssocA (-1). This performance, consistent with our other variants, further validates the generalizability of our MASA approach across different backbones and pre-training methodologies.

\begin{table}[t]
    \caption{State-of-the-art comparison on BDD MOTS benchmark. All methods in the table use the same object detection observations. AssocA, mIDF1, and IDF1 mainly focus on the association quality.}
    \centering
    {\resizebox{1\linewidth}{!}{
    \begin{tabular}{lccccc}
        \toprule
        Method & mIDF1$\uparrow$ & AssocA$\uparrow$ & TETA$\uparrow$ & mMOTSA$\uparrow$ & mHOTA$\uparrow$ \\
        \midrule
        \underline{\textit{Fully-supervised, in-domain}} & & & & &  \\
        UNINEXT-H~\cite{UNINEXT}$^{\dagger}$ & 48.5 & 53.2 & {53.6} & {35.7} & {40.6} \\
        \midrule
        \underline{\textit{Self-supervised, zero-shot}} & & & & &  \\
        \textbf{Ours-Detic}$^{\dagger}$ & \ranksecond{49.5} & {53.5} & {54.4} & \rankfirst{36.4} & 40.2 \\
        \textbf{Ours-Grounding-DINO}$^{\dagger}$ & 48.6 & {52.3} & {54.0} & \ranksecond{36.1} & {40.0} \\
        \textbf{Ours-SAM-B}$^{\dagger}$ & {49.2} & \ranksecond{53.9} & \rankfirst{54.8} & 35.2 & \ranksecond{40.7} \\
        \textbf{Ours-SAM-H}$^{\dagger}$ & \rankfirst{49.7} & \rankfirst{54.5} & \ranksecond{54.7} & {35.8} & \rankfirst{40.8} \\ \midrule
        \textbf{Ours-R50}$^{\dagger}$ & {48.7} & {53.6} & \ranksecond{54.7} & {35.2} & {40.4} \\
        \bottomrule
    \end{tabular}
}}
    \label{tab:mots_sup}
    \vspace{6mm}
\end{table}

\begin{table}[t]
    \caption{State-of-the-art comparison on BDD MOT benchmark. All methods in the table use the same object detection observations. Our training method learns the most robust and accurate association.}
    \centering
    {\resizebox{1\linewidth}{!}{
    \begin{tabular}{lcccccc}
        \toprule
        Method & mIDF1$\uparrow$ & IDF1$\uparrow$ & TETA$\uparrow$ & AssocA$\uparrow$ & mMOTA$\uparrow$ \\
        \midrule
        \underline{\textit{Fully-supervised, in-domain}} & & & & &  \\
        ByteTrack~\cite{bytetrack}$^{\dagger}$ & 54.8 & 70.4 & \rankfirst{55.7} & 51.5 & \rankfirst{45.5} \\
        \midrule
        \underline{\textit{Self-supervised, zero-shot}} & & & & &  \\
        \textbf{Ours-Detic}$^{\dagger}$ & \rankfirst{55.8} & {71.3} & 54.4 & \rankfirst{52.9} & \ranksecond{44.6} \\
        \textbf{Ours-Grounding-DINO}$^{\dagger}$ & \ranksecond{55.6} & \rankfirst{71.7} & \ranksecond{54.5} & \ranksecond{52.7} & {44.5} \\
        \textbf{Ours-SAM-B}$^{\dagger}$ & {55.6} & \ranksecond{71.6} & 54.0 & {52.6} & 44.1 \\
        \textbf{Ours-SAM-H}$^{\dagger}$ & {55.3} & \rankfirst{71.7} & {54.2} & {51.9} & {44.5} \\ \midrule
        \textbf{Ours-R50}$^{\dagger}$ & {54.8} & {71.3} & {54.0} & {51.7} & {44.2} \\
        \bottomrule
    \end{tabular}
}
}
    \label{tab:bdd_sup}
    % \vspace{-5mm}
\end{table}

\begin{table}[t]
    \caption{State-of-the-art comparison on TAO TETA benchmark. All methods in the table use the same object detection observations.}%
    \centering%
    \resizebox{0.9\linewidth}{!}{\resizebox{\linewidth}{!}{
\begin{tabular}{lccccccc}
    \specialrule{.1em}{.05em}{.05em} 
    Method & AssocA & TETA & LocA & ClsA \\
    \hline%\hline
    \multicolumn{5}{l}{\underline{\textit{Fully-supervised, in-domain}}}  \\
    TETer~\cite{TETer}$^{\dagger}$ & \ranksecond{36.7} & {34.6} & \rankfirst{52.1} & 15.0 \\
    \midrule
    \multicolumn{5}{l}{\underline{\textit{Self-supervised, zero-shot}}}  \\
    \textbf{Ours-Detic}$^{\dagger}$ & 36.4 & \ranksecond{34.7} & \ranksecond{51.9} & \rankfirst{15.8} \\
    \textbf{Ours-Grounding-DINO}$^{\dagger}$ & \rankfirst{37.6} & \rankfirst{34.9} & 51.8 & \ranksecond{15.4} \\
    \textbf{Ours-SAM-B}$^{\dagger}$ & 36.6 & 34.5 & \ranksecond{51.9} & 15.1 \\
    \textbf{Ours-SAM-H}$^{\dagger}$ & 36.4 & 34.5 & 51.8 & \ranksecond{15.4} \\ \midrule
    \textbf{Ours-R50}$^{\dagger}$ & 35.7 & 34.1 & \textbf{52.1 }& 15.0 \\ 
    \specialrule{.1em}{.05em}{.05em} 
\end{tabular}
}}%
    \label{tab:tao_sup}%
    \vspace{6mm}%
\end{table}

\section{Zero-shot Evaluation on YoutubeVIS}
\label{sec:supp-youtubeVIS}

In this section, we evaluate our association method in a zero-shot setting on the Youtube-VIS 2019~\cite{VIS} benchmark. To be specific, we test our MASA adapter with SAM-ViT-B as the base model directly on Youtube-VIS 2019 for association. Our method uses the same object detection observations as the state-of-the-art VIS method UNINEXT-R50~\cite{UNINEXT}. As shown in Table~\ref{tab:vis}, our method achieves comparable performance with SOTA UNINEXT trained with the in-domain YoutubeVIS data, while outperforming all other approaches significantly. This outcome underlines the robust zero-shot association capabilities of our method, highlighting its effectiveness in scenarios without domain-specific training.

\begin{table}[t]
    \caption{State-of-the-art comparison on Youtube-VIS 2019. $^{\dagger}$ represents that we provide the same object detection observations. Our method does not train using any image or any annotation from Youtube-VIS 2019.}%
    \centering
    \resizebox{1\linewidth}{!}{\resizebox{1.0\columnwidth}{!}{
\begin{tabular}{lcccccc}
\toprule
% \noalign{\smallskip}
\multirow{2}{*}{Method}   & \multirow{2}{*}{Zero-shot} & \multirow{2}{*}{Association Label} &\multirow{2}{*}{Video} &\multicolumn{3}{c}{VIS2019 val}\\
% \cmidrule(lr){4-6} \cmidrule(lr){7-9}
\arrayrulecolor{white}
\arrayrulecolor{black}
\arrayrulecolor{black}
\arrayrulecolor{black}
\arrayrulecolor{white}
&&&&$\rm AP$  &$\rm AP_{75}$ \\
% \noalign{\smallskip}
\arrayrulecolor{white}\hline
\arrayrulecolor{black}\hline
\arrayrulecolor{white}\hline
% \noalign{\smallskip}

 VisTR~\cite{VISTR}   & \ding{55} & \ding{51}      &\ding{51} &36.2 &36.9\\
 MaskProp~\cite{MaskProp}  &  \ding{55} & \ding{51}  &\ding{51} &40.0 &42.9\\  
 IFC~\cite{IFC}  &  \ding{55} & \ding{51} &\ding{51} &42.8 &46.8\\  
{SeqFormer}~\cite{SeqFormer}   &  \ding{55} & \ding{51}  &\ding{51} &47.4 &51.8\\
 IDOL~\cite{IDOL}  & \ding{55} & \ding{51} &\ding{51} & 49.5 &52.9\\
 MFVIS~\cite{maskfreevis}  & \ding{55} & \ding{51} &\ding{51} & 46.6 &49.7\\
 VITA~\cite{VITA}&\ding{55} & \ding{51}&\ding{51}&49.8&54.5\\
UNINEXT-R50~\cite{UNINEXT}$^{\dagger}$  & \ding{55} & \ding{51} &\ding{51} &\rankfirst{53.0}&\rankfirst{59.1}\\ 
\textbf{Ours-SAM-B}$^{\dagger}$ & \ding{51} & \ding{55} & \ding{55} & \ranksecond{51.8} & \ranksecond{58.1} \\
\arrayrulecolor{white}\hline
\arrayrulecolor{black}\hline
\arrayrulecolor{white}\bottomrule
\end{tabular}
}}%
    \label{tab:vis}%
\end{table}

\section{Visualization of Instance Embeddings} 
\label{sec:supp-vis-instance-embeddings}
\begin{figure*}[!t]
  \centering%
\includegraphics[width=0.8\linewidth]{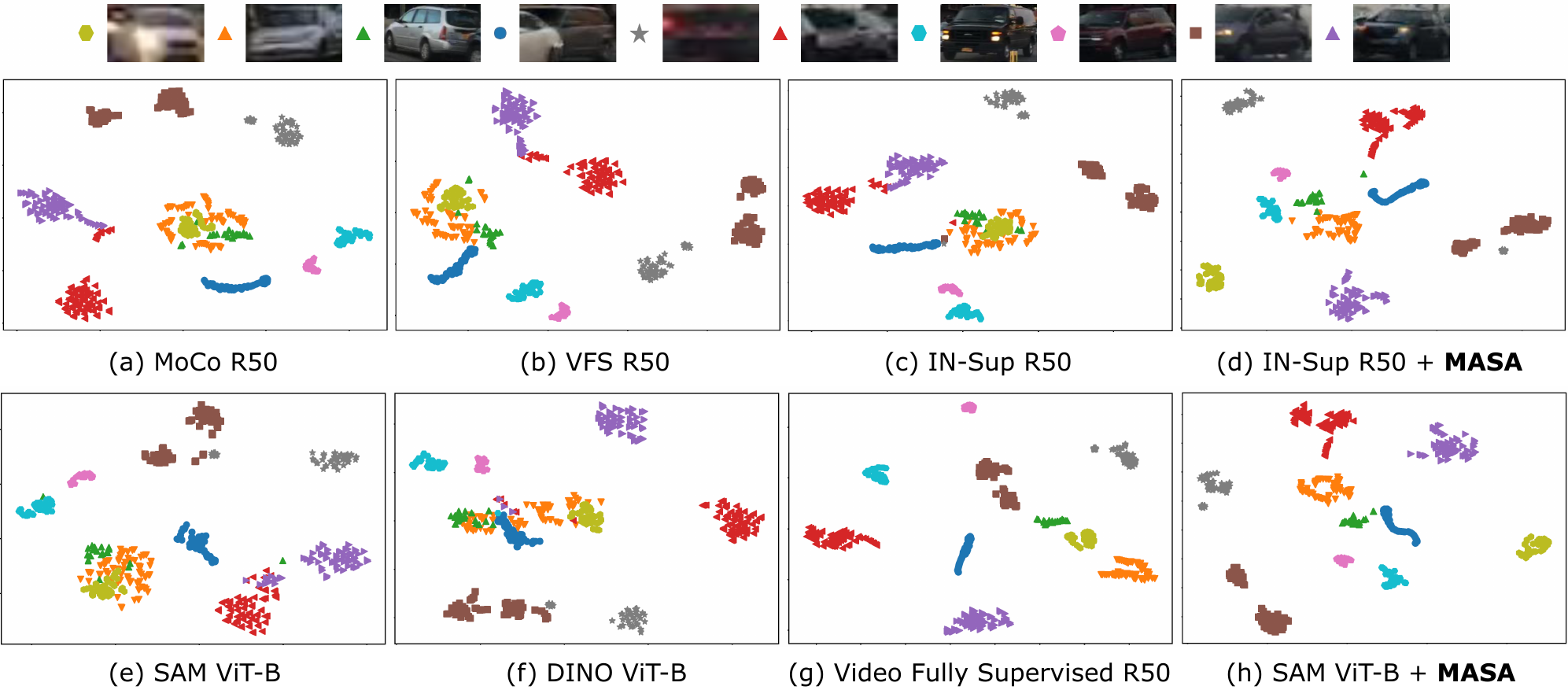}%
  \caption{t-SNE visualization demonstrates the distinctiveness of instance embeddings across various methods on a selected BDD100k sequence. The embeddings generated by our method (indicated by MASA-enhanced models) exhibit greater inter-instance separation and tighter intra-instance clustering than other self-supervised methods (MoCo, VFS, DINO) and the original supervised methods (IN-Sup, SAM). This enhanced discrimination highlights the effectiveness of our adapted features for downstream tasks.}%
  \label{fig-visual-embeddings-supp}
  % \vspace{-4mm}%
\end{figure*}

In Figure~\ref{fig-visual-embeddings-supp}, we use t-SNE to visualize instance embeddings learned in different ways. We compare self-supervised approaches such as MoCo-v2~\cite{MoCov2}, VFS~\cite{VFS}, and DINO~\cite{DINO-SSL}, alongside two base models: SAM ViT-B~\cite{SAM}, originally pre-trained on SA-1B for segmentation tasks, and IN-Sup R50~\cite{ResNet}, initially pre-trained on ImageNet for image classification. Additionally, we present embeddings from fully supervised in-domain video models~\cite{TETer} and the same base models enhanced with our MASA adapters.  In these visualizations, instances that share the same ground-truth ID are represented in the same colors. We use the BDD100K sequence as the data source. 

Our observations indicate that the embeddings from the original SAM, IN-Sup R50, as well as the self-supervised methods like MoCo, VFS, and DINO, do not consistently separate different instances within certain complex scenarios, as highlighted by the instances marked in green, orange, and yellow. In contrast,  by applying our MASA adapter to the original SAM ViT-B and IN-Sup R50 features, the resulting adapted embeddings exhibit a successful delineation of distinct instances. This performance is comparable to that of fully supervised methods that have been trained on labeled in-domain videos. Significantly, our method achieves these results without any labeled in-domain video data, demonstrating its considerable potential for robust instance-level correspondence learning.

\section{Domain Gap and Adaptation}
\label{sec:supp-domain-gap}
\begin{figure}[!t]
  \centering%
\includegraphics[width=0.95\linewidth]{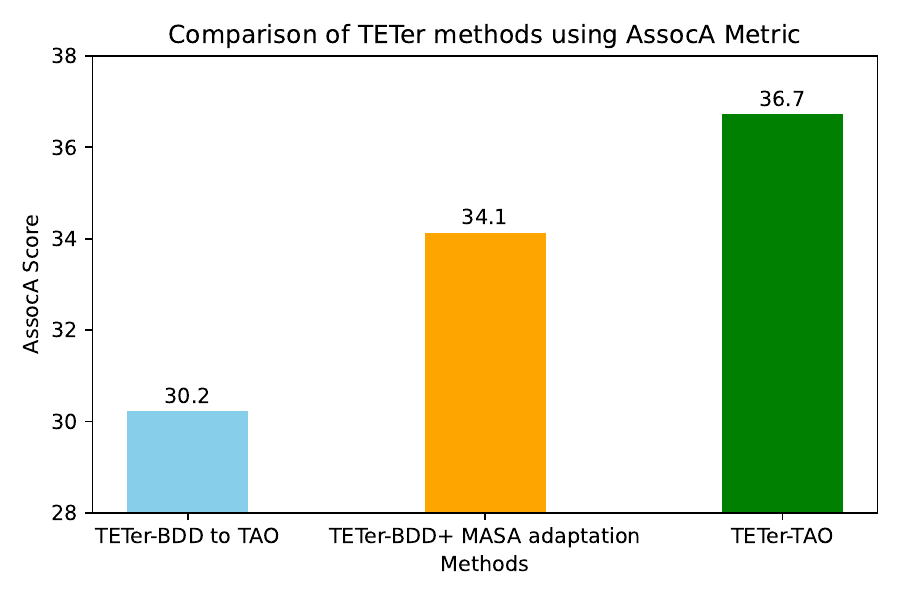}\vspace{-2mm}%
  \caption{Domain adaptation for TETer with MASA.}%
  \label{fig-domain_adapataion}
\end{figure}

Except for previously mentioned applications, MASA can also serve as a useful domain adaption method for instance association. To be specific, due to the domain gaps such as object categories, scenarios, and lighting conditions, trackers trained on data of domain A may suffer from performance drop when evaluating on domain B. For example, compared with BDD~\cite{BDD100K}, TAO~\cite{TAO} covers much more diverse scenarios and object categories. Thus, we choose BDD100K~\cite{BDD100K} as the source domain and TAO~\cite{TAO} as the target domain. Then we train two separate models with the same architecture as TETer~\cite{TETer} using labeled data of BDD and LVIS+ TAO~\cite{LVIS, TAO} respectively. These two models are represented by the blue and green bars in Figure~\ref{fig-domain_adapataion}. Please note that when evaluating their associating ability on TAO, they use the same object detection observations. As shown in Figure~\ref{fig-domain_adapataion}, directly applying embeddings trained on BDD to TAO (blue bar) leads to poor AssocA, which is 6.5\% lower than the model trained on in-domain TAO (green bar). To alleviate this performance gap, we fine-tune the track head of the original TETer model represented by the blue bar with the MASA training pipeline, while freezing all other parameters (orange bar). Specifically, we only fine-tune the model using unlabeled images of LVIS and TAO, while not using any original TAO annotation. As shown in Figure~\ref{fig-domain_adapataion}, compared with the blue bar, the orange bar achieves an improvement of 3.9\% on AssocA, reducing the domain gap by 60\%. This demonstrates that MASA can effectively improve the association performance in out-of-domain scenarios, only requiring unlabeled images from the target domain. 

\section{Impact of Photometric Augmentation}
\label{sec:supp-pho-aug}
\begin{figure*}[!t]
\centering
\includegraphics[width=0.95\linewidth]{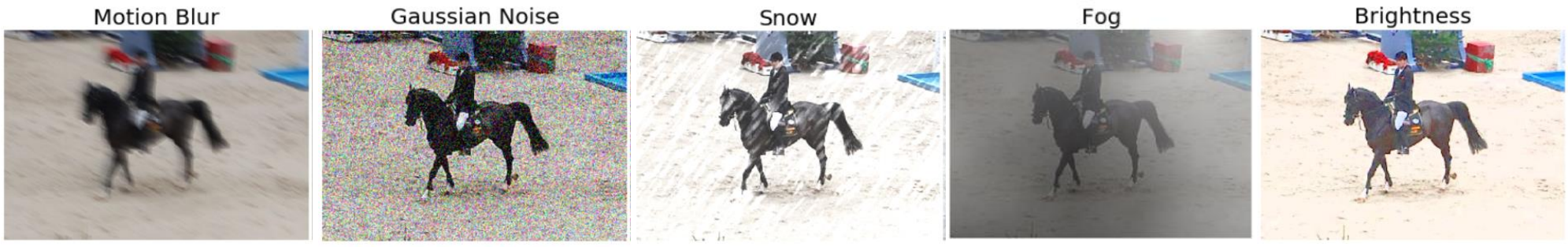}
\caption{Beyond the strong geometric augmentations utilized in the main study, this figure presents an exploration of five additional photometric augmentations: motion blur, Gaussian noise, snow, fog, and brightness adjustments.}
\label{fig-photometric-aug}
\end{figure*}

In Section 4.3 of our main paper, we focused on various geometric augmentations, including random affine transformations, and large-scale jittering. We also use MixUp to enhance the instance diversity and simulate the occlusion effect. This section delves into the impact of additional photometric augmentation. We specifically examine the effects of motion blur, Gaussian noise, snow, fog, and brightness adjustments. Photometric augmentations are characterized by their ability to modify pixel values in an image. These alterations often mimic changes in environmental factors such as lighting and weather, impacting how scenes are captured by cameras. Unlike geometric augmentations that change the spatial arrangement of pixels through rotation, scaling, or cropping, photometric augmentations do not alter the structural integrity of objects within an image. Figure~\ref{fig-photometric-aug} illustrates these augmentations visually.

We maintained the same training regimen as our ablation study in the main paper, and using SAM-ViT-B as the foundational model for our experiments. Table ~\ref{tab-supp-photometric-aug} presents the results, indicating that the inclusion of photometric augmentation yields only modest improvements. We observed a marginal increase of +0.1 mIDF1 and +0.2 AssocA on the BDD dataset and +0.1 AssocA on the TAO dataset. Consequently, these augmentations are not included as a default in our methodology to achieve a better balance between performance improvement and the potential increase in computational complexity.

\begin{table}[t]
\caption{Assessing the Impact of Additional Photometric Augmentation. The standard augmentation set includes flipping, color jittering, and random cropping. The more intensive "Strong Aug" set comprises random affine transformations, large-scale jittering, and mix-up techniques. The photometric augmentation set tested here includes motion blur, Gaussian noise, snow, fog, and brightness adjustments.}
\centering
\resizebox{1\linewidth}{!}{\resizebox{1\linewidth}{!}{
        \begin{tabular}{c|c|c|l|l|l}
            \toprule
              \multirow{2}{*}{Standard Aug}  & \multirow{2}{*}{Strong Aug}  & \multirow{2}{*}{Photometric Aug} &\multicolumn{2}{c|}{BDD MOT} & TAO \\
        
             & & &  mIDF1& AssocA & AssocA \\ \midrule
             \ding{51}&  & &  48.2 & 43.9   &   28.5 \\
             \ding{51}&\ding{51}& & 54.9& 51.9 & 35.8\\
             \ding{51}&\ding{51}&\ding{51}& 55 & 52.1 &35.9\\
            \bottomrule
        \end{tabular}
    }

}
\label{tab-supp-photometric-aug}
\end{table}

\section{Comparison of Proposal Diversity}
\label{sec:supp-prop-diverse}
In our main paper, we assessed different proposal generation mechanisms within the context of association learning. Specifically, we focused on training using raw images from the BDD dataset. We experimented by replacing SAM in our MASA pipeline with Mask2former-SwinL, pre-trained on the COCO dataset (see \cite{Mask2Former}). As detailed in Table 9c of the main paper, the model utilizing SAM's proposals demonstrated enhanced performance. This was evident both in in-domain tracking on the BDD dataset and in zero-shot tracking scenarios on the TAO dataset. Such findings highlight the crucial role of SAM's dense and diverse object proposals in facilitating effective contrastive similarity learning.

\begin{figure*}[!t]
\centering
\includegraphics[width=0.95\linewidth]{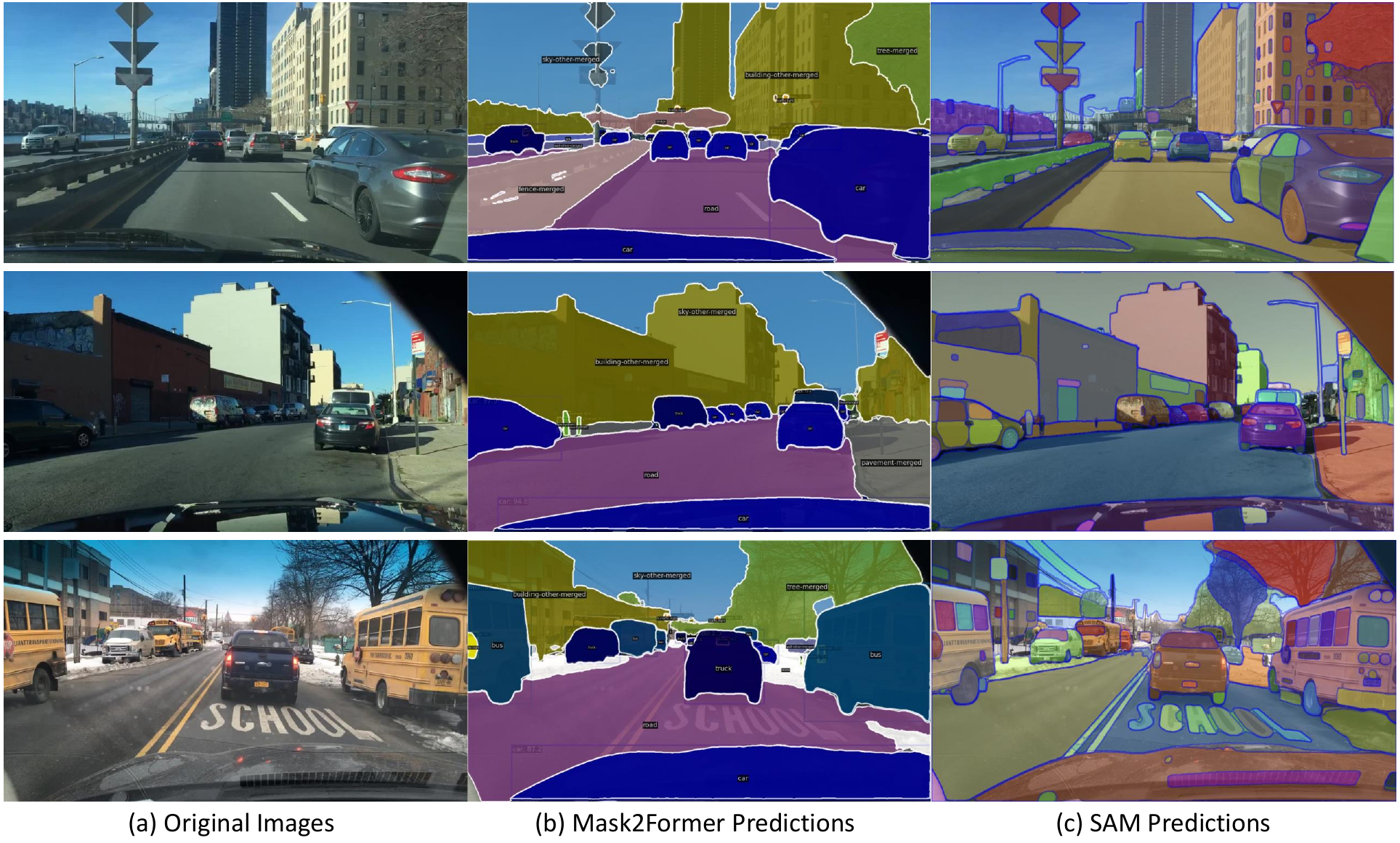}
\caption{Comparison between predictions of Mask2Former and SAM.While Mask2Former is limited to identifying 'things' and 'stuff' from categories included in its training set, SAM demonstrates a broader detection scope. It effectively identifies objects of more diverse classes and finer granularity, such as windows, wheels, and traffic signs.}

\label{fig-mask2former-compare}
\end{figure*}

Further, we present visual comparisons of the proposals generated by Mask2former and SAM in Figure~\ref{fig-mask2former-compare}. These comparative visualizations distinctly showcase the superior diversity in SAM's proposals relative to those generated by Mask2former. SAM exhibits an enhanced ability to identify a wider array of instances within raw images, providing proposals with greater diversity. This diversity is pivotal in instance similarity learning and significantly contributes to the out-of-domain generalization capabilities of the learned instance representations.

\section{Compare with Self-Supervised Methods}
\label{sec:supp-ssl-compare}
\begin{figure*}[!t]
\centering
\includegraphics[width=0.9\linewidth]{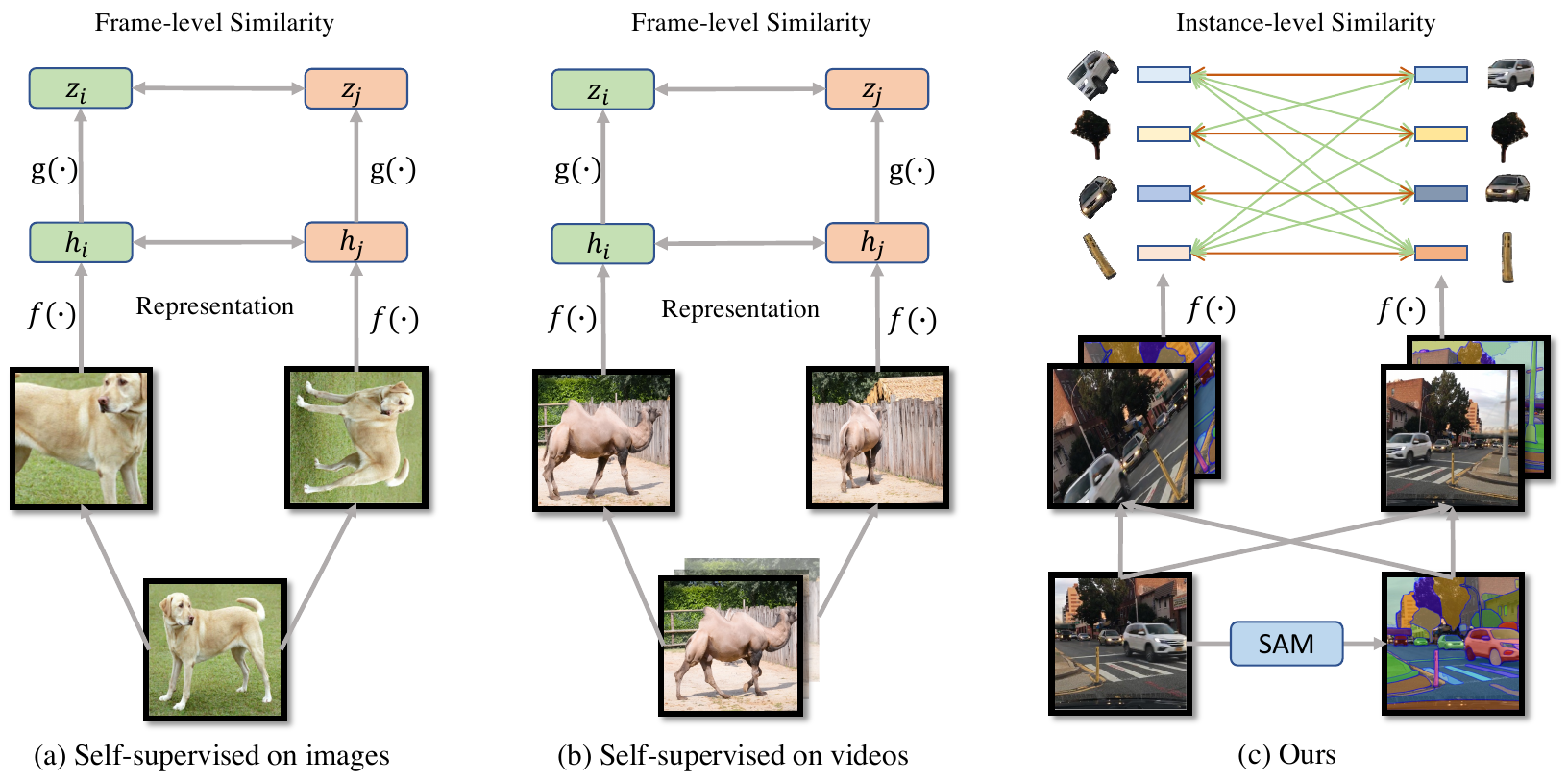}
\caption{Comparison of self-supervised representation learning methods for object association. (a) Traditional methods, such as SimCLR~\cite{SimCLR}, MoCo~\cite{MoCov2}, focus on learning representations by leveraging frame-level similarity. They utilize augmented views of entire images to extract meaningful features. These methods often struggle with complex scenarios involving multiple objects. The reliance on frame-level similarity can be limiting in environments where object-centric learning is crucial. (b) Methods like VFS~\cite{VFS} take a different route by extracting positive pairs from different frames within the same video. This approach aims to capture temporal consistency and object dynamics. Similar to traditional methods, it also requires clean, object-centred video data. The complexity increases significantly in multi-object environments, where distinguishing between different objects becomes challenging.  (c) Our method innovatively combines data augmentation with SAM's~\cite{SAM} mask generation technique. This synergy allows for learning dense instance-level correspondences from unlabeled images. By focusing on dense correspondences at the instance level, it can effectively disentangle and learn from intricate object interactions and dynamics in complex environments.
}
\label{fig-self-supervised-compare}
\end{figure*}

The task of extracting meaningful information from purely unlabeled images is notably challenging. UniTrack \cite{uniTrack} has showcased the potential of self-supervised trained representations, such as MoCo \cite{MoCov2} and VFS \cite{VFS}, in generalizing to various tracking tasks across different domains. However, as depicted in Figure~\ref{fig-self-supervised-compare}, current self-supervised methods predominantly employ contrastive training with clean, object-centered images or videos. In particular, VFS trains on the Kinetics dataset, while MoCo and DINO utilize ImageNet. 

However, these approaches primarily focus on frame-level similarities and fail to leverage instance information effectively. Consequently, they struggle to learn accurate instance representations in complex domains with multiple instances appearing together, demonstrating a notable weakness in extracting robust and generalized representations.

\begin{figure*}[!t]
\centering
\includegraphics[width=0.95\linewidth]{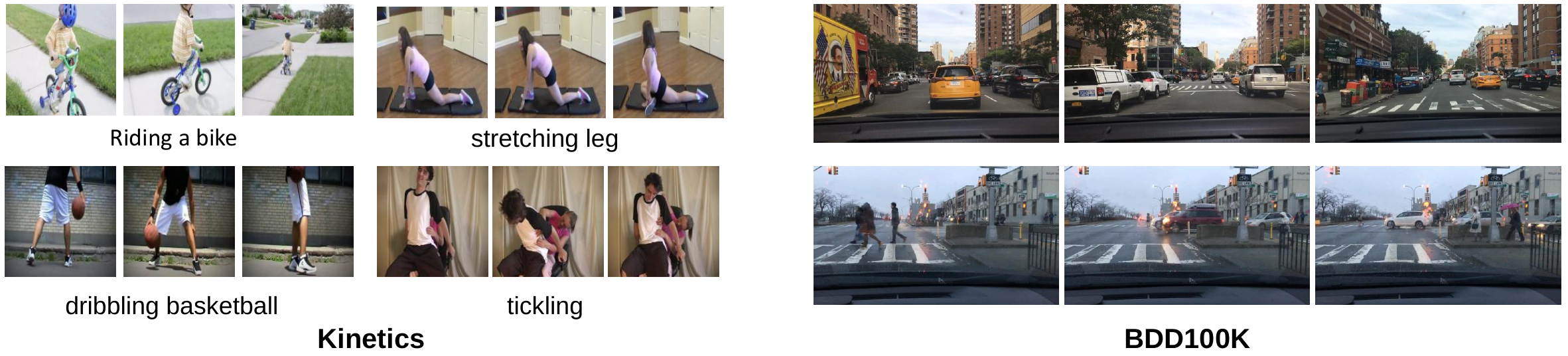}
\caption{Comparison between Kinetics and BDD100K videos. Kinetics, as an action recognition dataset, ensures that actions are contained within selected videos, thus guaranteeing the presence of instances throughout the video. Centred entities in Kinetics videos usually do not change over time. This makes VFS's sampling strategy reasonable for Kinetics. 
  However, in BDD videos, objects get into and out of the frames frequently. Two frames sampled from the same video may not contain the same instances at all.}
\label{fig-kinetics}
\end{figure*}

\noindent\textbf{Visualization of Object-Centered Training Data} We visualize the training data of VFS~\cite{VFS}, the Kinetics dataset, and compare it with the driving videos from BDD100K in \figureautorefname~\ref{fig-kinetics}. Kinetics, being an action recognition dataset, ensures the presence of instances throughout its videos by focusing on contained actions. Centred entities in Kinetics videos usually remain consistent over time, making VFS's sampling strategy suitable for Kinetics. In contrast, BDD100K driving videos present a more dynamic and unpredictable environment. These videos frequently feature objects that enter and exit the frame, leading to a significant variation in the presence of instances across different frames. This characteristic of BDD100K poses a challenge as two frames sampled from the same video may not share the same instances, highlighting a fundamental difference in the nature of training data between the two datasets.

\noindent\textbf{Training with Different Data Sources} For a fair comparison, when comparing our method with other self-supervised counterparts in Table 6 of the main paper, we train all methods using the same raw training images (BDD and COCO), which are not object-centered and usually contain multiple instances in complex environments. In this section, we also present the tracking performance of those self-supervised methods using their original object-centered training data. As shown in the table below, the AssocA of MoCo trained on images from BDD and COCO remains relatively stable compared to its original version trained on ImageNet, with only a slight drop on the BDD MOT dataset. However, for VFS, training on images with multiple instances leads to a significant performance drop of 15.9 AssocA on BDD MOT and 12.7 AssocA on TAO, respectively. The reason is as follows: VFS considers frames from the same video as positive samples and frames from different videos as negative samples. This strategy is reasonable for Kinetics but not for BDD, as demonstrated in \figureautorefname~\ref{fig-kinetics}. Specifically, centred entities in Kinetics videos usually do not change over time, but in BDD videos, objects frequently move in and out of frames. Two frames from the same BDD video may not contain the same instances at all. Lastly, in DINO's training process, it forces representations of two augmented views from the same image to be similar without explicitly using negative samples. However, for images in BDD and COCO, two augmented views may contain many different instances, considering the complex scenes of these two datasets. This training strategy may cause the learned embeddings to be less discriminative. 

Our approach, which leverages instance-level knowledge from the pre-trained SAM, moves beyond frame-level similarity to embrace a more nuanced instance-level similarity. The strong results obtained underscore the effectiveness of our proposed methods in learning robust representations for tracking purposes.

\begin{table}[t]
    \caption{Compare with self-supervised based methods. All methods use
the same detection observations for testing. Object-centred data means ImageNet for MoCO and DINO, and Kinetics for VFS.}%
    \centering
    \resizebox{1\linewidth}{!}{\begin{tabular}{@{}lc|cc|c|c|c@{}}
\toprule
\multirow{2}{*}{Method} &\multirow{2}{*}{Video}& \multicolumn{2}{c|}{BDD MOT} & \multicolumn{1}{c|}{TAO} & \multicolumn{2}{c}{BDD MOTS} \\
& & AssocA &mIDF1 & AssocA & AssocA &mIDF1 \\
\midrule
\underline{\textit{Train on object-centred data}}&&&&&& \\
VFS & \ding{51} &45.1&49.9&31.8 & 50.3 & 44.9 \\
MoCov2 &\ding{55} &44.1 &48.6&31.1 & 50.5 & 45.6 \\
DINO &\ding{55}  &41.7 & 46.5&26 & 46 & 40.5 \\
\midrule
\underline{\textit{Train on BDD \& COCO}} &&&&&&\\
VFS & \ding{51} &29.2&35.0&19.1 & 30.7 & 30.1 \\
MoCov2 &\ding{55} &42.7 &46.7&30.7 & 51 & 45.3  \\
DINO &\ding{55}  &23.1 & 16.8&12.9 & 20.2 & 22.2 \\
\textbf{Ours-SAM-B} &\ding{55} &\textbf{51.9}&\textbf{54.9}&\textbf{35.8} & \textbf{53.7}&\textbf{49.1} \\
\bottomrule
\end{tabular}
}%
    \label{tab-supp-self-sup}%
\end{table}

\section{Comparison with VOS-based Methods}
\label{sec:supp-vos}
The recent segmentation foundation model, SAM, has demonstrated exceptional ability in segmenting any object. However, simultaneously tracking all instances generated by SAM in videos remains a challenging task. Current methods typically employ SAM as a mask generator for the first frame of a video, then apply off-the-shelf video object segmentation (VOS) methods to propagate the initialized mask to subsequent frames~\cite{deva, SAMTrack, DeAOT, XMem}. One notable method, Deva~\cite{deva}, utilizes XMem~\cite{XMem} for mask propagation to track multiple instances simultaneously. However, these methods encounter several key disadvantages.

\begin{figure*}[t]
\centering
\includegraphics[width=0.8\linewidth]{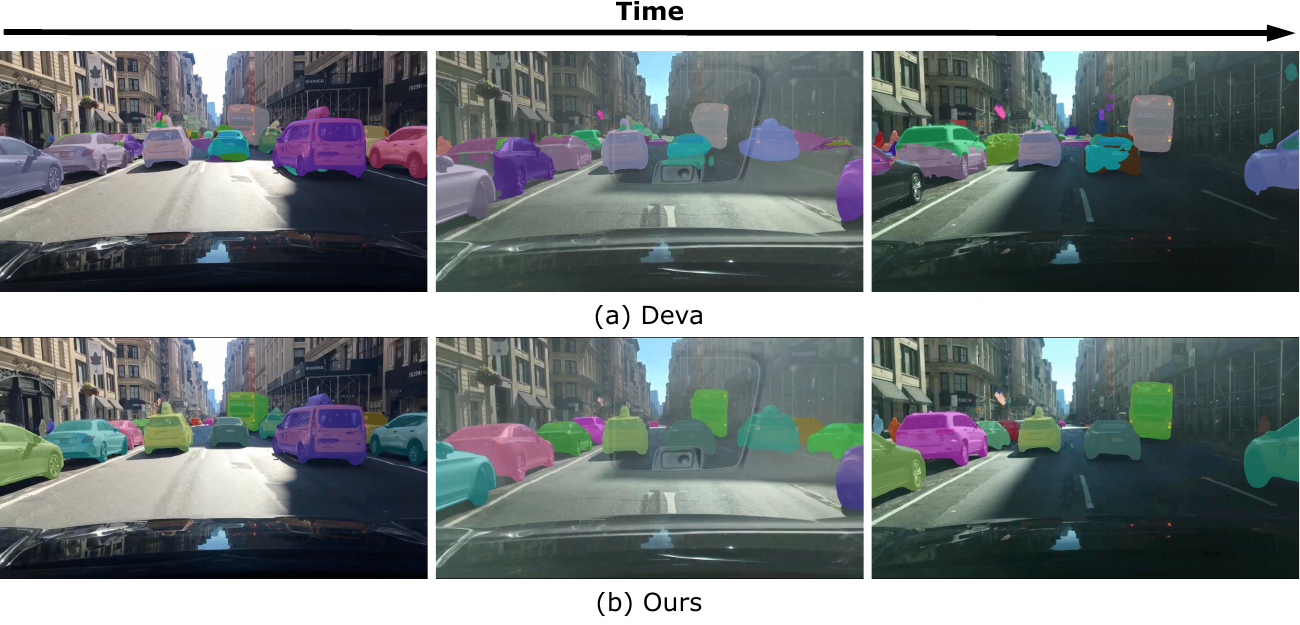}
\caption{Qualitative Comparison between Our Method and Deva~\cite{deva} on BDD100K. This figure illustrates the challenges Deva faces in driving scenarios, a domain beyond its training environment. Key issues include inadequate mask propagation and an increasing incidence of false positives over time. }
\label{fig-deva-ours-bdd}
\end{figure*}

\noindent\textbf{Inadequate Mask Propagation Quality:} Trained on relatively small-scale video segmentation datasets, these methods experience substantial domain gaps when tasked with tracking any object in any domain, resulting in inadequate mask propagation quality. Our main paper illustrates that our method significantly outperforms Deva~\cite{deva} in zero-shot testing across various multiple object tracking benchmarks, especially in driving scenes, which are out-of-domain for both Deva and our method. We further provide a qualitative comparison in Figure~\ref{fig-deva-ours-bdd}. Additional video comparisons can be found in the provided video file. Testing Deva in the driving domain, which differs significantly from its training data, results in poor mask quality and accumulating errors over time. Moreover, there is no effective mechanism to handle the rapid entry and exit of objects in a scene, a common occurrence in real-world applications like autonomous driving. In contrast, our method exhibits stable performance in such scenarios.

\begin{figure*}[t]
\centering
\includegraphics[width=0.9\linewidth]{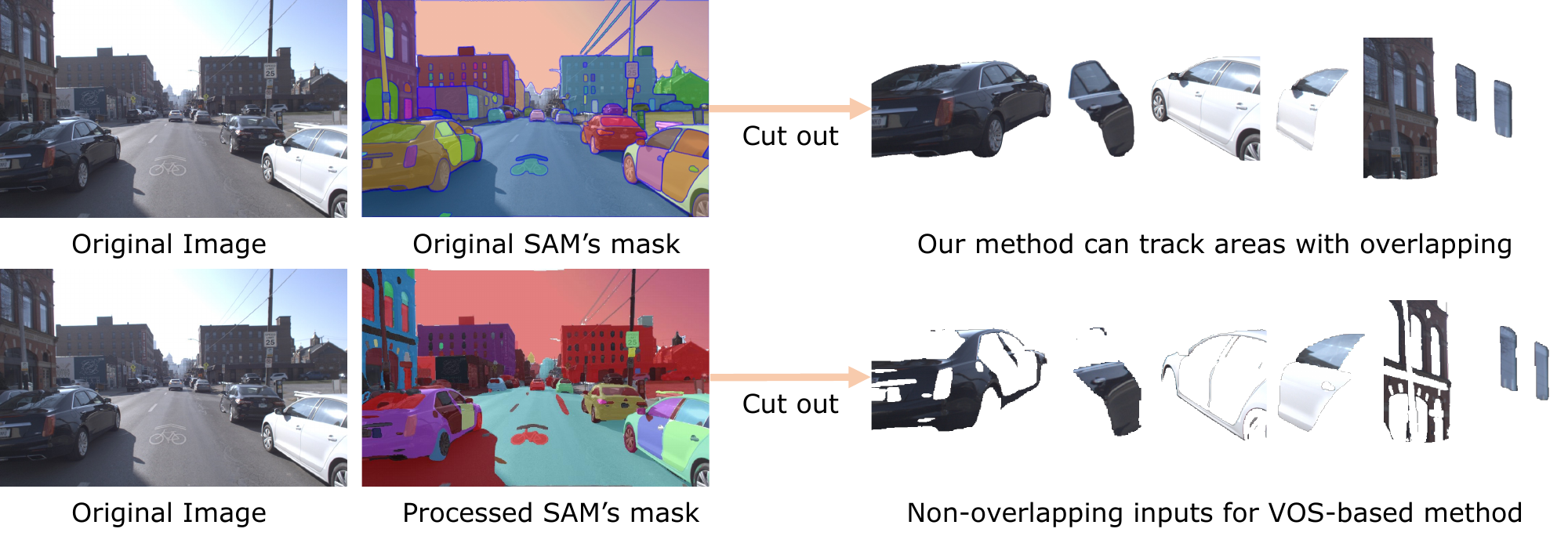}
\caption{Challenges of VOS-Based Methods with Multi-Granular Pixel Overlaps. This figure illustrates the complexity encountered when dealing with overlapping masks in SAM's output, where a single pixel may be associated with multiple instances at different granularities. Traditional VOS methods, operating under the assumption that each pixel belongs to only one instance, often resort to heuristics to resolve these overlaps, as depicted in the second row. In contrast, our method effectively handles such overlapping masks, showcasing its adaptability in complex scenarios.}
\label{fig-vos-overlapping}
\end{figure*}

\noindent\textbf{Difficulty in Managing Multiple Granularities of Pixels:}
Furthermore, these methods are primarily developed for video object segmentation (VOS) tasks, which typically involve videos and annotations of single, rather than multiple, diverse objects. As a result, most VOS-based approaches are designed to track only one instance at a time. While recent advancements like those in \cite{deva, DeAOT} allow for the simultaneous tracking of multiple instances, they often work on the premise that each pixel is part of a single instance. This overlooks complexities in pixel granularity, where a pixel may be part of multiple instances depending on the level of granularity—a common situation in the outputs of SAM, as depicted in Figure~\ref{fig-vos-overlapping}. This issue is further illustrated using the UVO dataset, which contains only coarse object-level annotations, often omitting finer details of object parts.

We apply SAM to generate mask predictions for each frame in the UVO dataset for both methods. To track objects segmented by SAM, a VOS-based method like Deva has to resolve overlaps by assigning each pixel to a unique instance. For example, if a group of pixels belongs to a part of an object, it must decide whether to track the part or the whole object. Assigning pixels to a part implies that the corresponding object is partially excluded, as shown with the cars in Figure~\ref{fig-vos-overlapping}. Conversely, assigning pixels to the object results in the removal of the part mask. We present the quantitative results of these scenarios on the UVO dataset in Table~\ref{tab-supp-track-parts-UVO}. Tracking parts leads to an incomplete representation of object masks on UVO, thus affecting performance negatively. In contrast, our method, capable of handling multiple granularities, tracks both entire objects and their parts without compromising performance on the UVO dataset.

\begin{table}[t]
    \caption{Performance Comparison on the UVO Dataset for tracking objects and their parts. This table presents a detailed analysis of tracking performance using the UVO dataset. VOS-based methods like Deva have to resolve overlaps by assigning each pixel to a unique instance. Tracking parts leads to an incomplete representation of object masks on UVO, thus affecting performance negatively. In contrast, our method, capable of handling multiple granularities, tracks both entire objects and their parts without compromising performance on the UVO dataset.}.%
    \centering
    \resizebox{1\linewidth}{!}{    \resizebox{1\linewidth}{!}{%
        \begin{tabular}{llc}
            \toprule
             Track & Method & AR100 \\
            \midrule
            \multirow{4}{*}{video} 
            &\multicolumn{2}{l}{\underline{\textit{Zero-shot test}}}  \\
    
            & Deva-SAM-H: track all instances (with parts)~\cite{deva} & 19.4 \\
            & Deva-SAM-H: track only whole objects (no parts) ~\cite{deva} & 36.0 \\
            & \textbf{Ours-SAM-H}: track all instances (with parts) & \rankfirst{37.5} \\
            \bottomrule
        \end{tabular}%
        }}%
    \label{tab-supp-track-parts-UVO}%
\end{table}

\section{More Qualitative Results}
\label{sec:qua-res}
We provide \textbf{a video file} containing our qualitative tracking results on multiple domains. Here we provide some visualization results regarding fast proposal generation and dense object association. 

\subsection{Fast Proposal Generation}
In ~\figureautorefname~\ref{fig-qualitative}, we compare the segmentation quality of our fast proposal generation with SAM's original everything mode on raw images from COCO validation set. By default, we output 300 bounding boxes per image, and use a bounding box NMS with 0.5 threshold as the only post-processing during inference. The results show that our fast proposal generation can achieve similar segmentation quality to the everything mode of SAM, despite using much less time.

 \subsection{Open-Vocabulary Tracking} 
\begin{figure*}[t]
    \centering
    \small
    \setlength\tabcolsep{0.5mm}
    \resizebox{1.0\linewidth}{!}{
    \begin{tabular}{ccccc}
        \toprule
        $t$ & $t +1$ & $t + 2$& $t + 3$& $t + 4$\\ \midrule
        \includegraphics[width=0.2\linewidth]{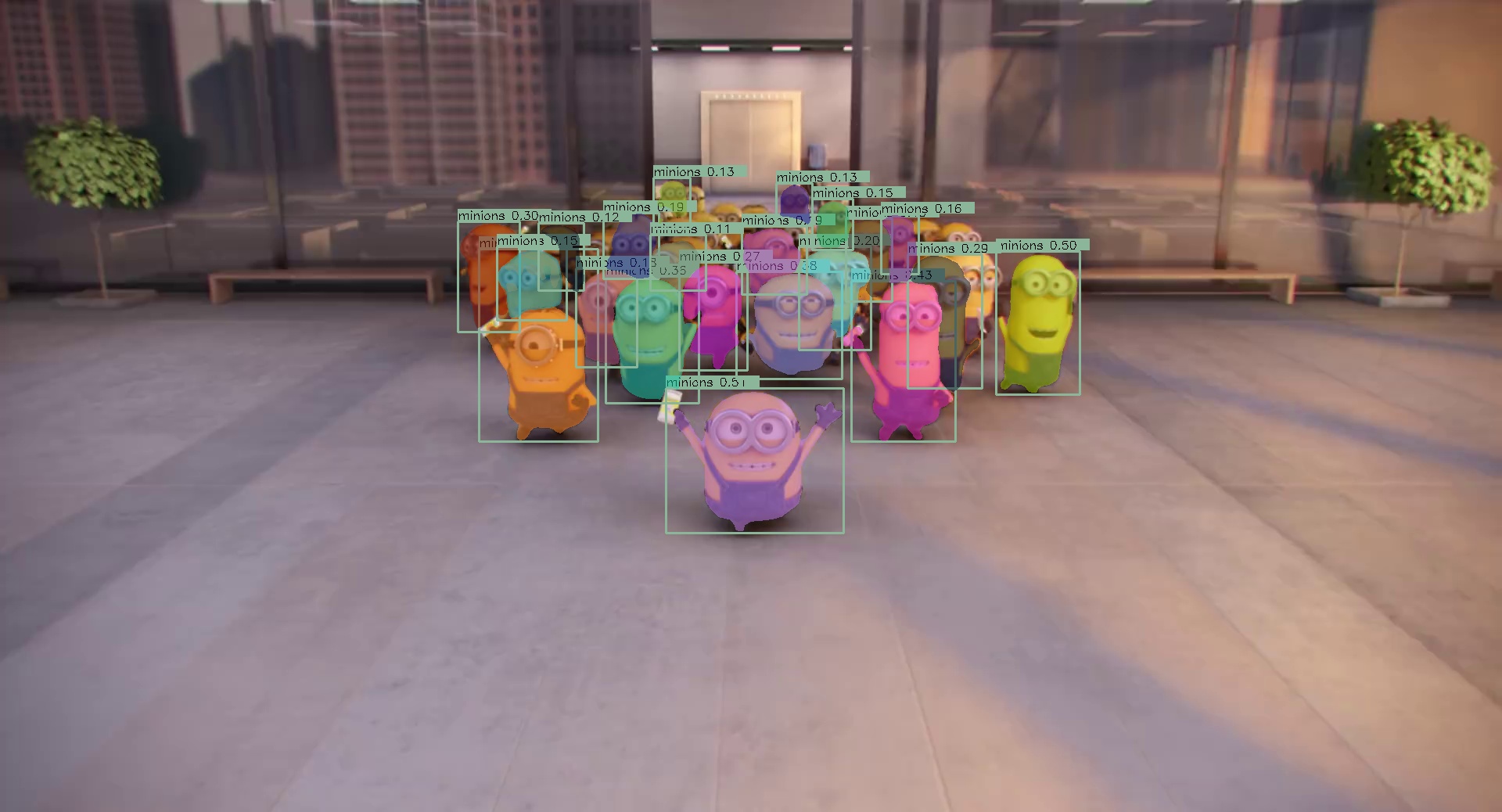} &
        \includegraphics[width=0.2\linewidth]{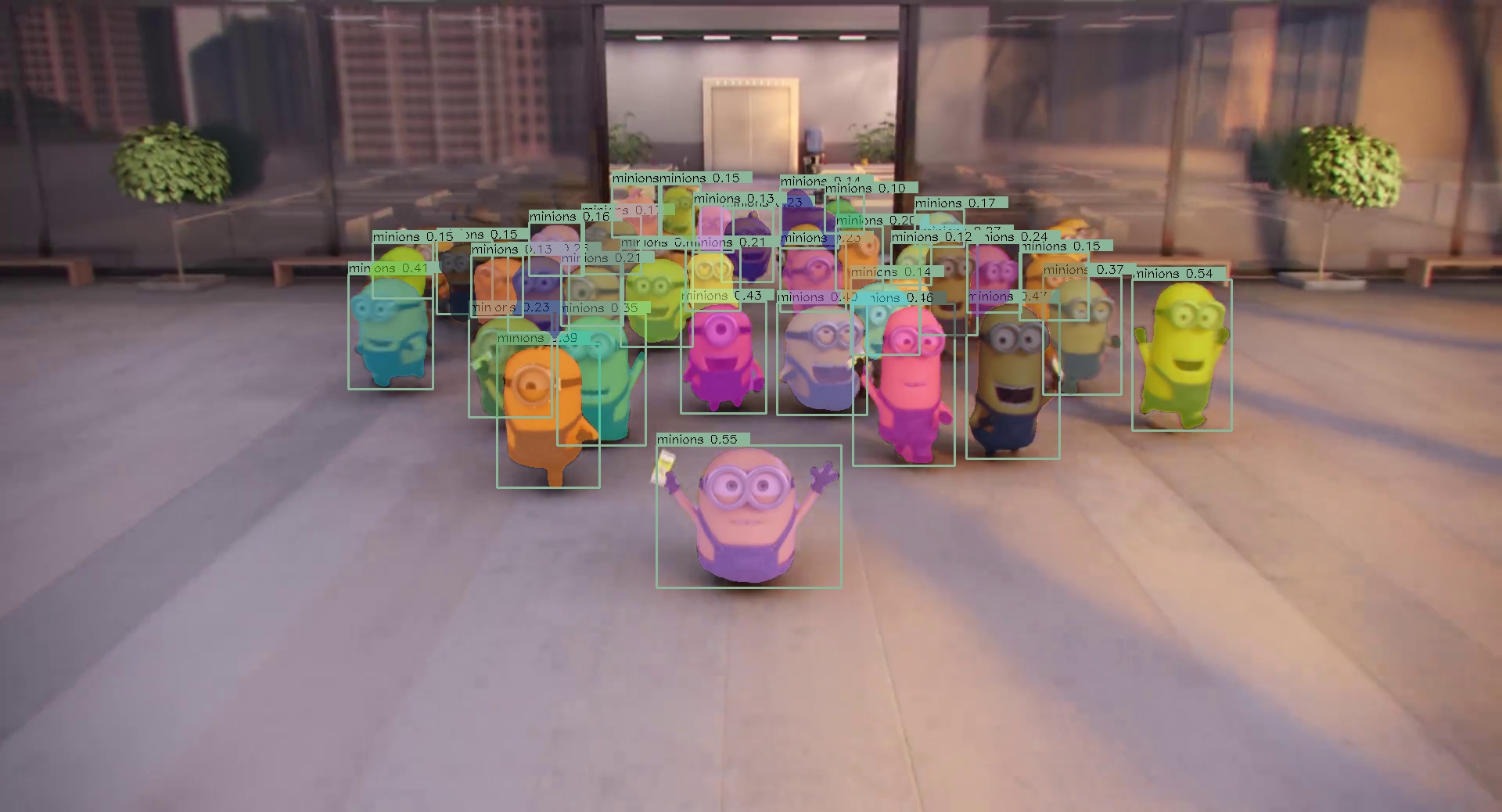 } &
        \includegraphics[width=0.2\linewidth]{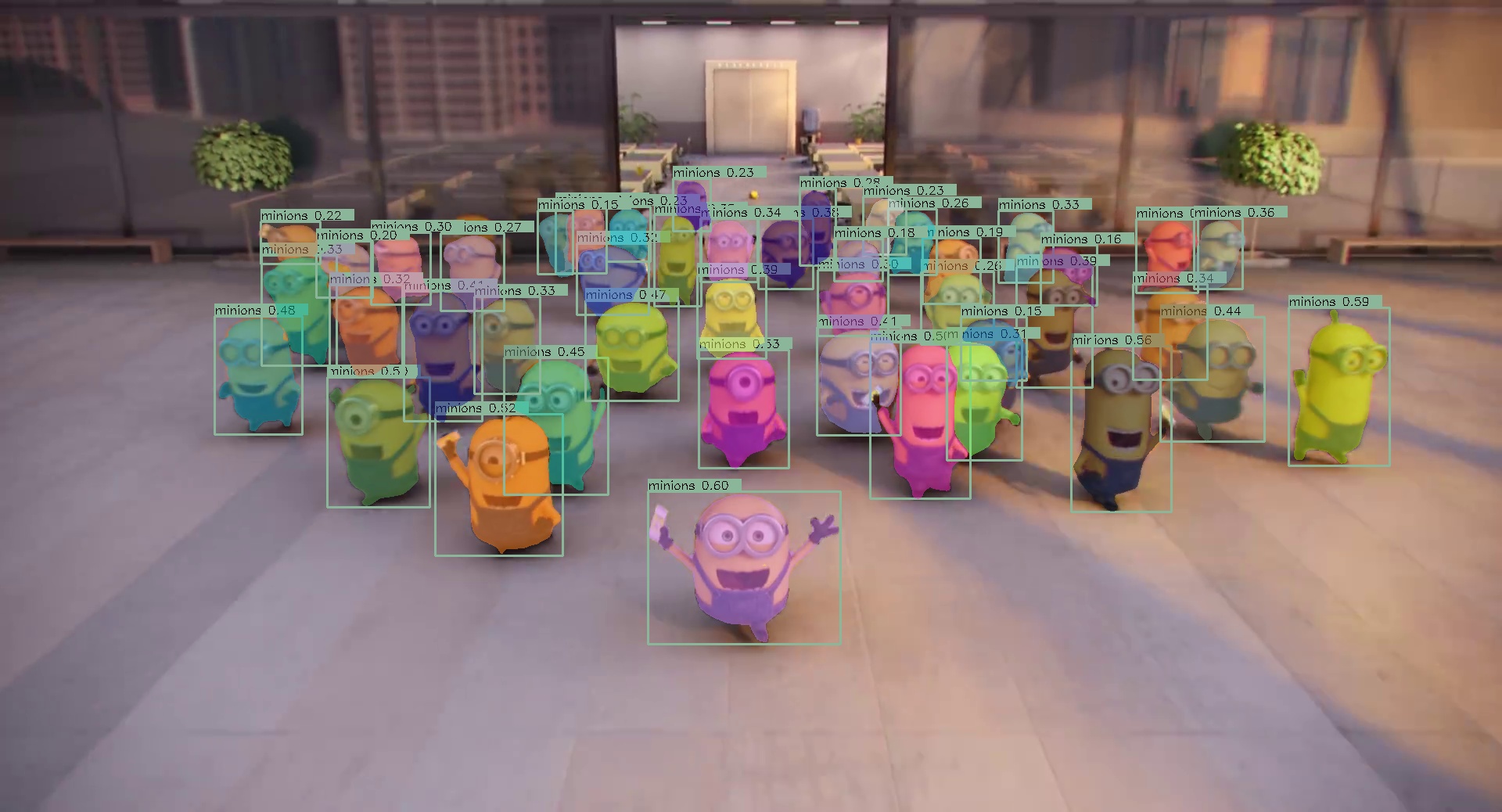} &
        \includegraphics[width=0.2\linewidth]{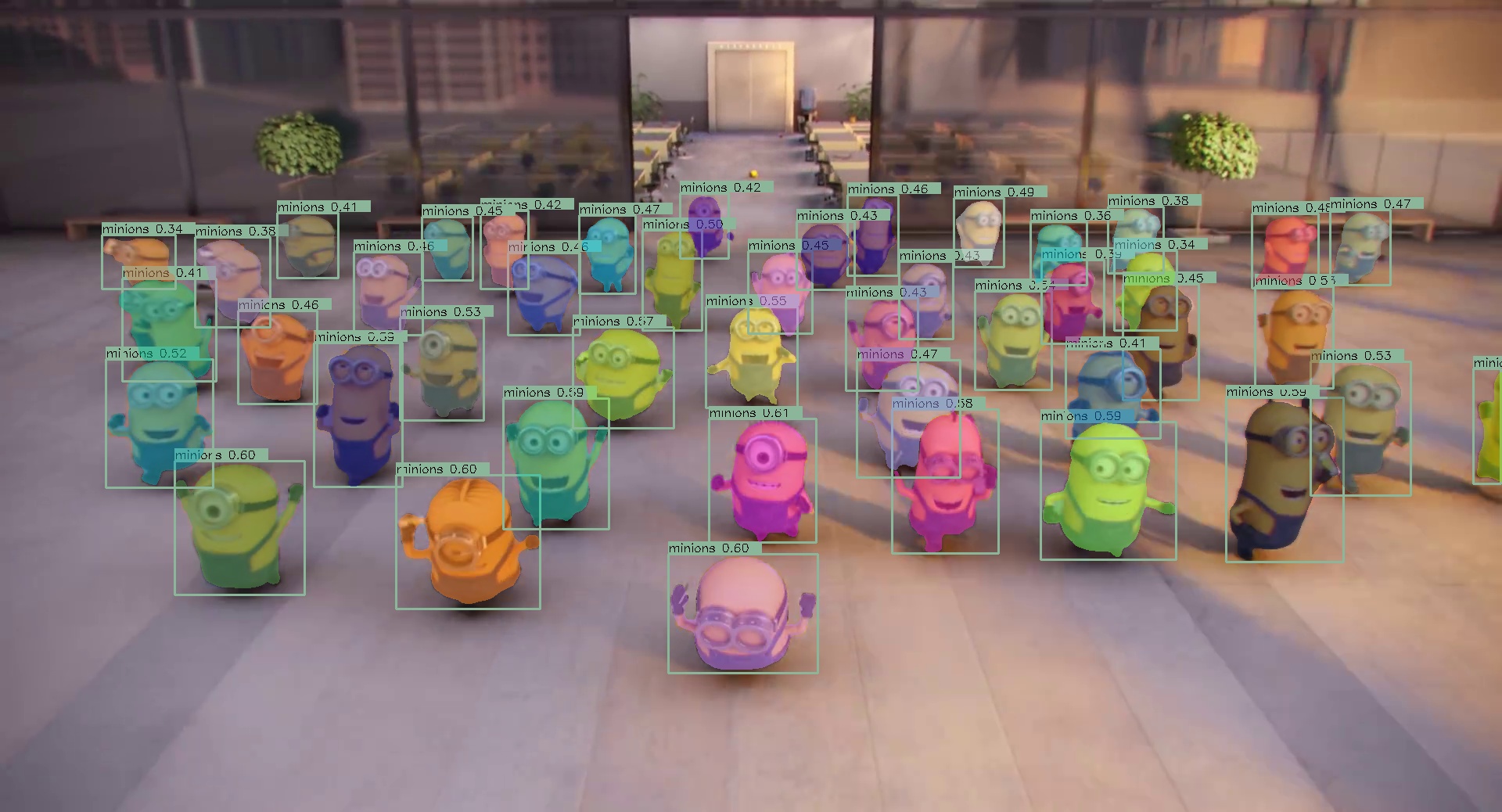} &
        \includegraphics[width=0.2\linewidth]{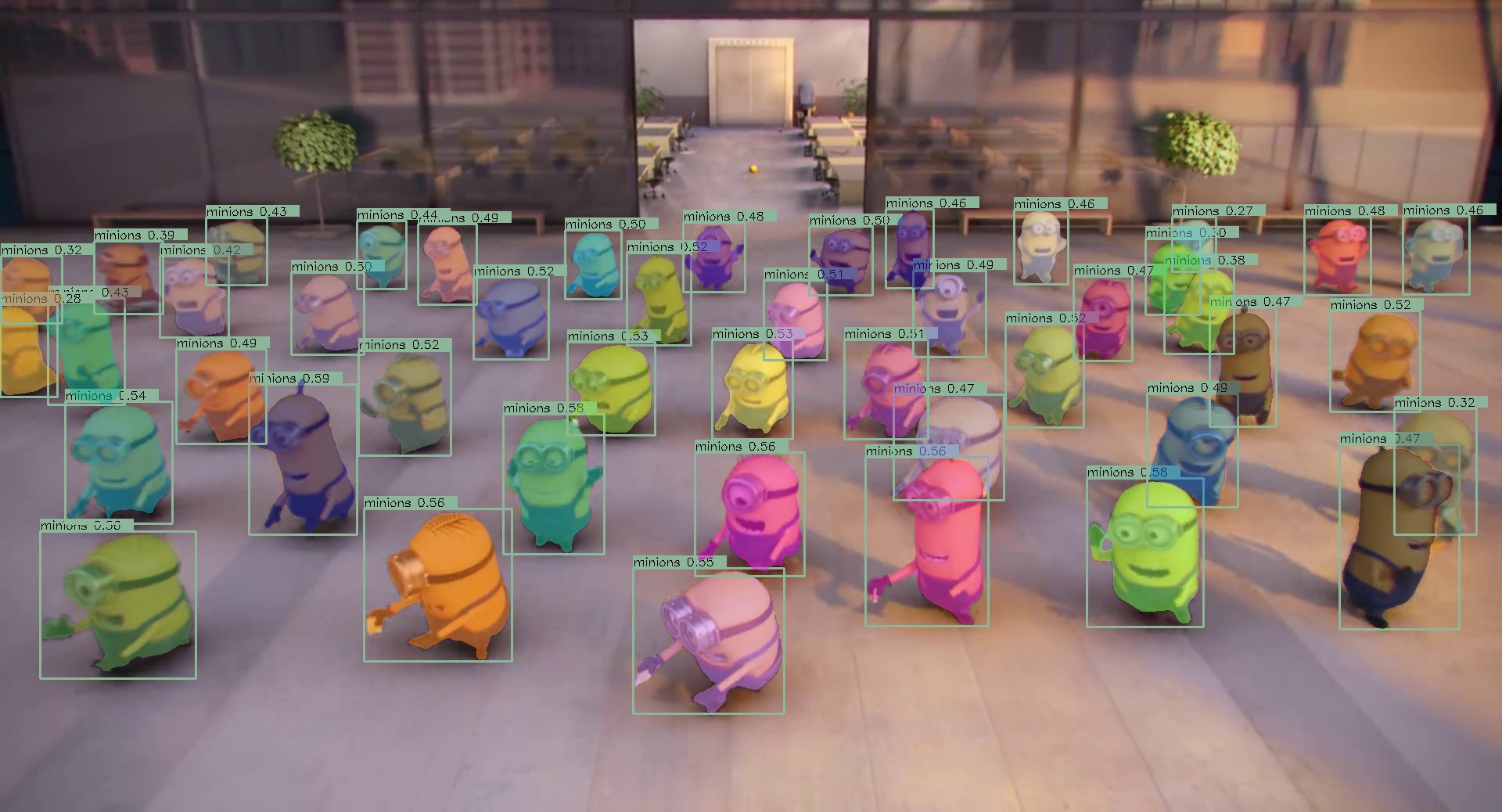} \\

        \includegraphics[width=0.2\linewidth]{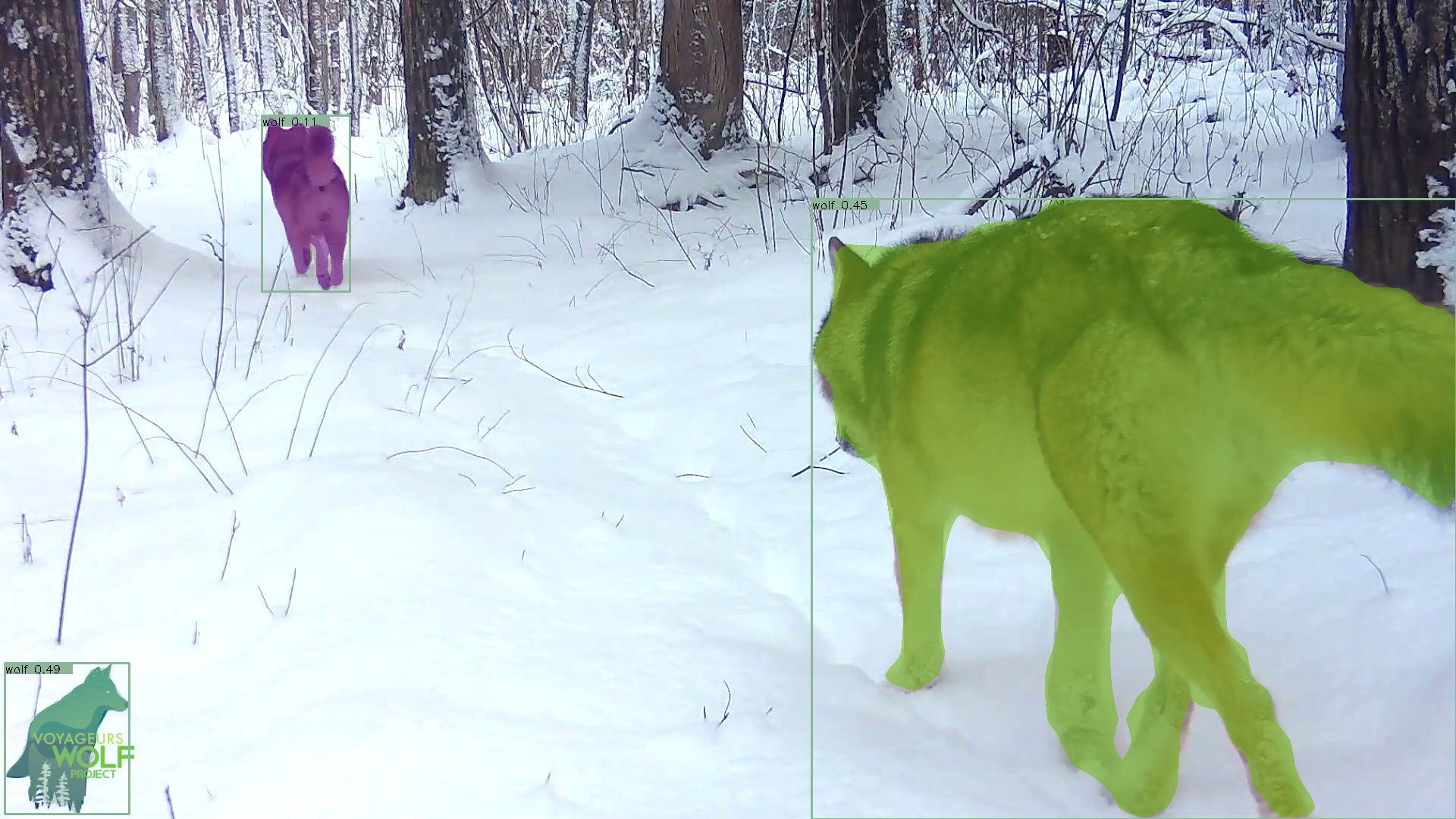} &
        \includegraphics[width=0.2\linewidth]{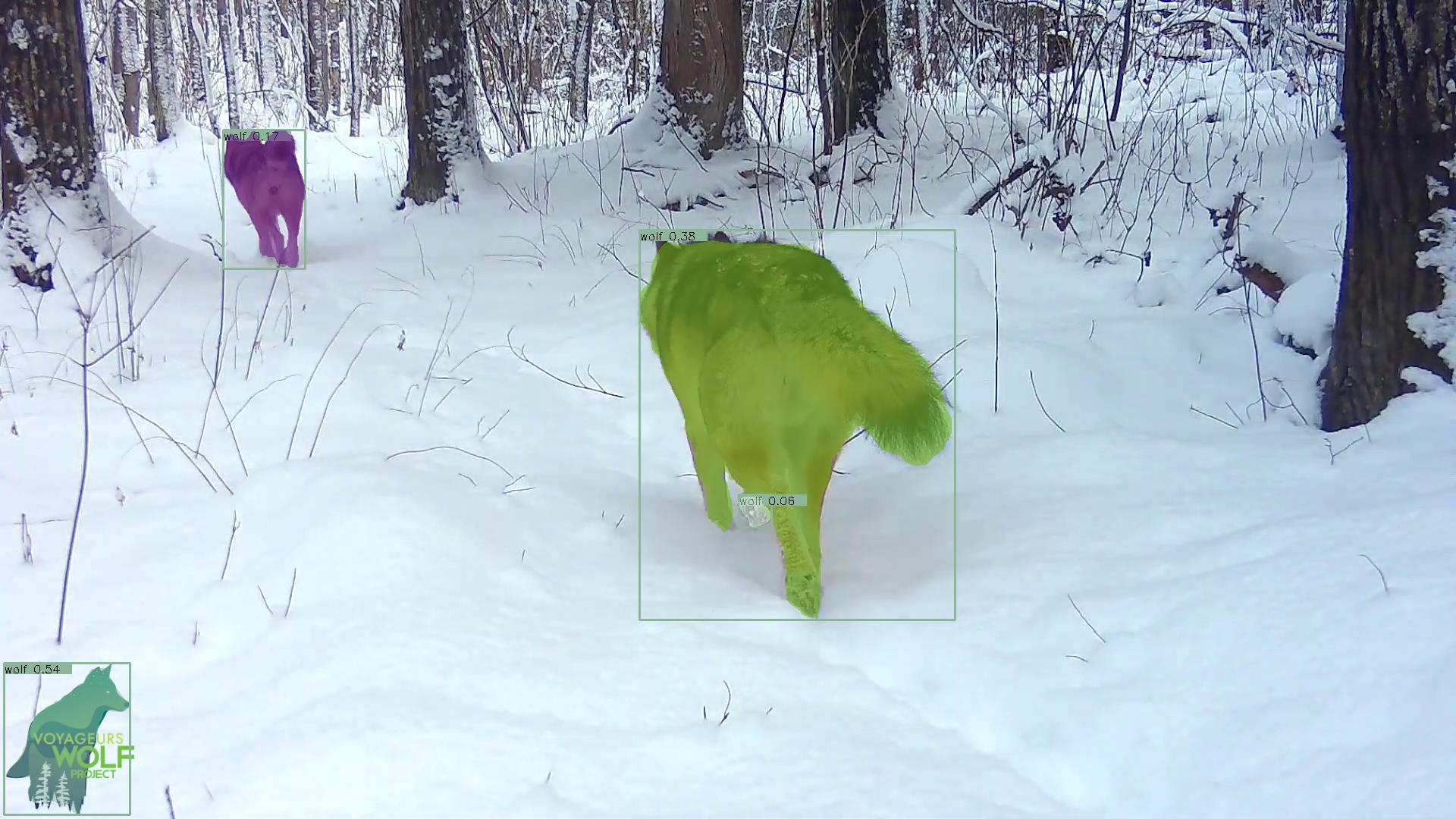} &
        \includegraphics[width=0.2\linewidth]{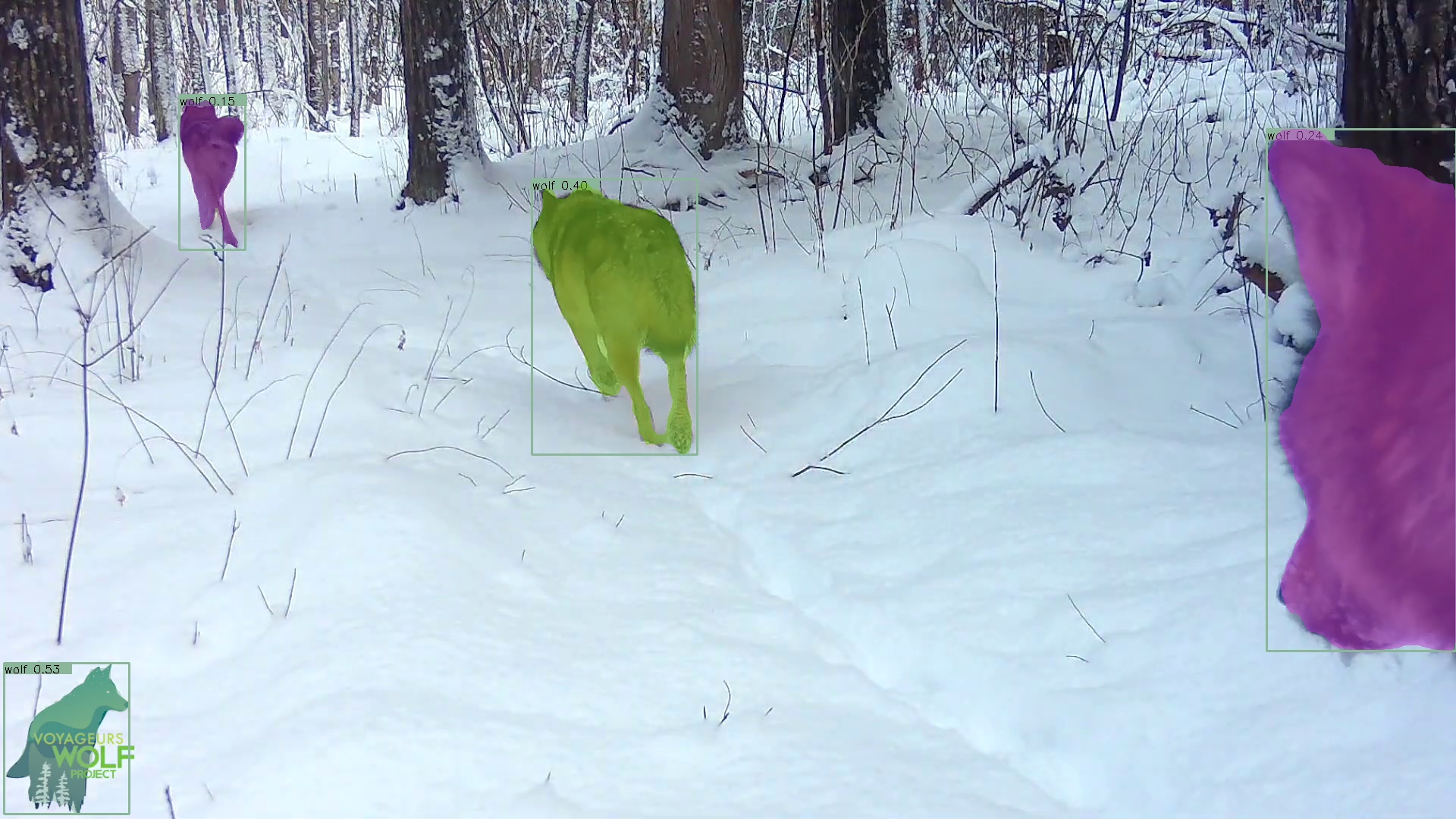} &
        \includegraphics[width=0.2\linewidth]{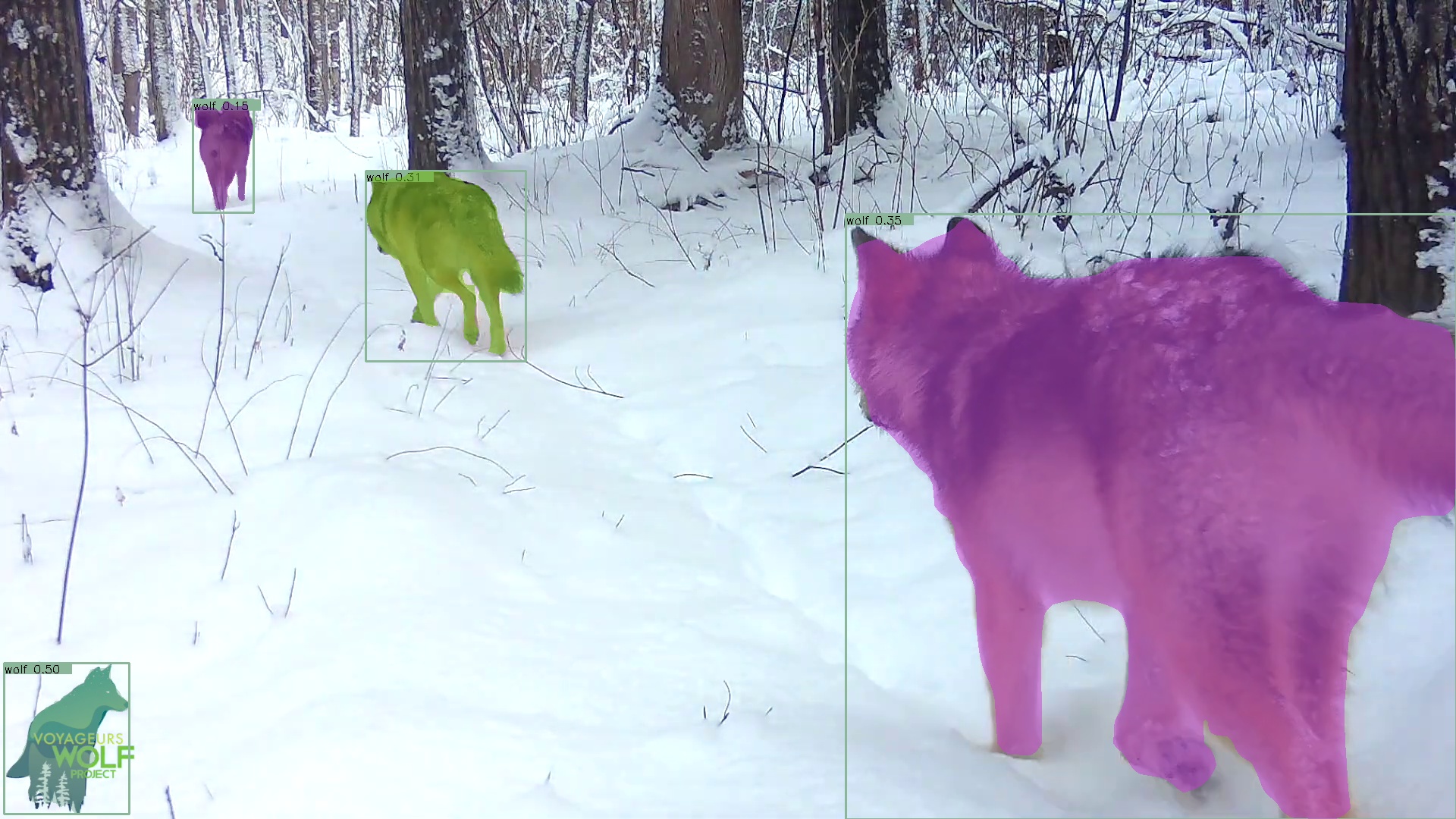} &
        \includegraphics[width=0.2\linewidth]{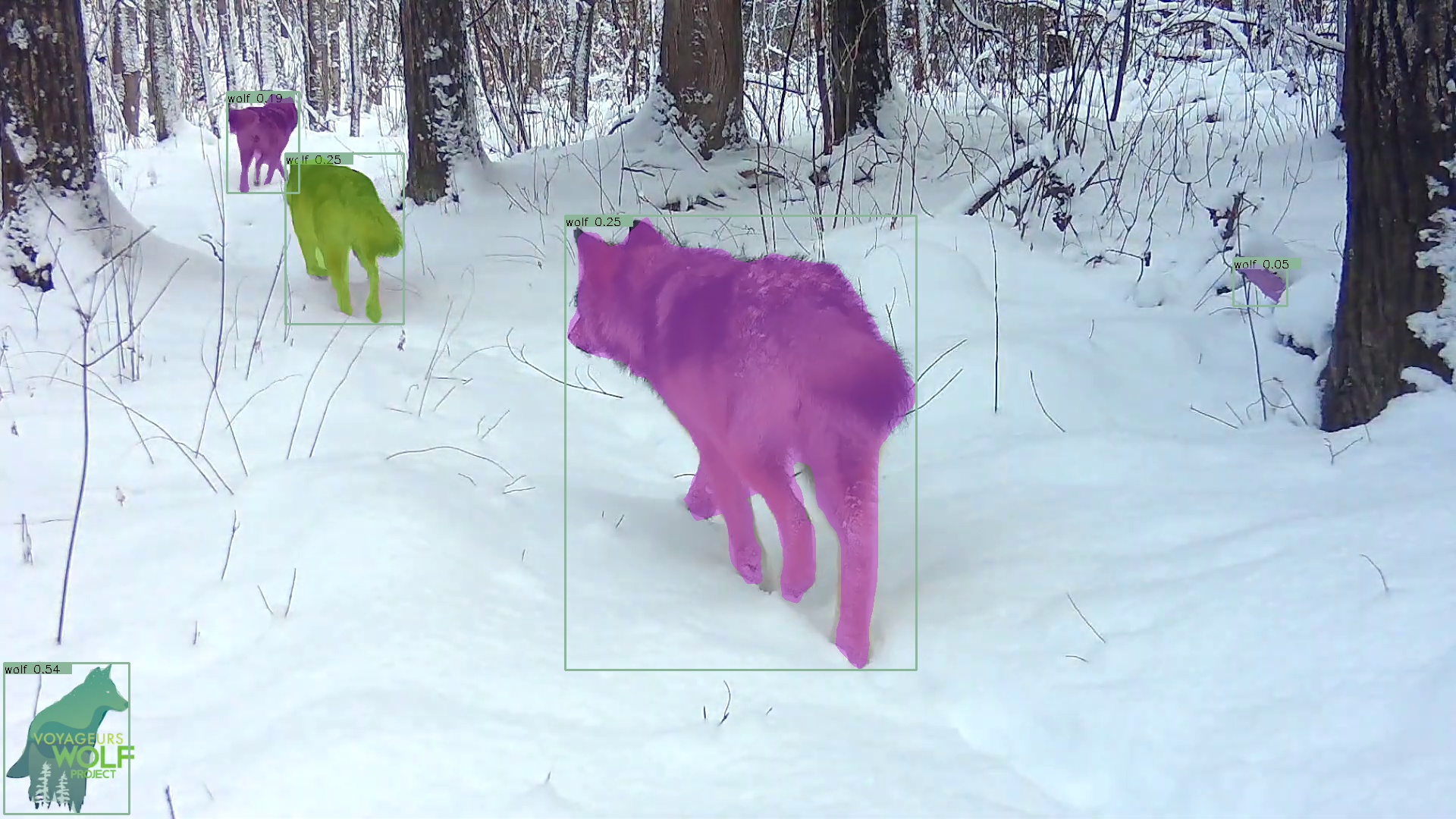} \\
        
        \includegraphics[width=0.2\linewidth]{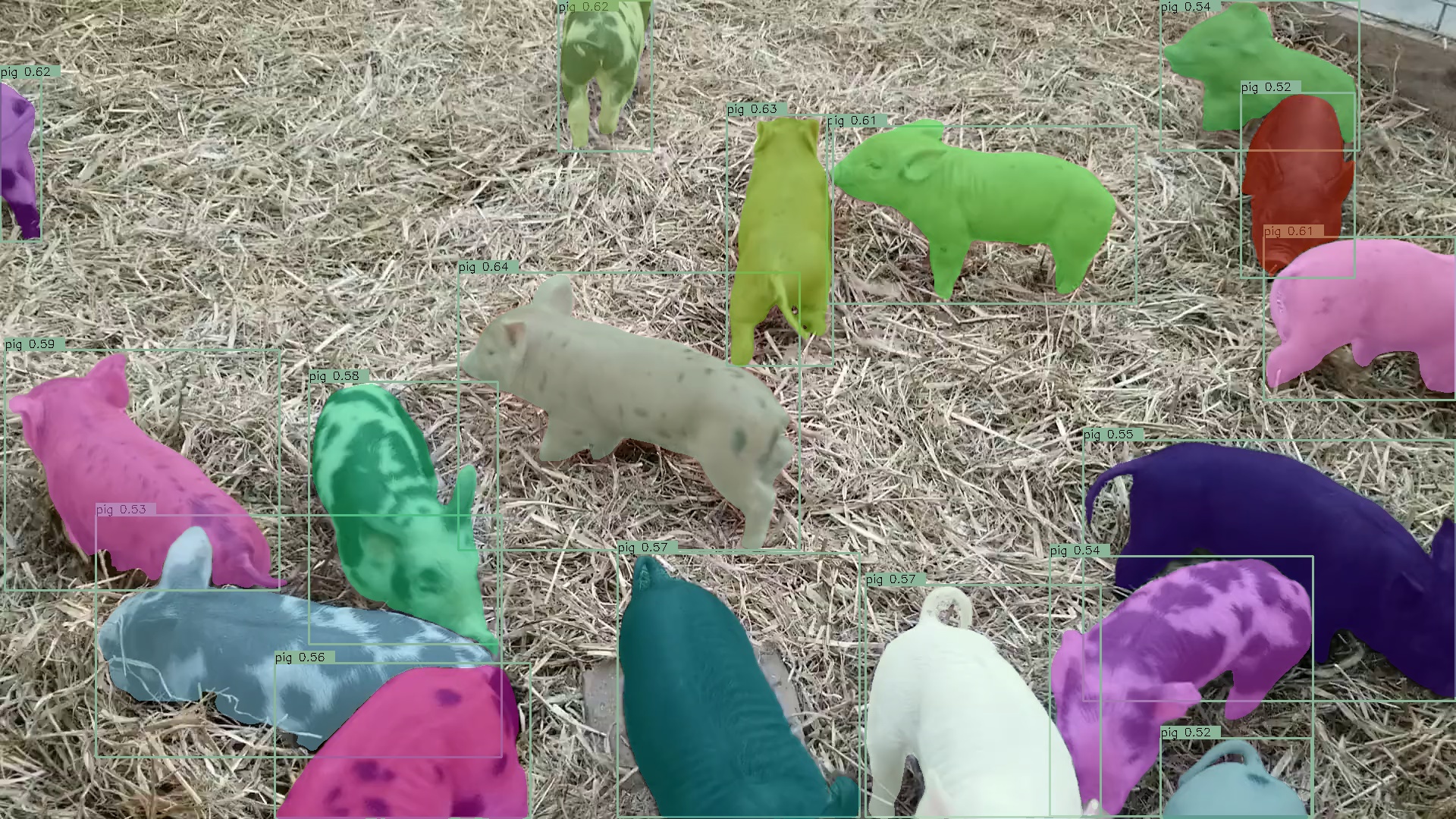} &
        \includegraphics[width=0.2\linewidth]{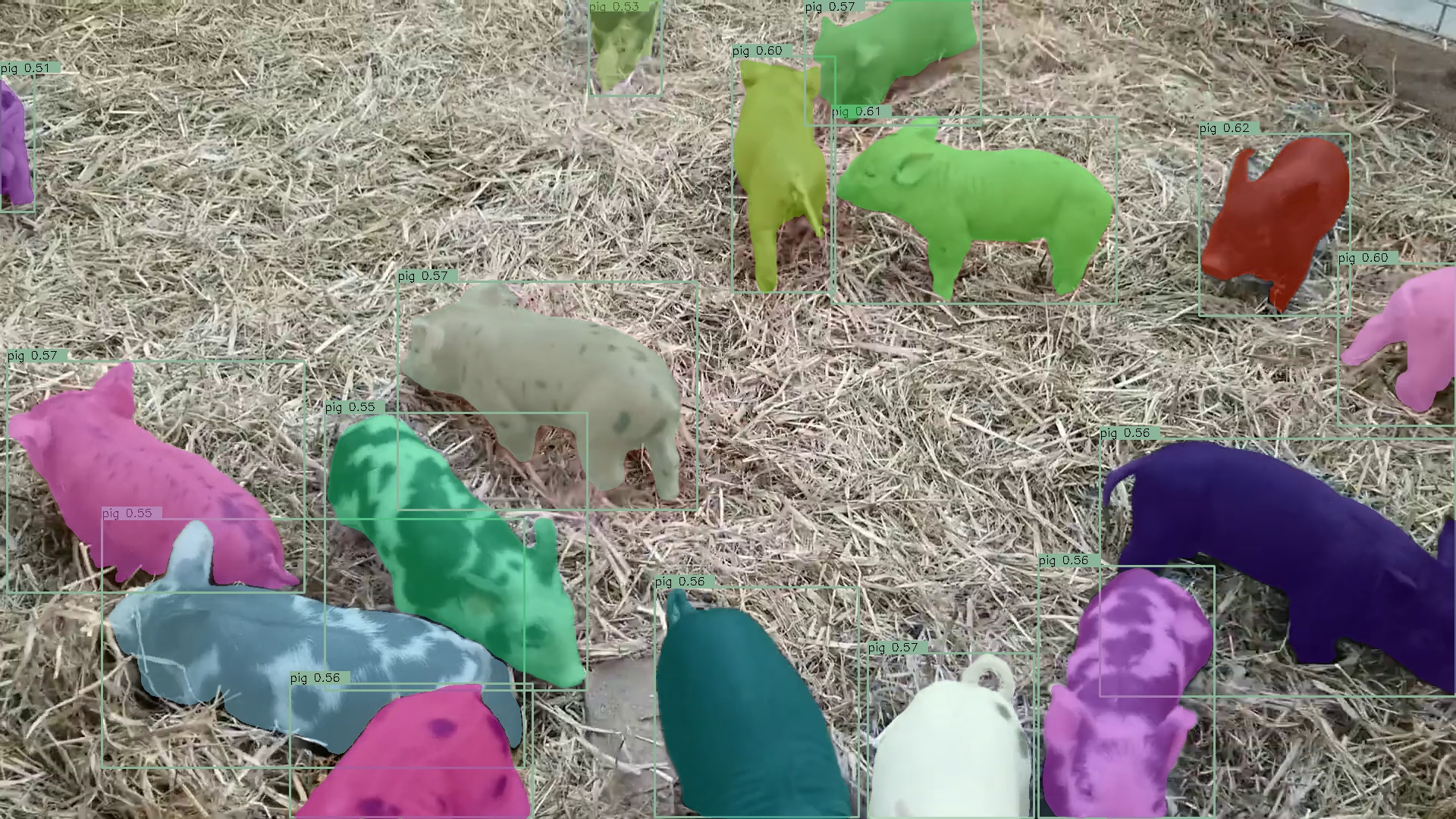} &
        \includegraphics[width=0.2\linewidth]{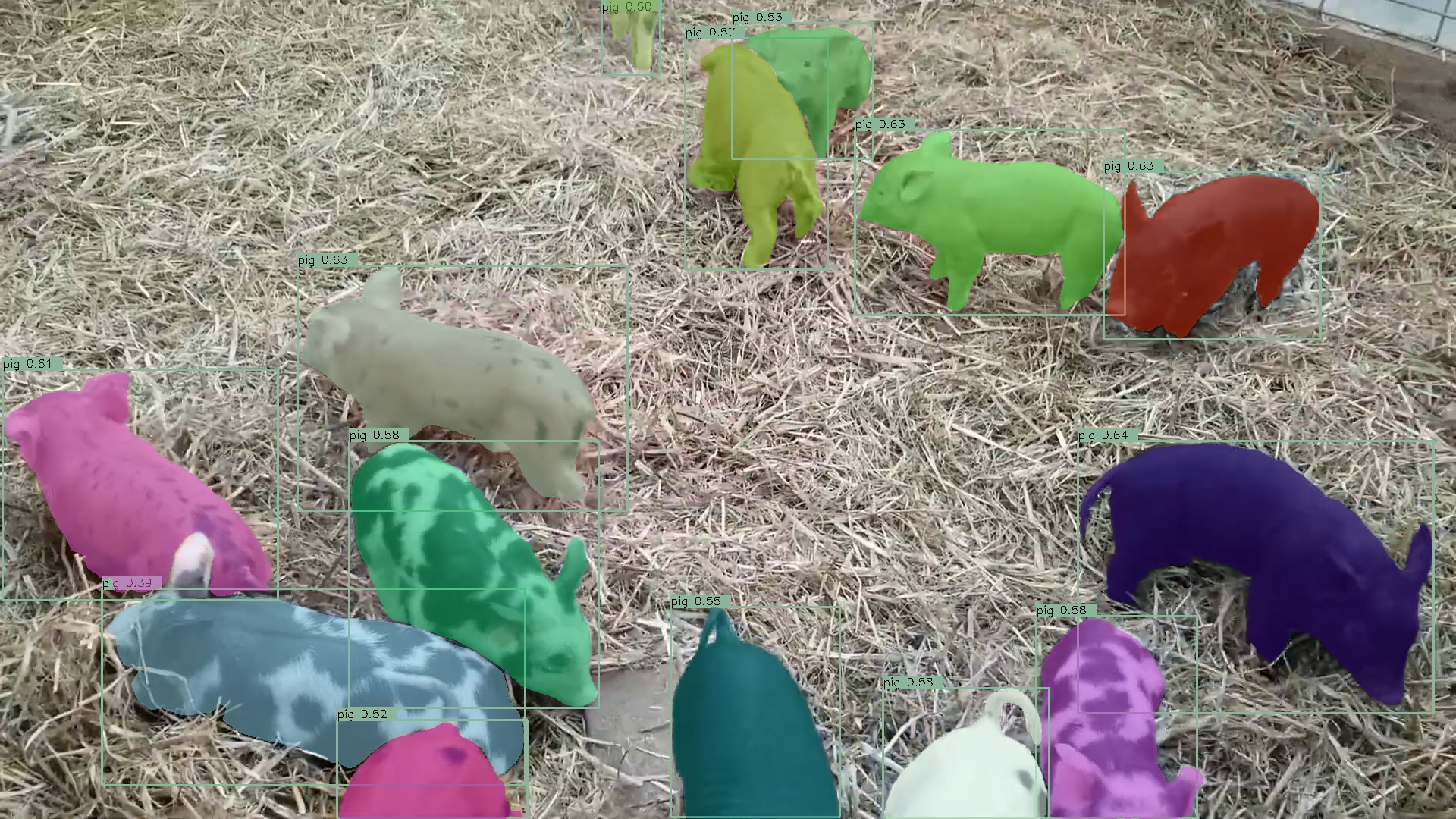} &
        \includegraphics[width=0.2\linewidth]{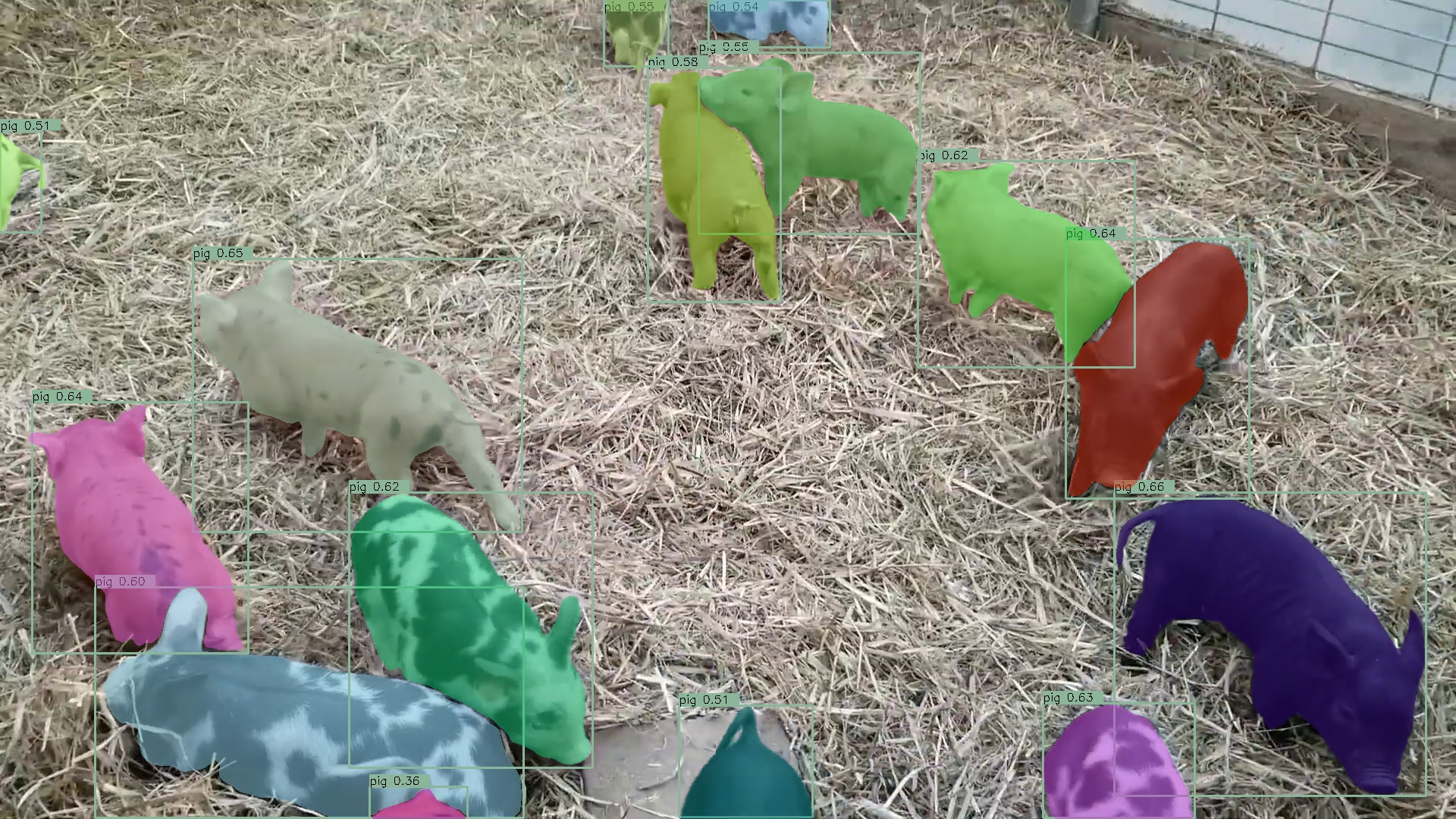} &
        \includegraphics[width=0.2\linewidth]{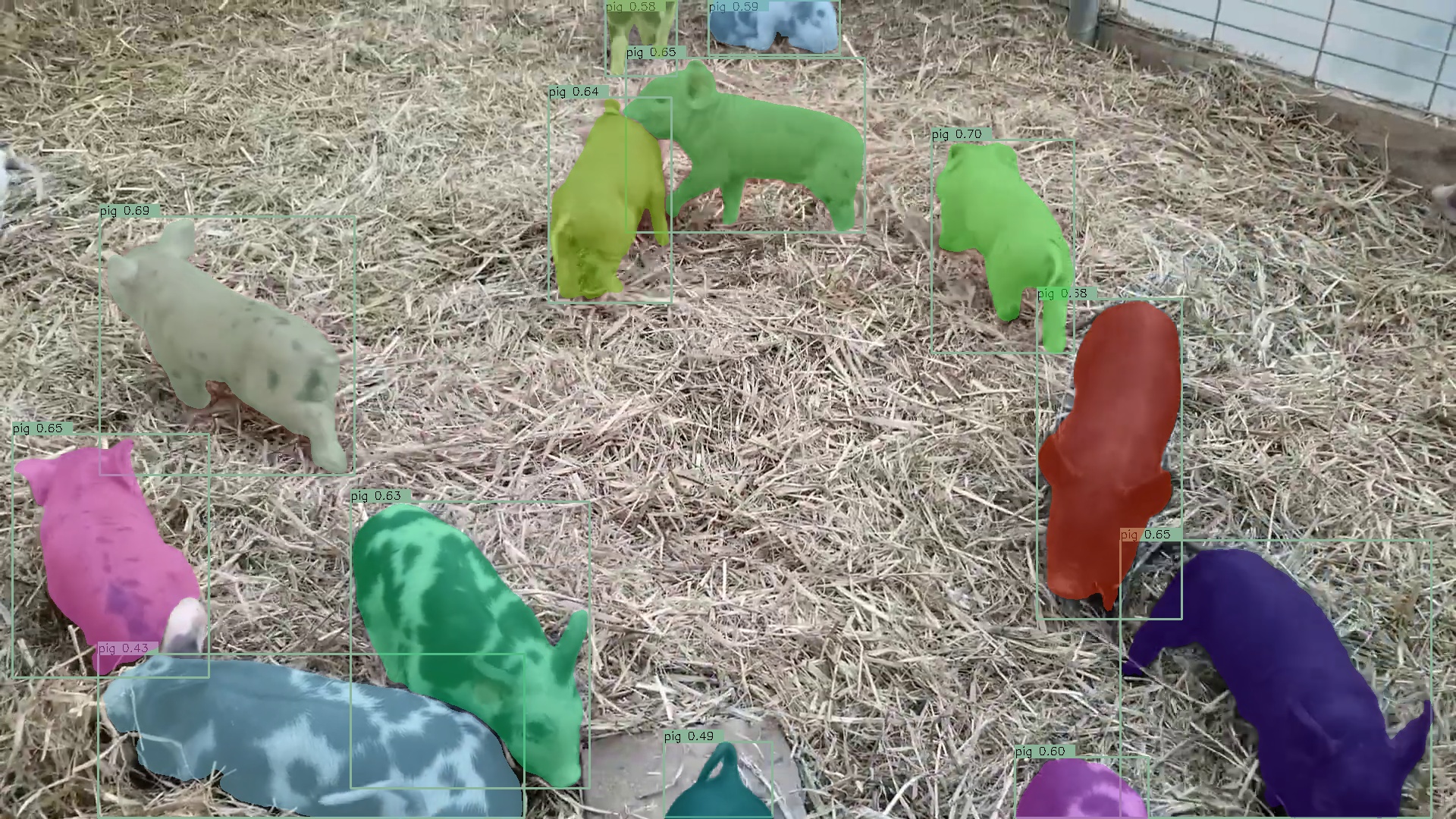} \\
        
        \includegraphics[width=0.2\linewidth]{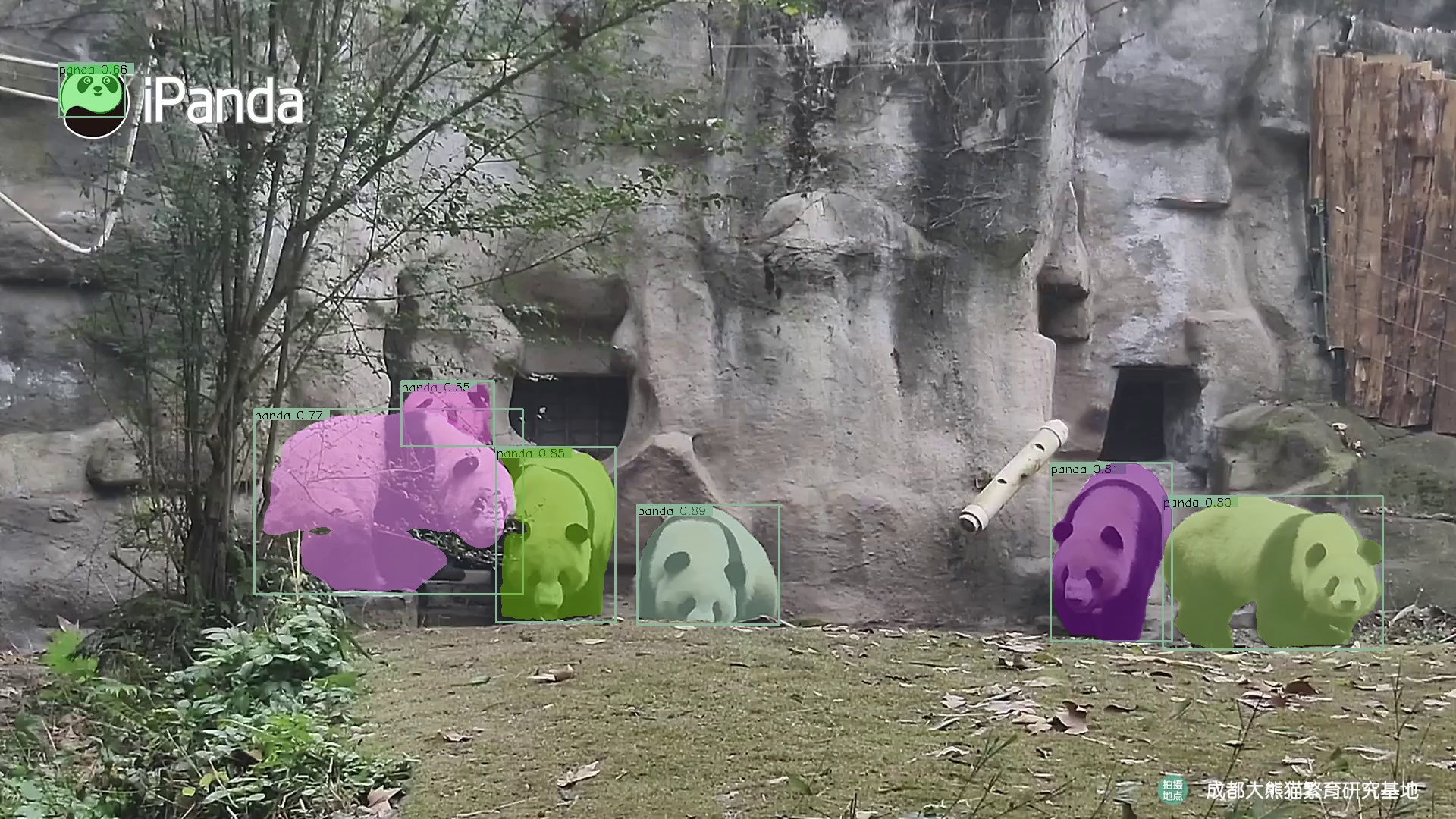} &
        \includegraphics[width=0.2\linewidth]{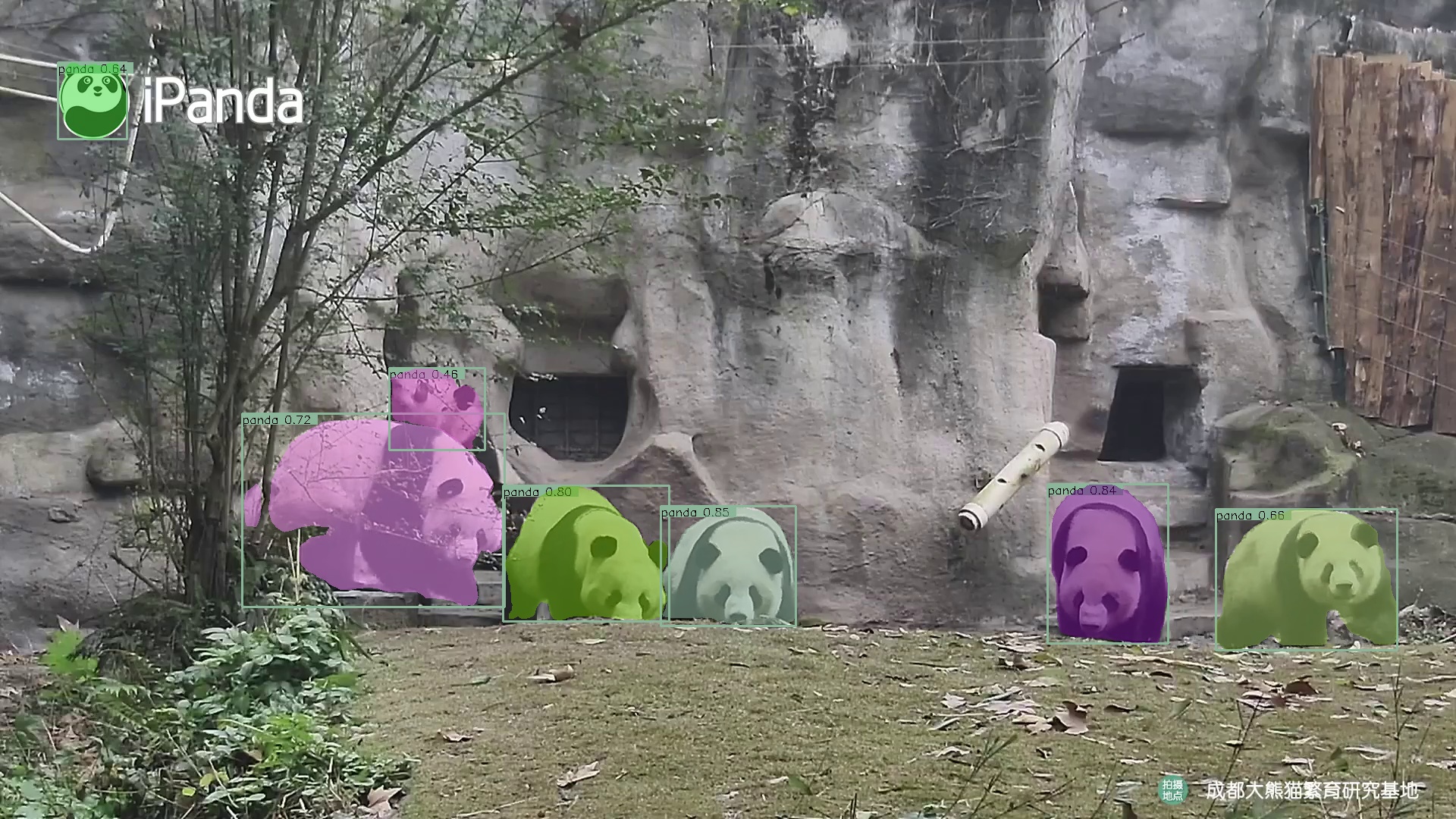} &
        \includegraphics[width=0.2\linewidth]{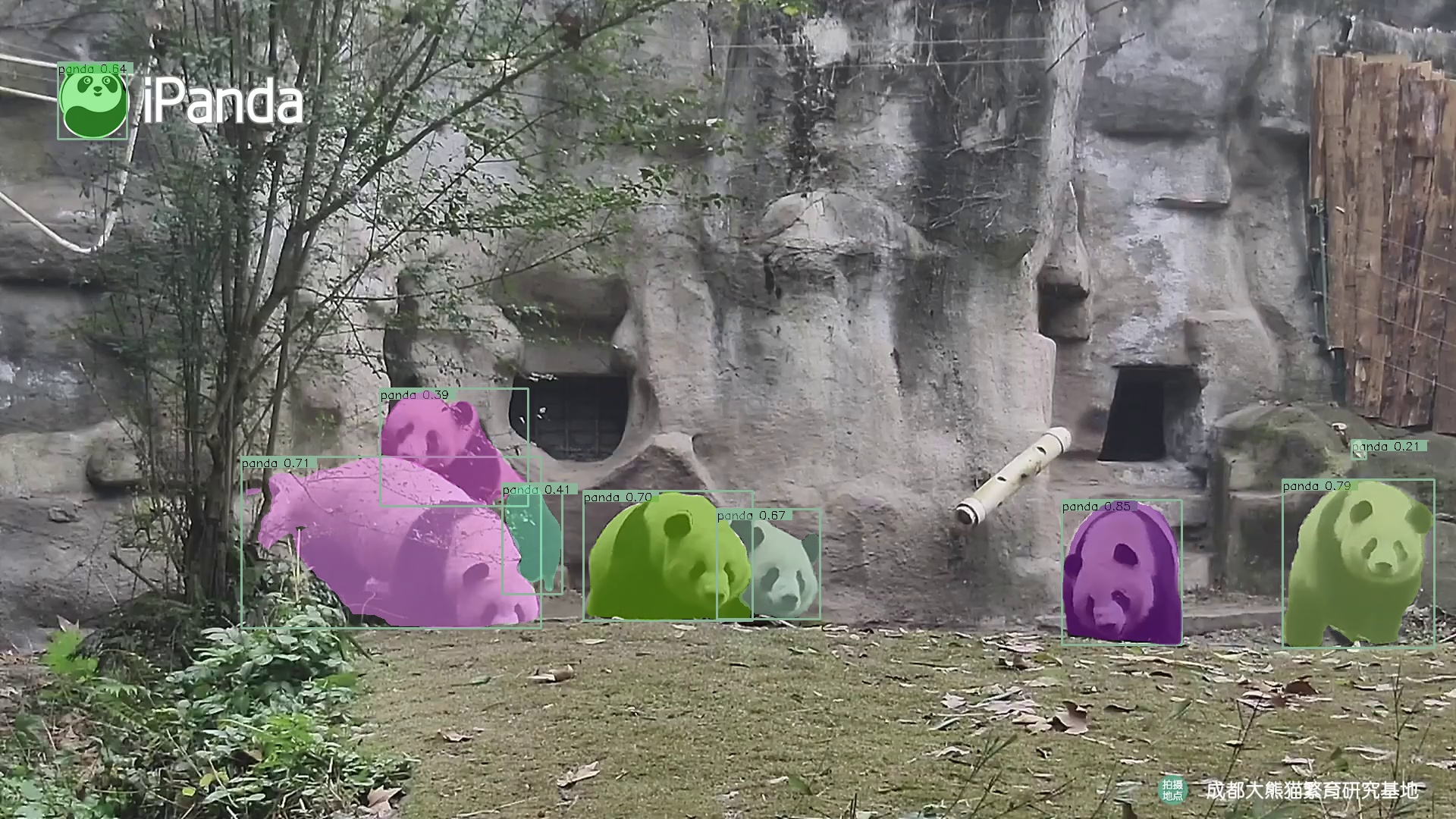} &
        \includegraphics[width=0.2\linewidth]{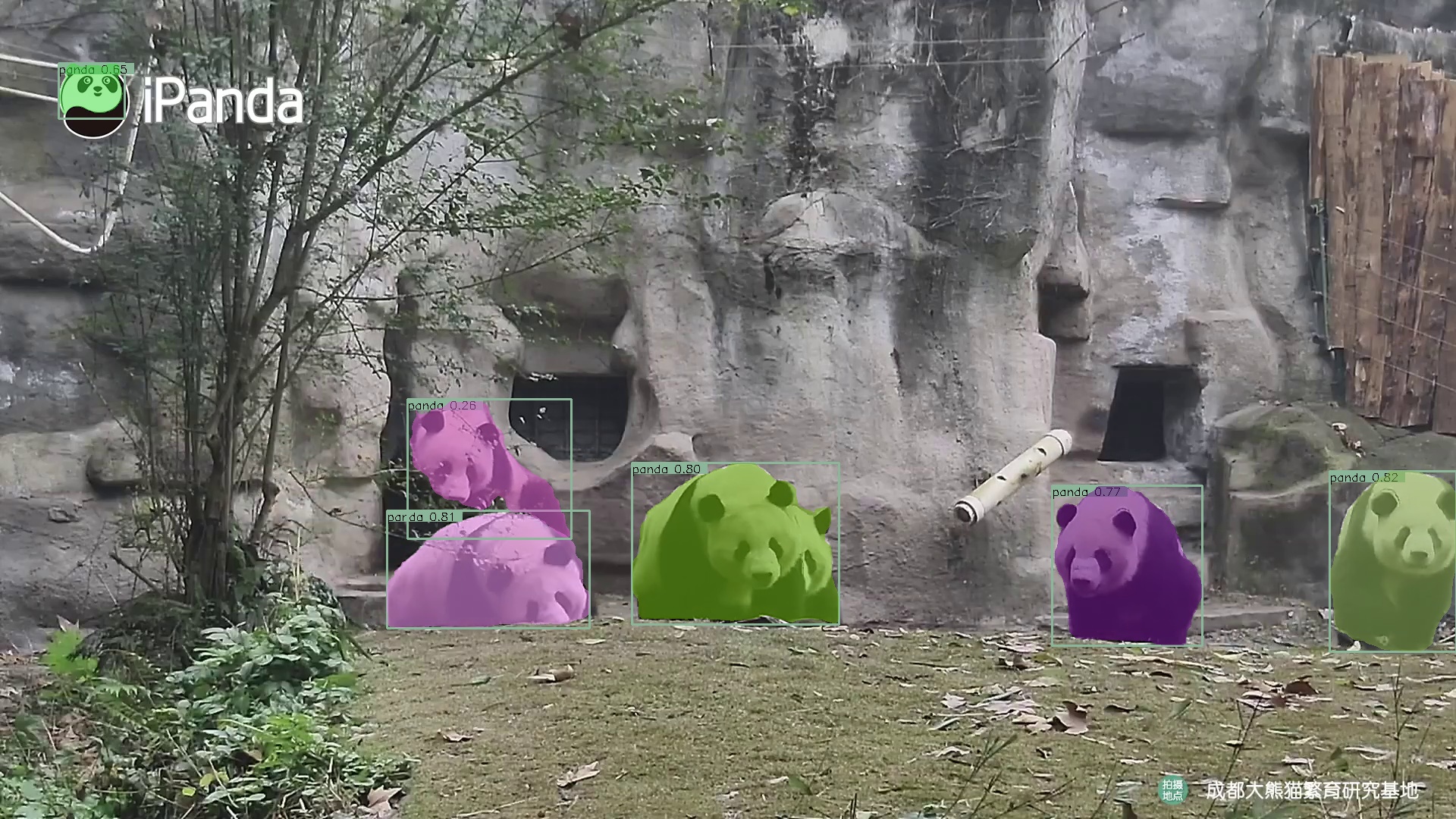} &
        \includegraphics[width=0.2\linewidth]{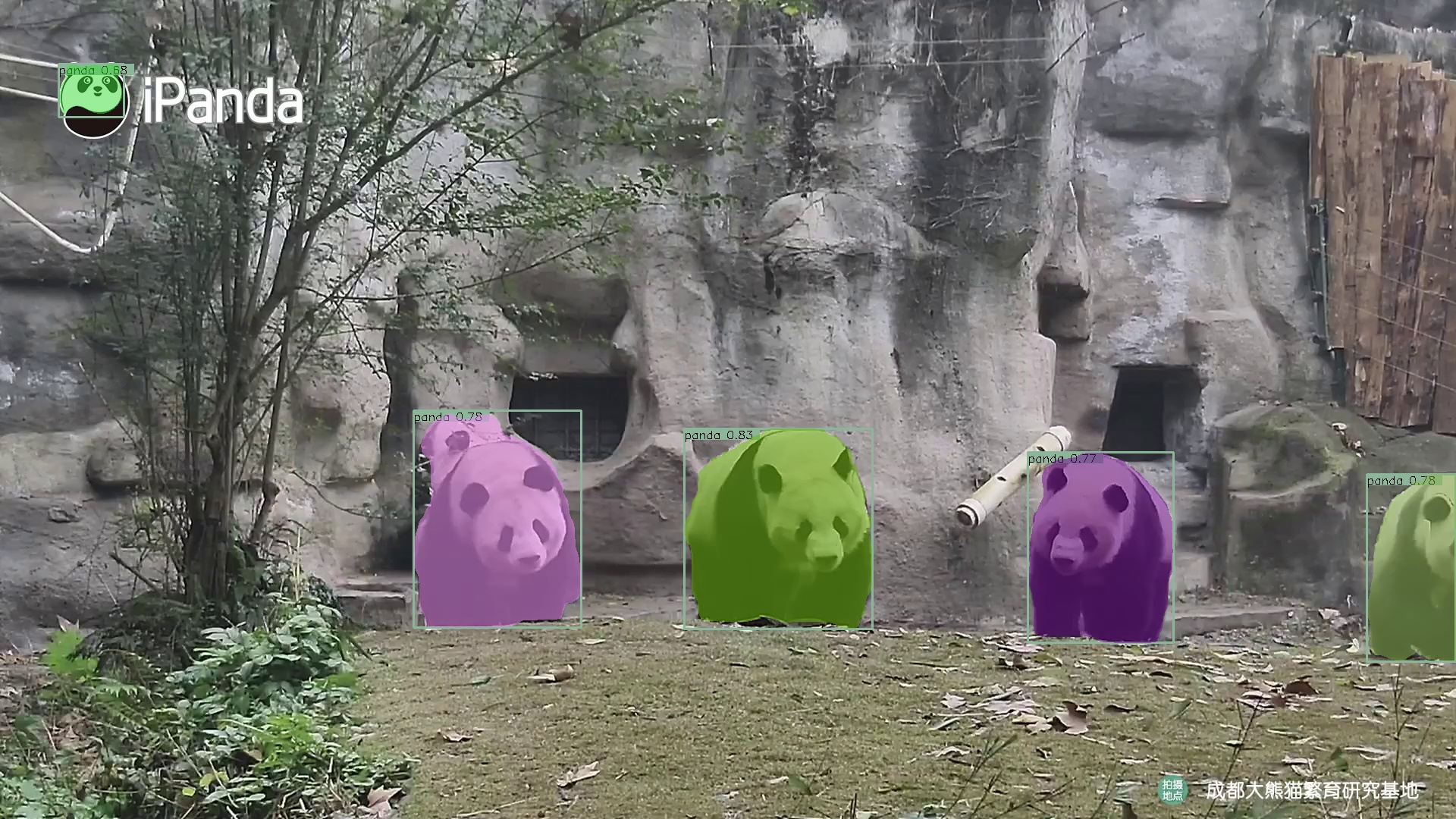} \\
        
        \includegraphics[width=0.2\linewidth]{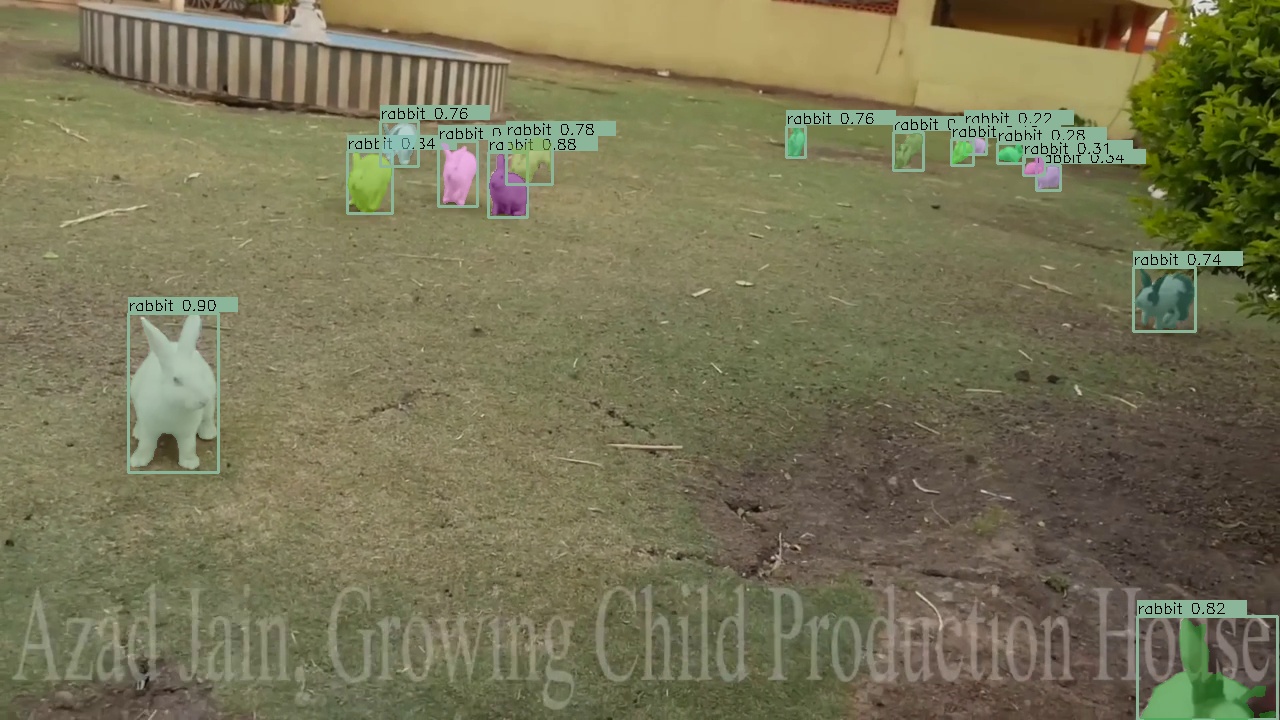} &
        \includegraphics[width=0.2\linewidth]{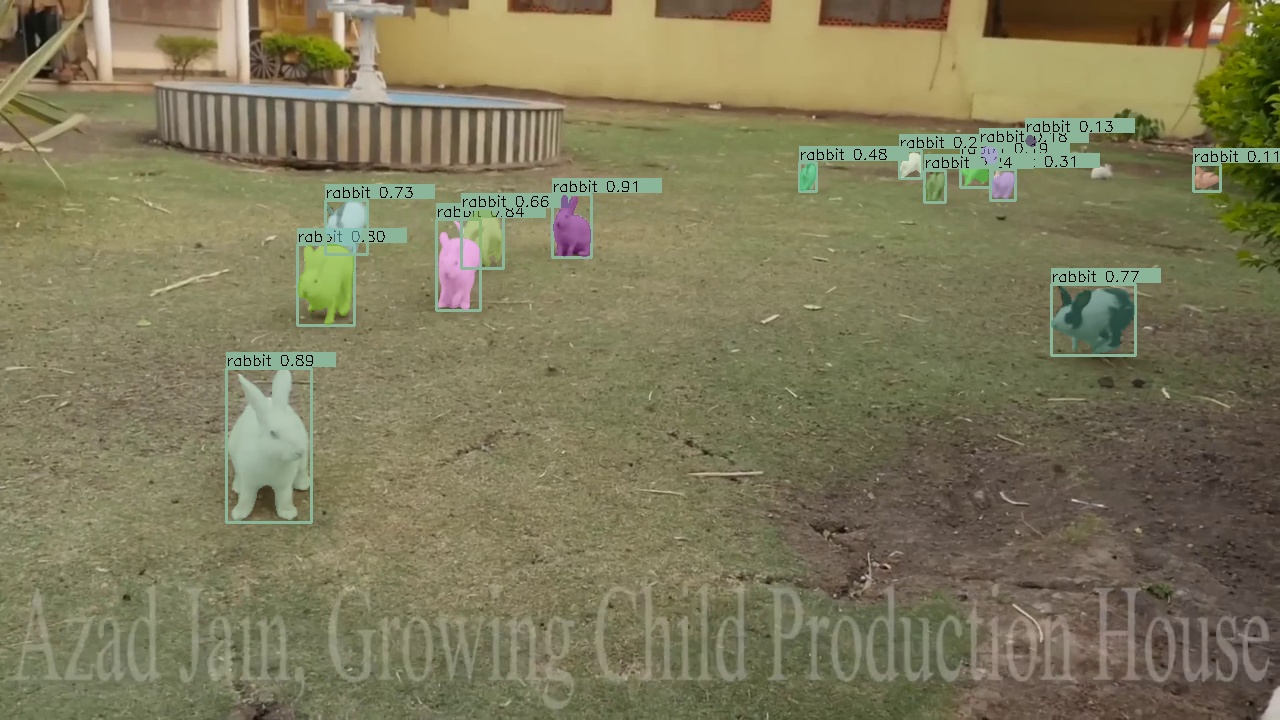} &
        \includegraphics[width=0.2\linewidth]{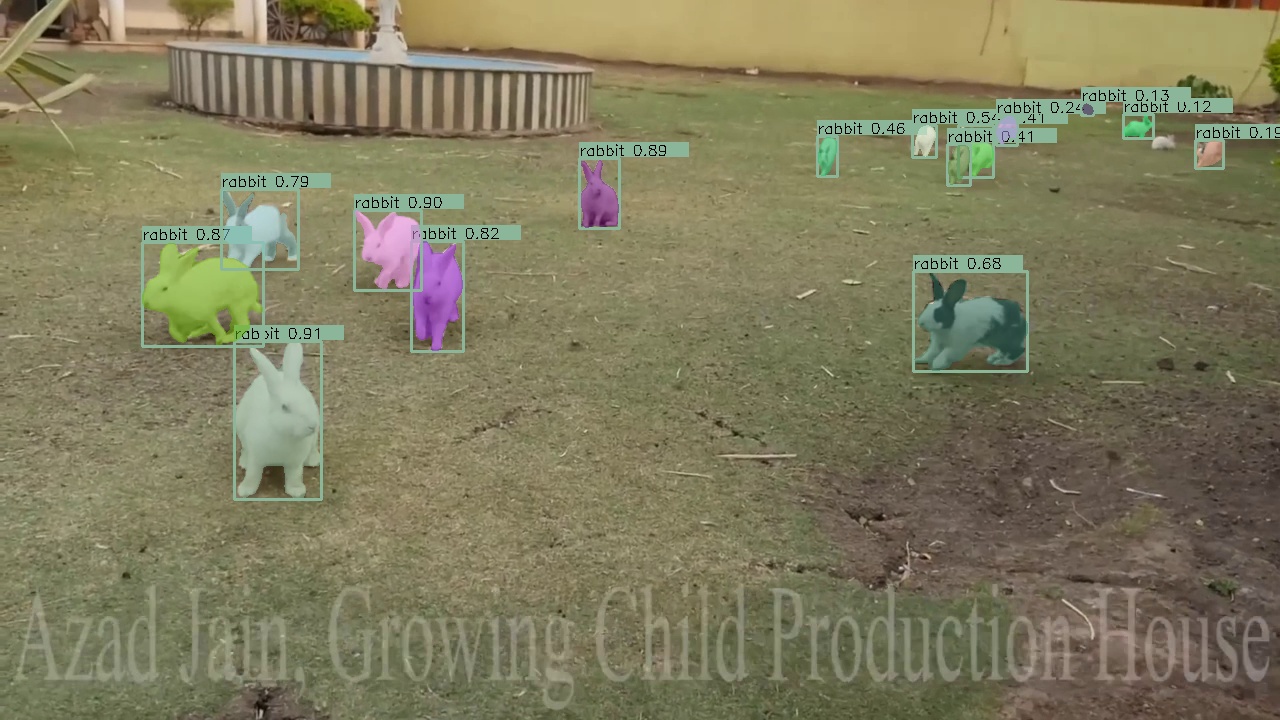} &
        \includegraphics[width=0.2\linewidth]{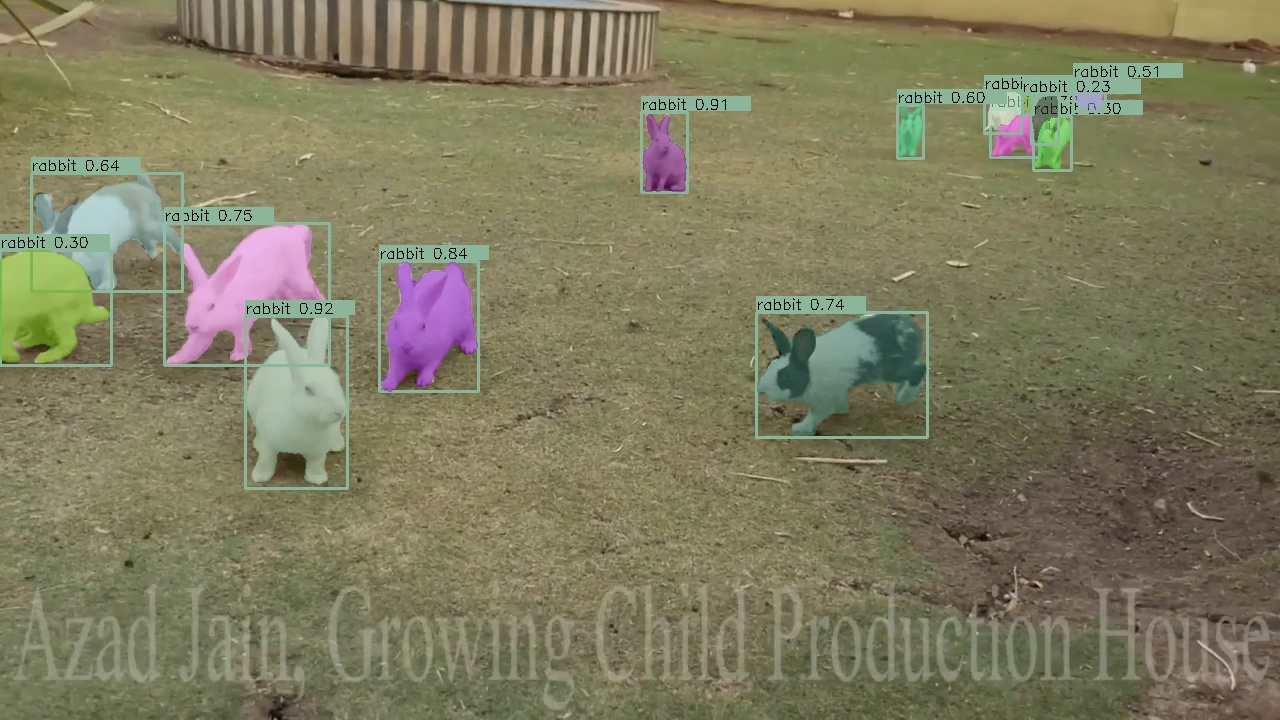} &
        \includegraphics[width=0.2\linewidth]{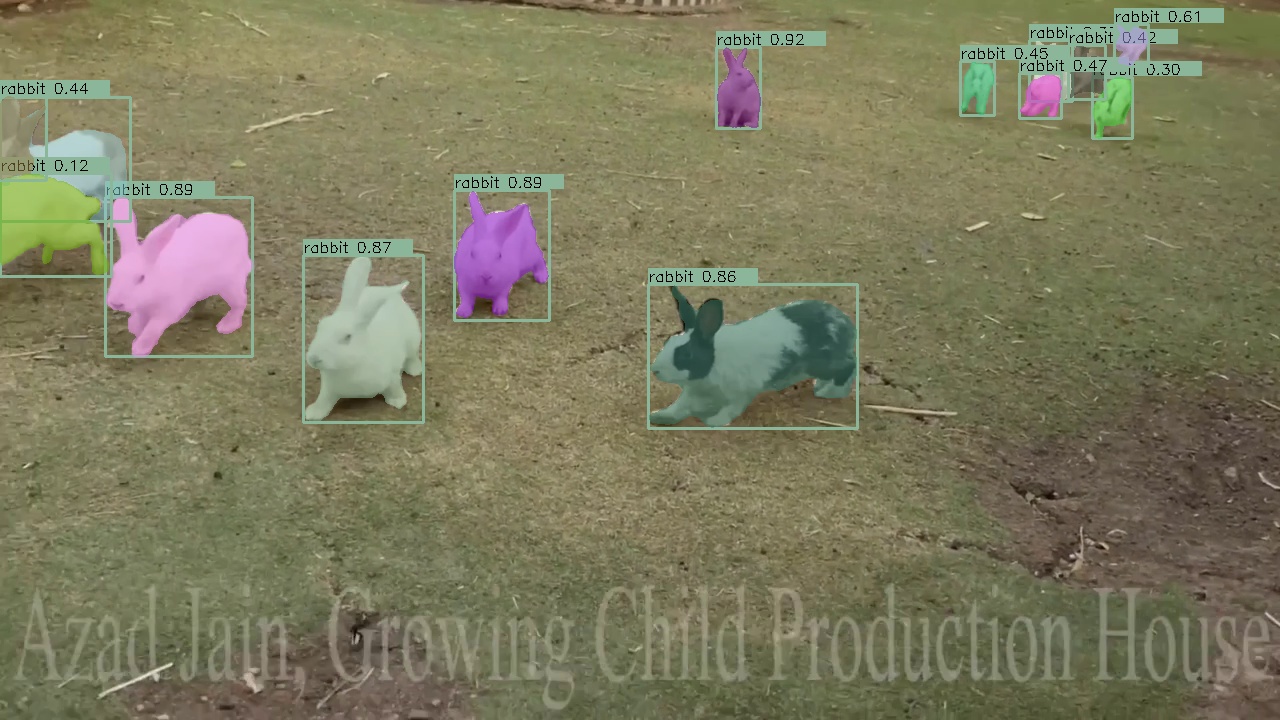} \\
        
        \includegraphics[width=0.2\linewidth]{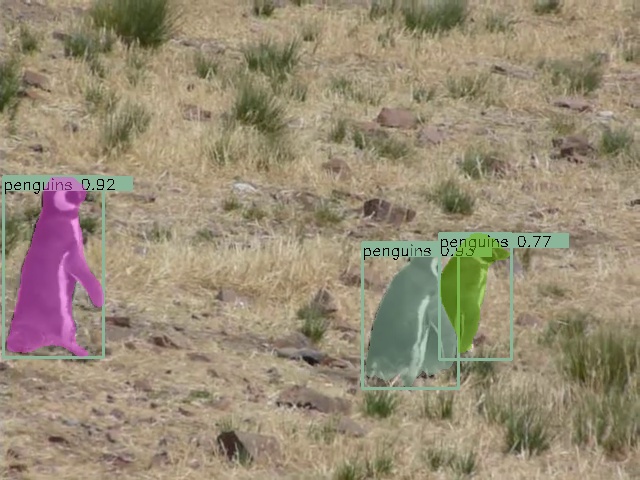} &
        \includegraphics[width=0.2\linewidth]{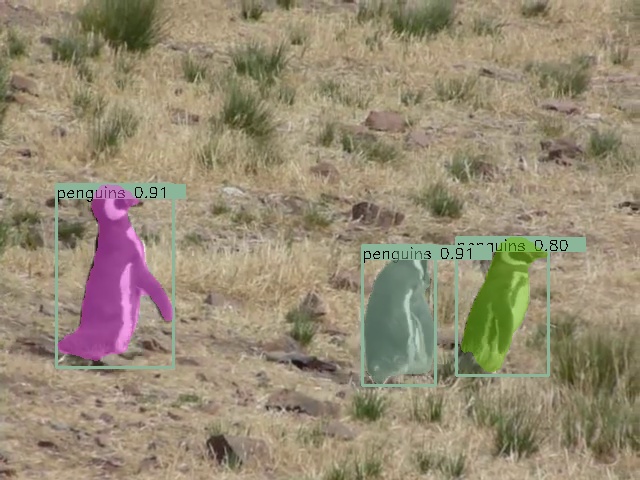} &
        \includegraphics[width=0.2\linewidth]{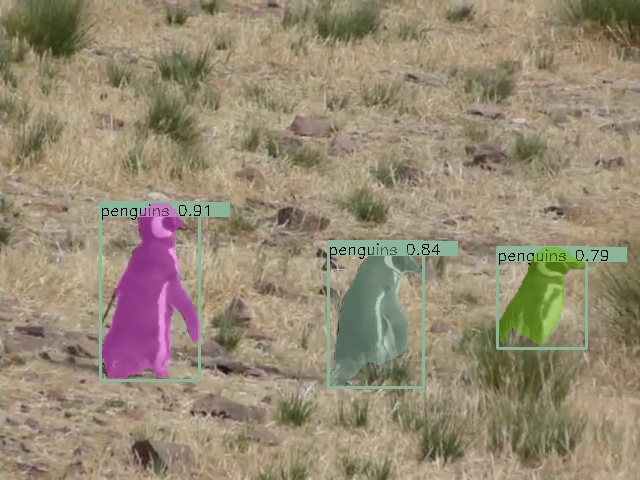} &
        \includegraphics[width=0.2\linewidth]{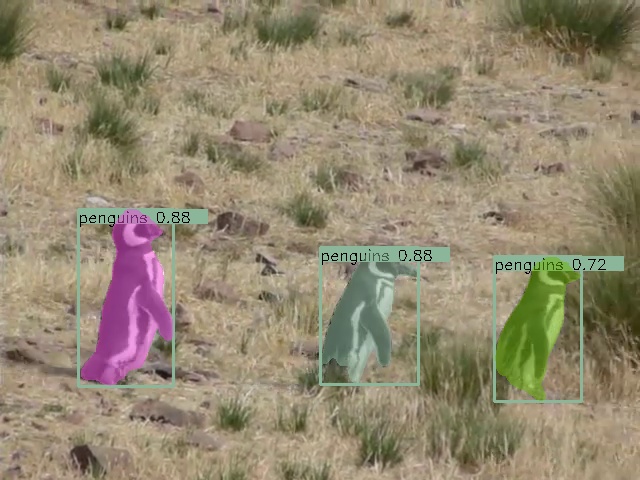} &
        \includegraphics[width=0.2\linewidth]{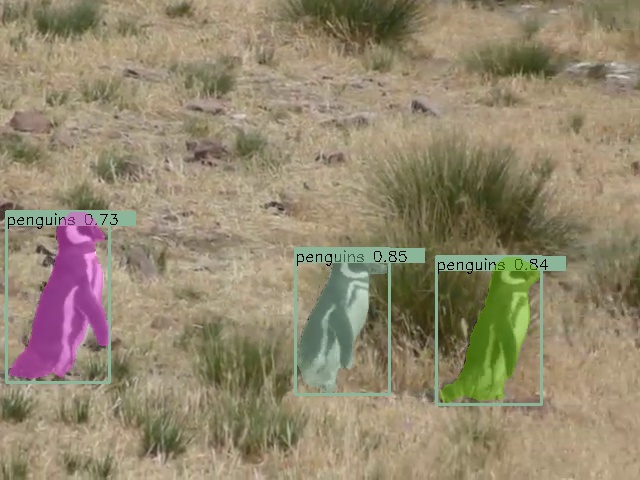} \\

        \includegraphics[width=0.2\linewidth]{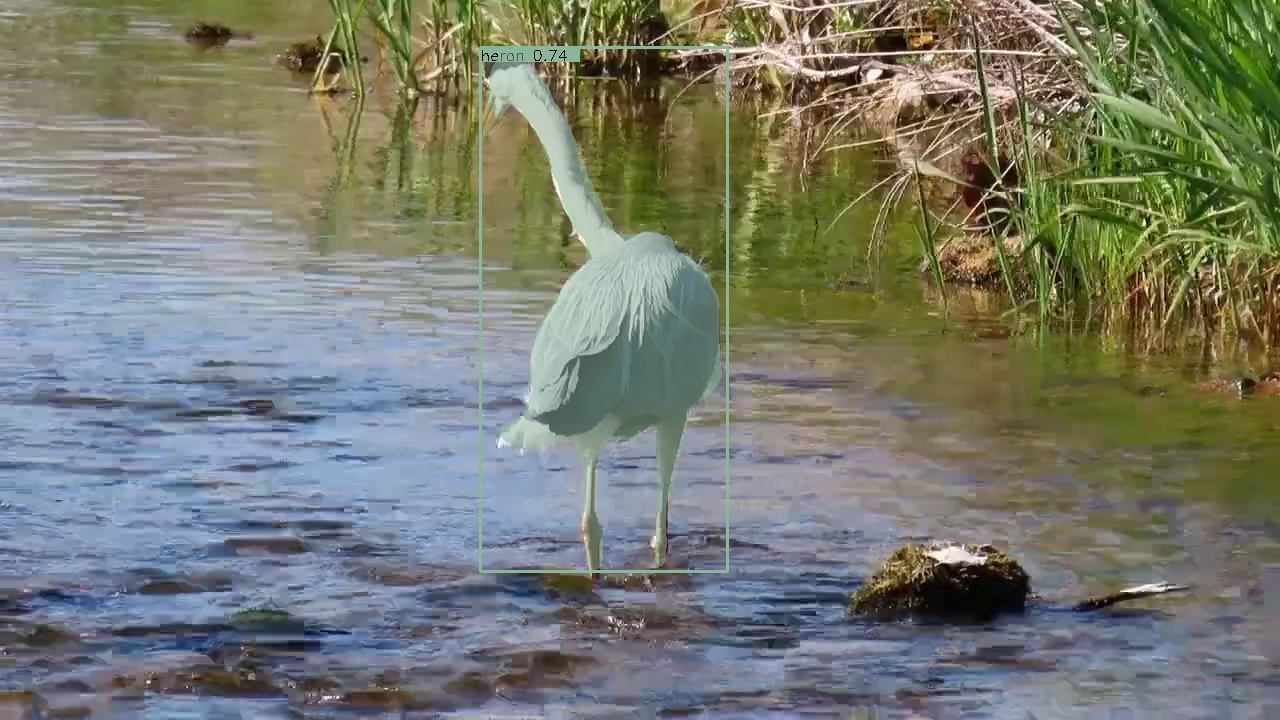} &
        \includegraphics[width=0.2\linewidth]{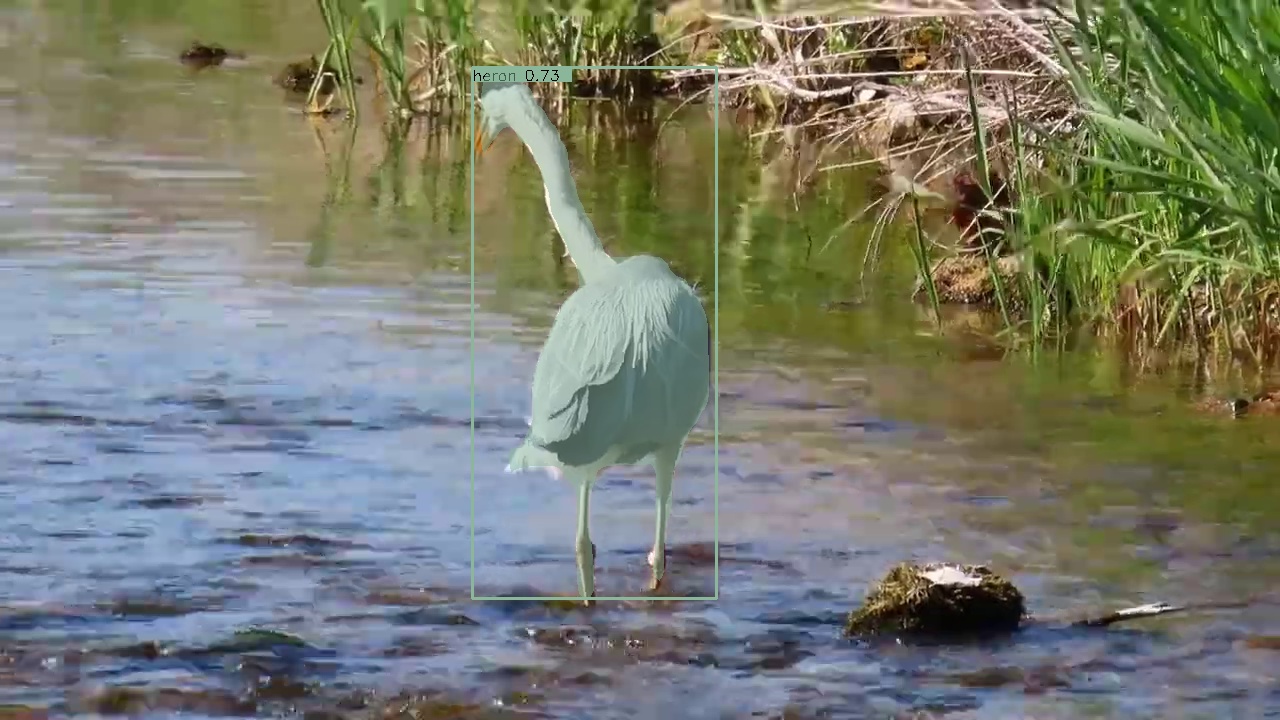} &
        \includegraphics[width=0.2\linewidth]{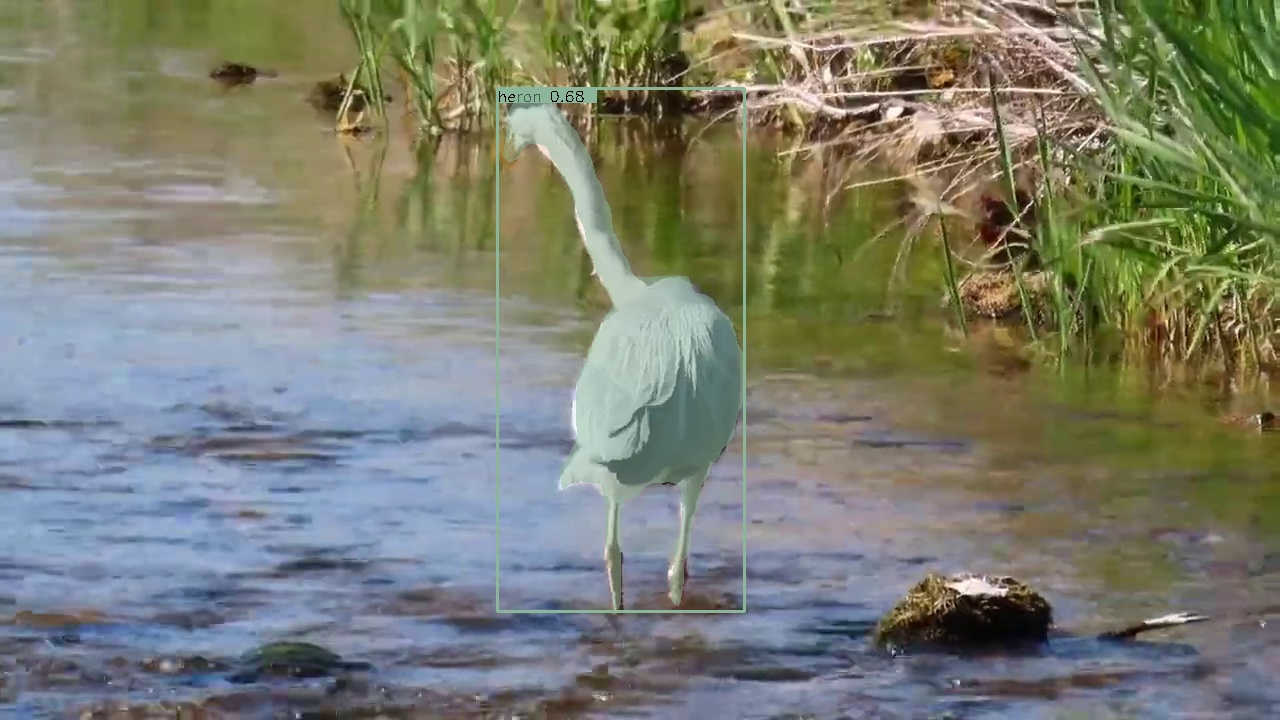} &
        \includegraphics[width=0.2\linewidth]{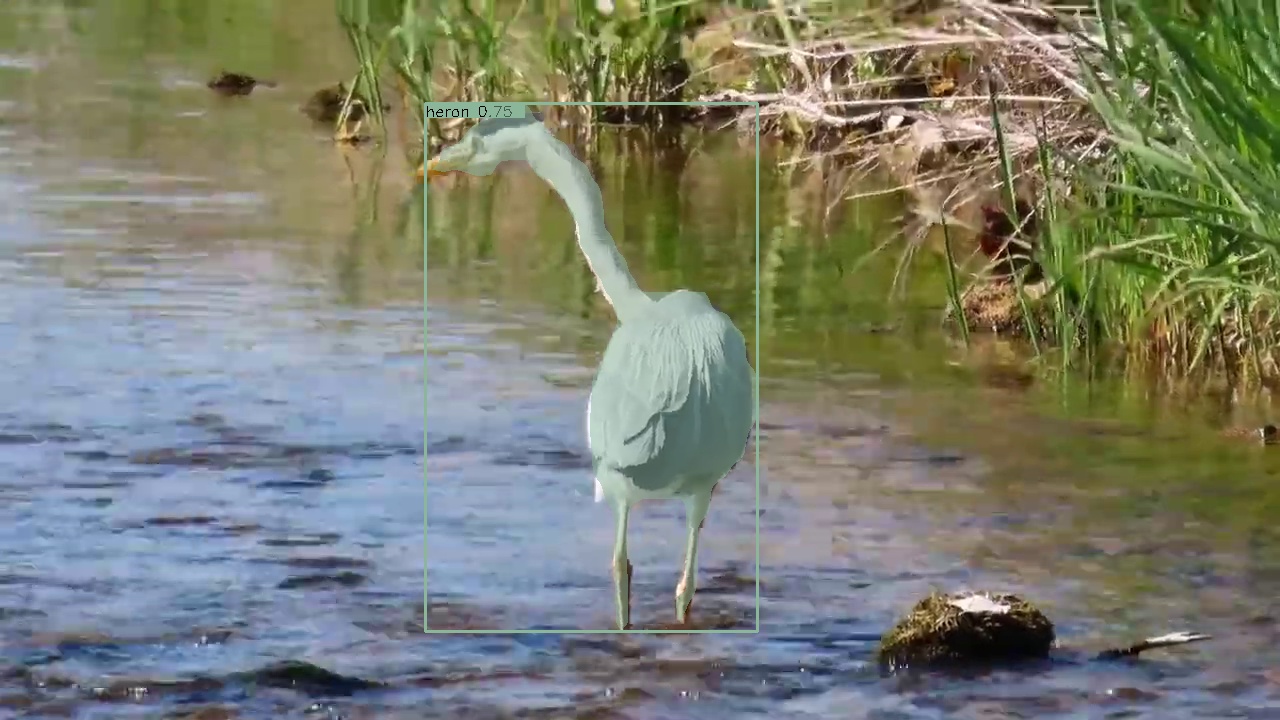} &
        \includegraphics[width=0.2\linewidth]{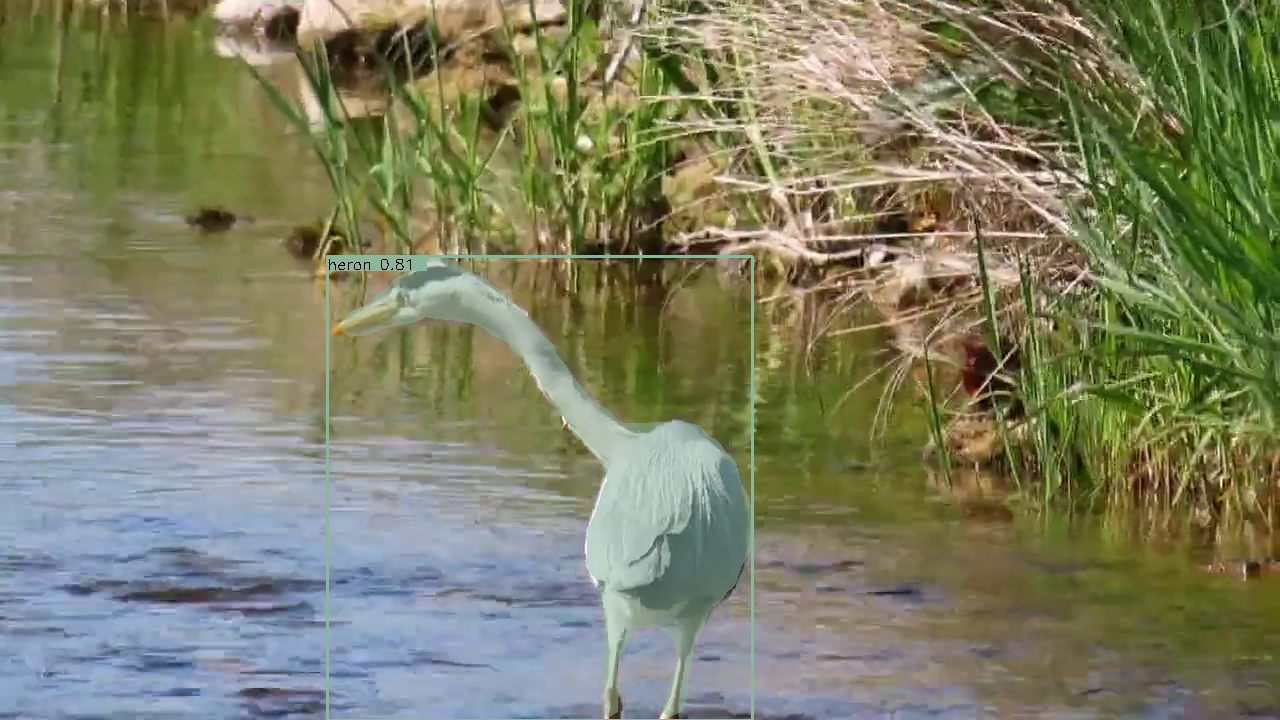} \\
        
        \includegraphics[width=0.2\linewidth]{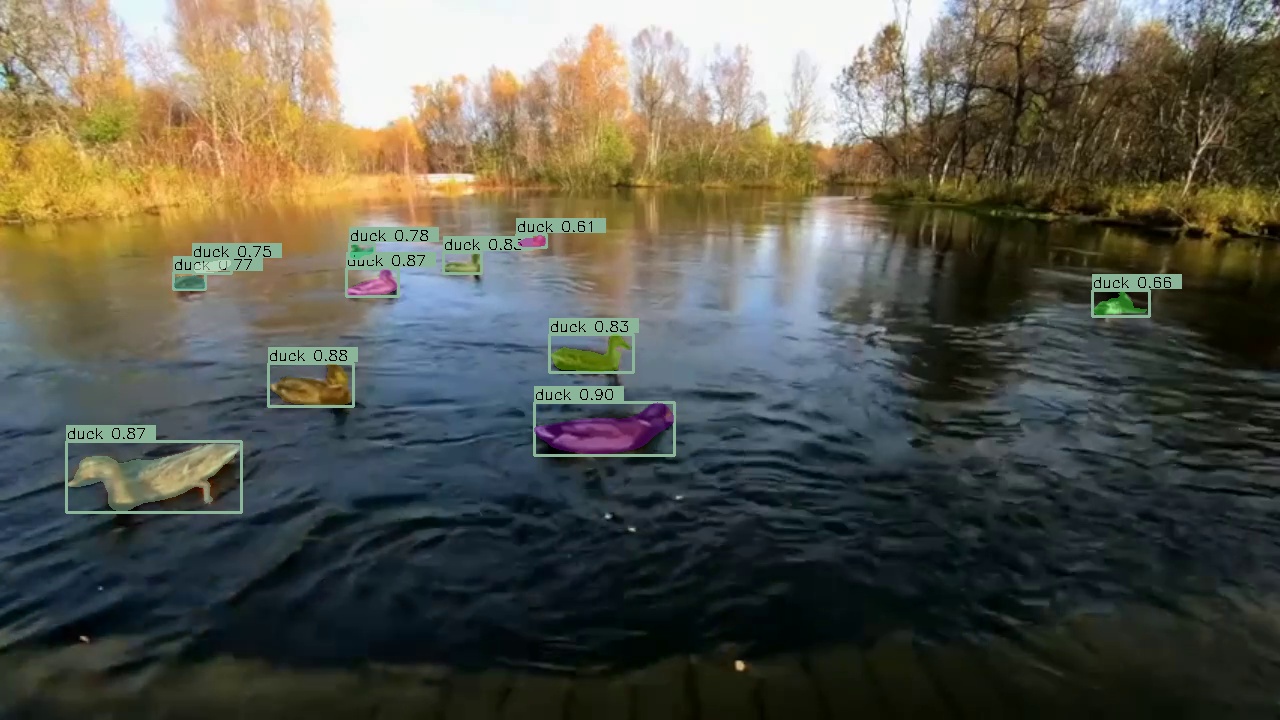} &
        \includegraphics[width=0.2\linewidth]{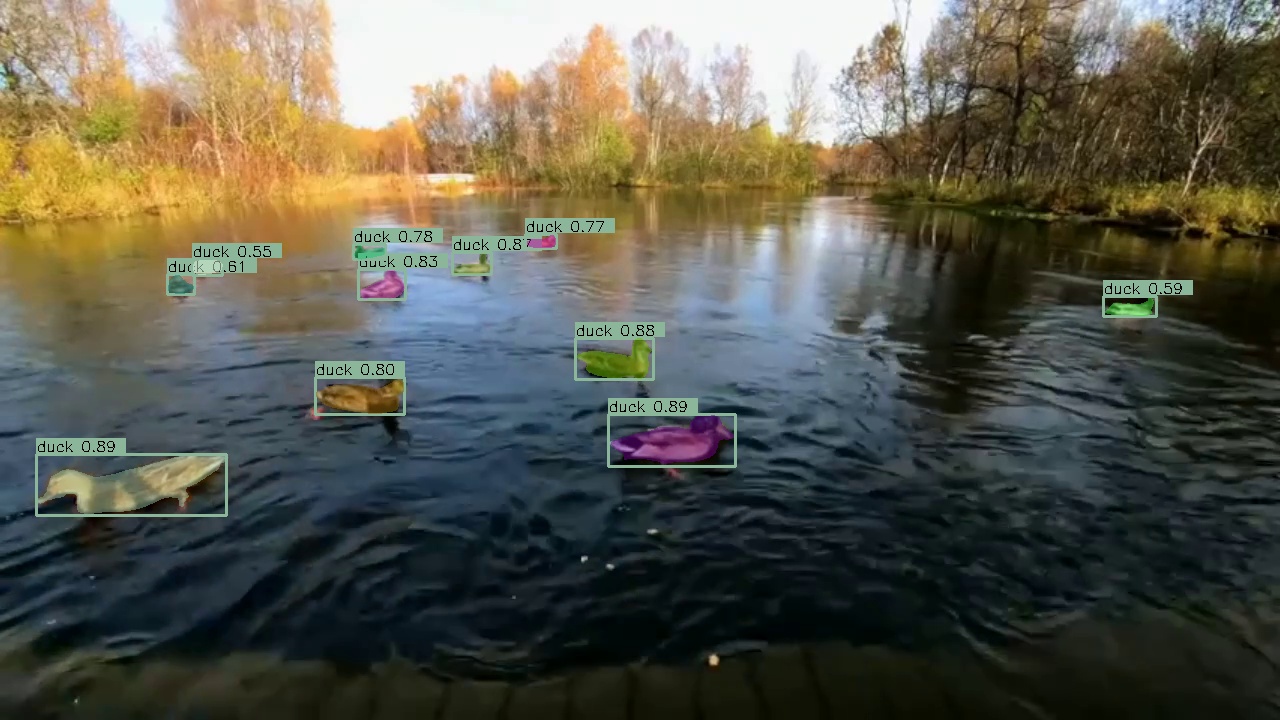} &
        \includegraphics[width=0.2\linewidth]{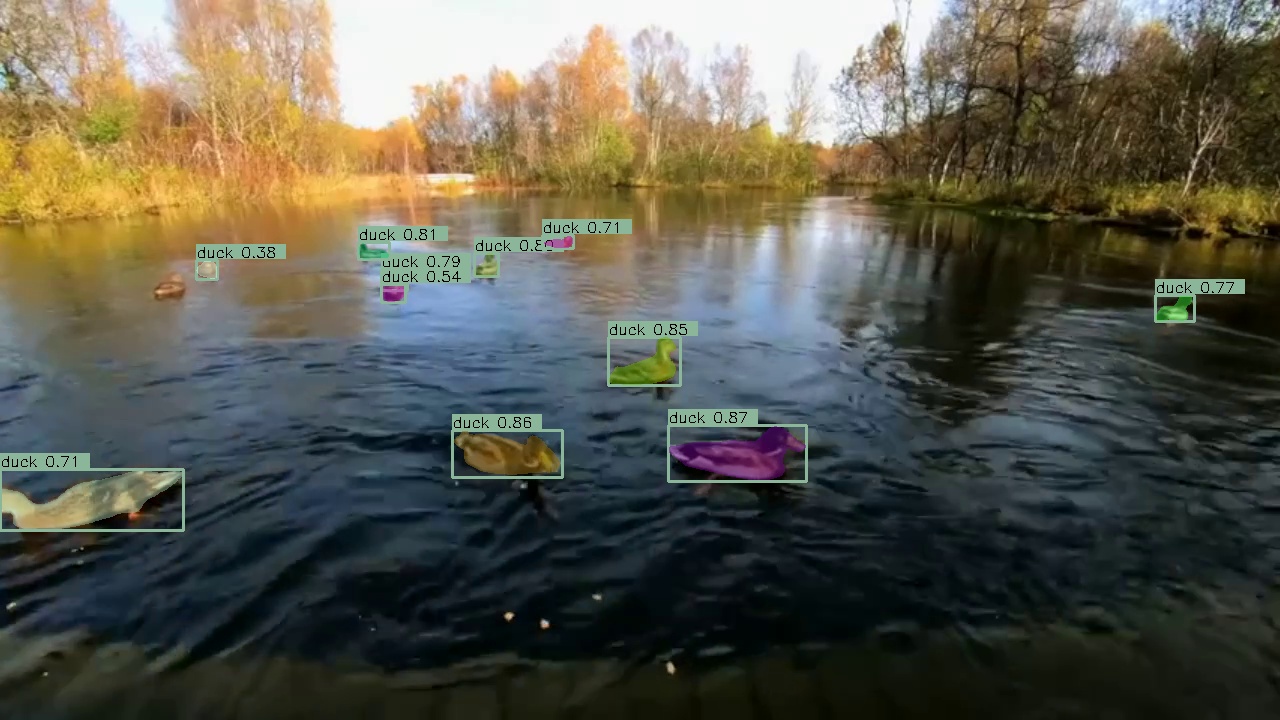} &
        \includegraphics[width=0.2\linewidth]{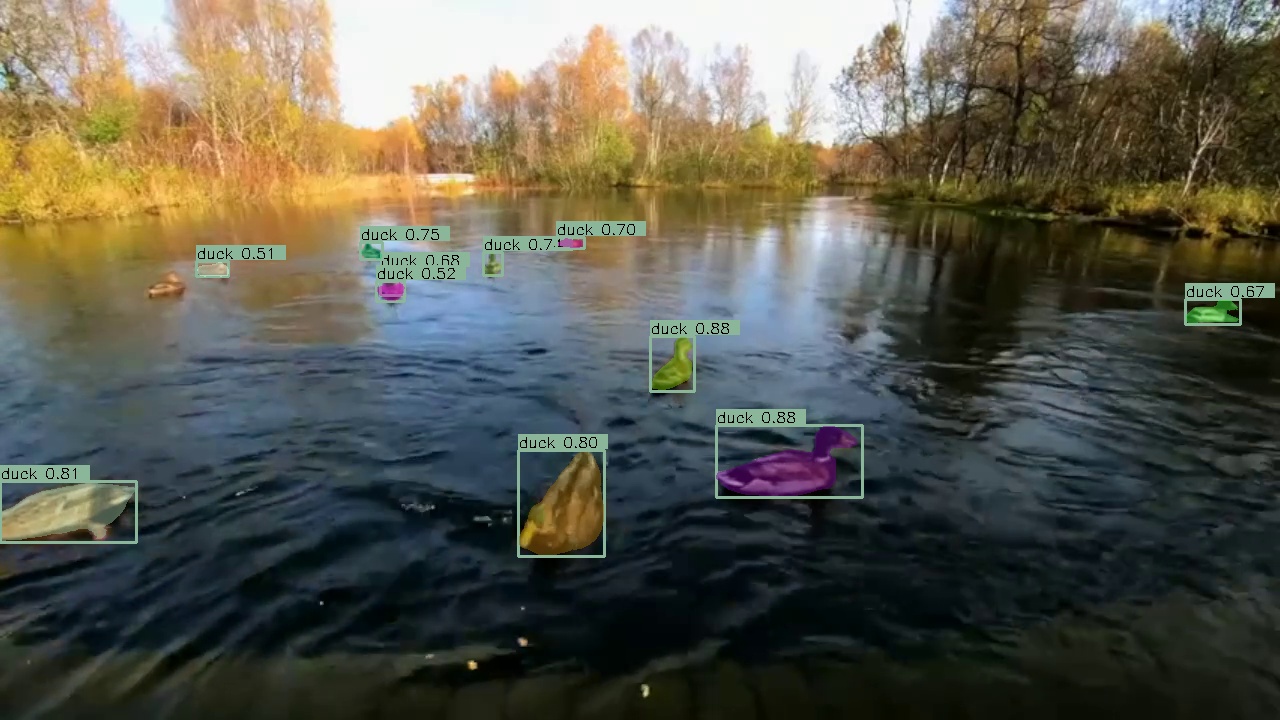} &
        \includegraphics[width=0.2\linewidth]{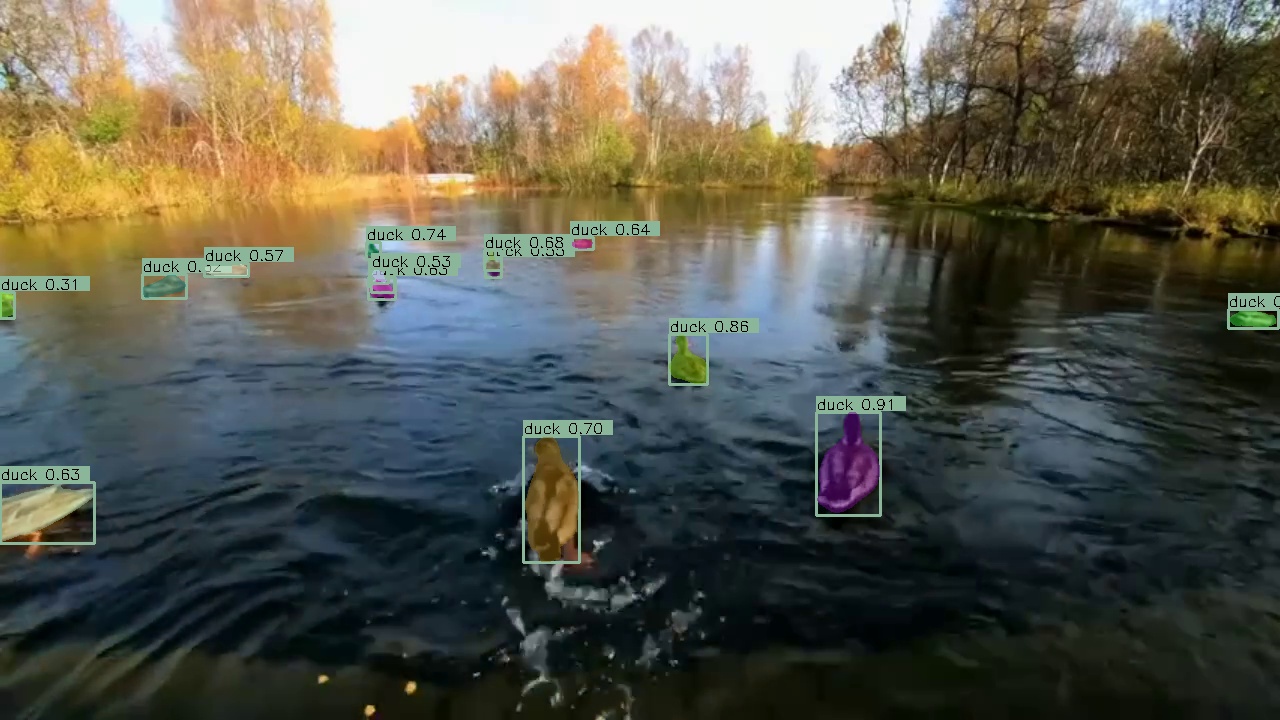} \\

        \includegraphics[width=0.2\linewidth]{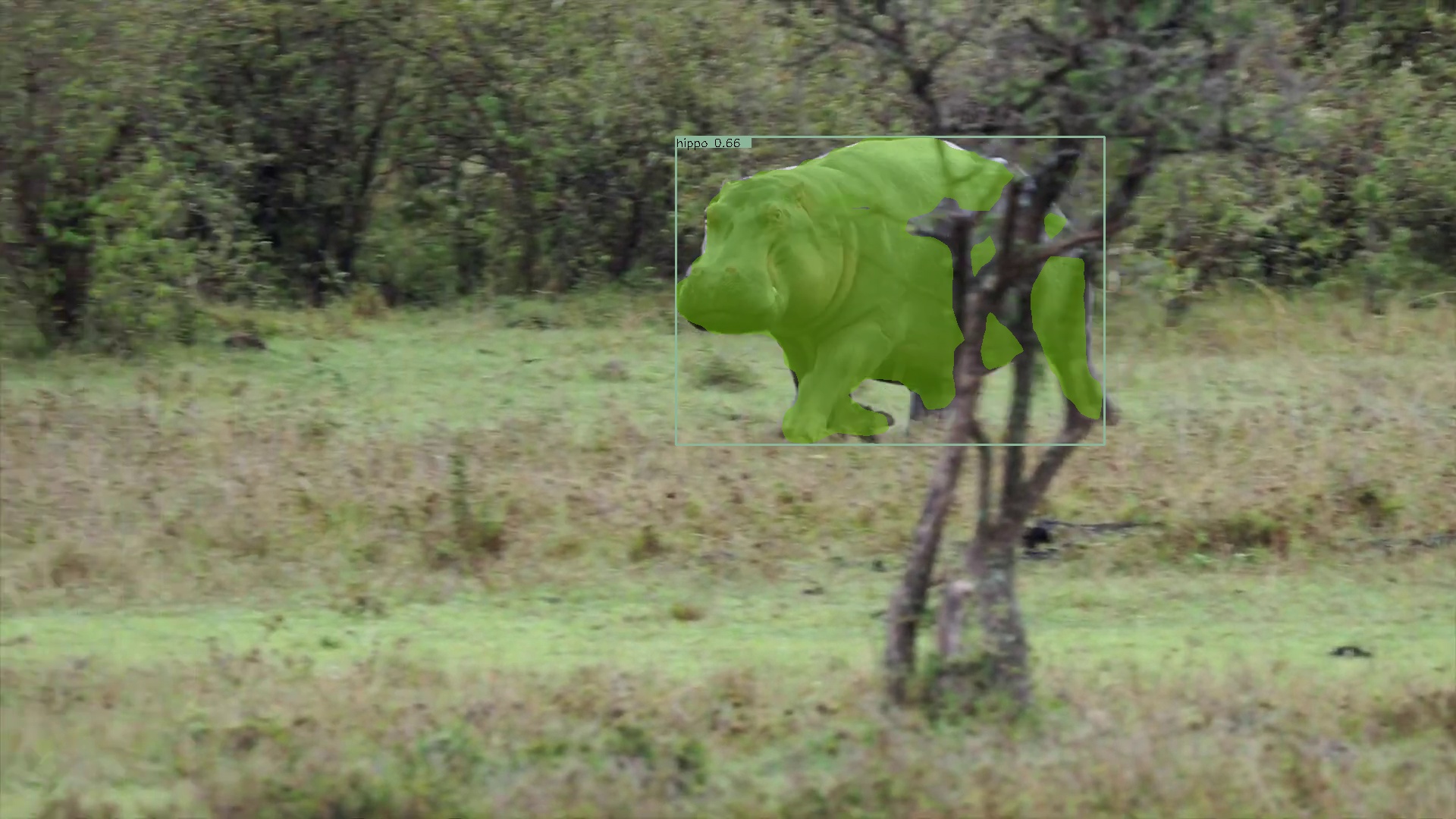} &
        \includegraphics[width=0.2\linewidth]{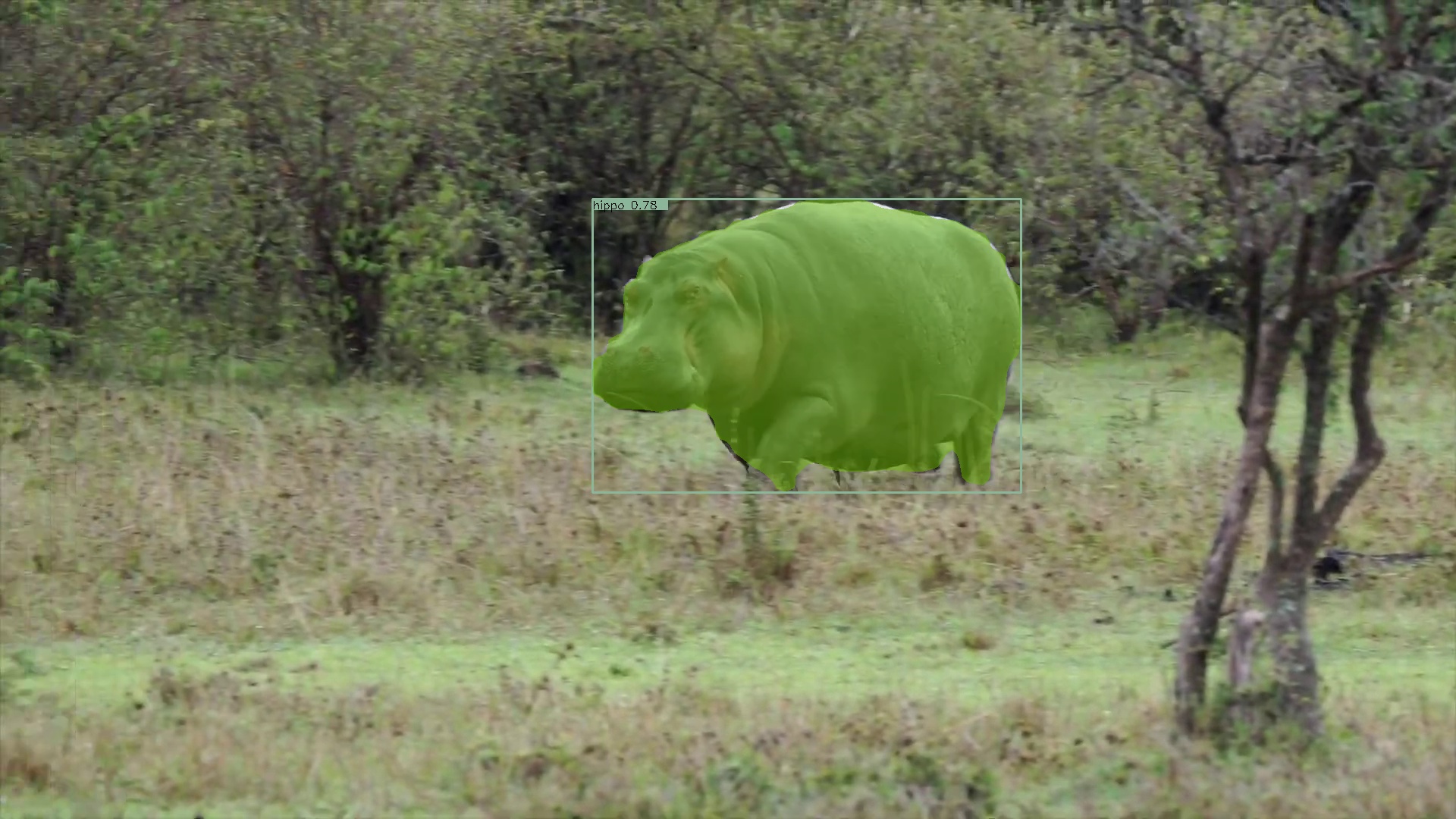} &
        \includegraphics[width=0.2\linewidth]{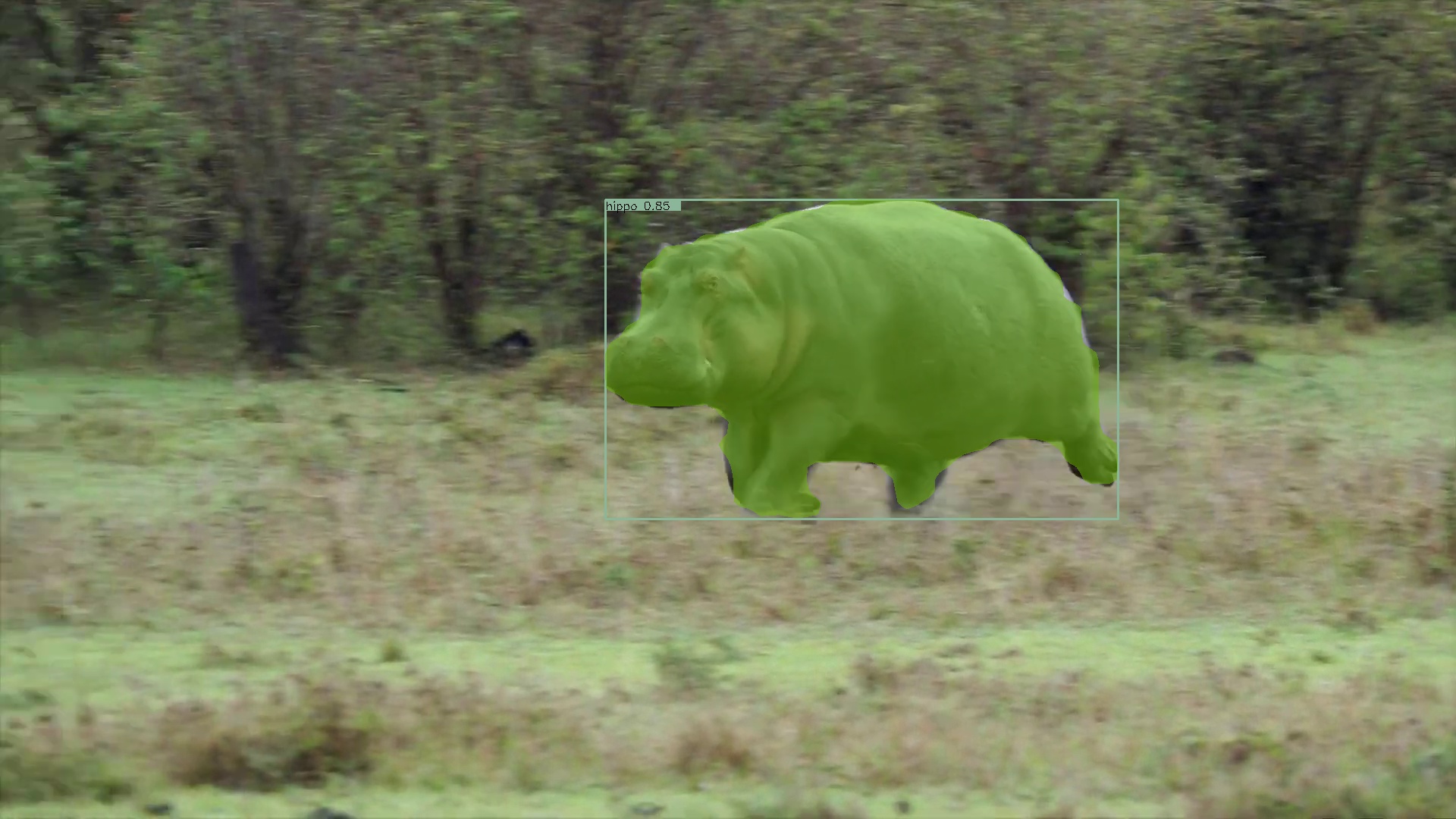} &
        \includegraphics[width=0.2\linewidth]{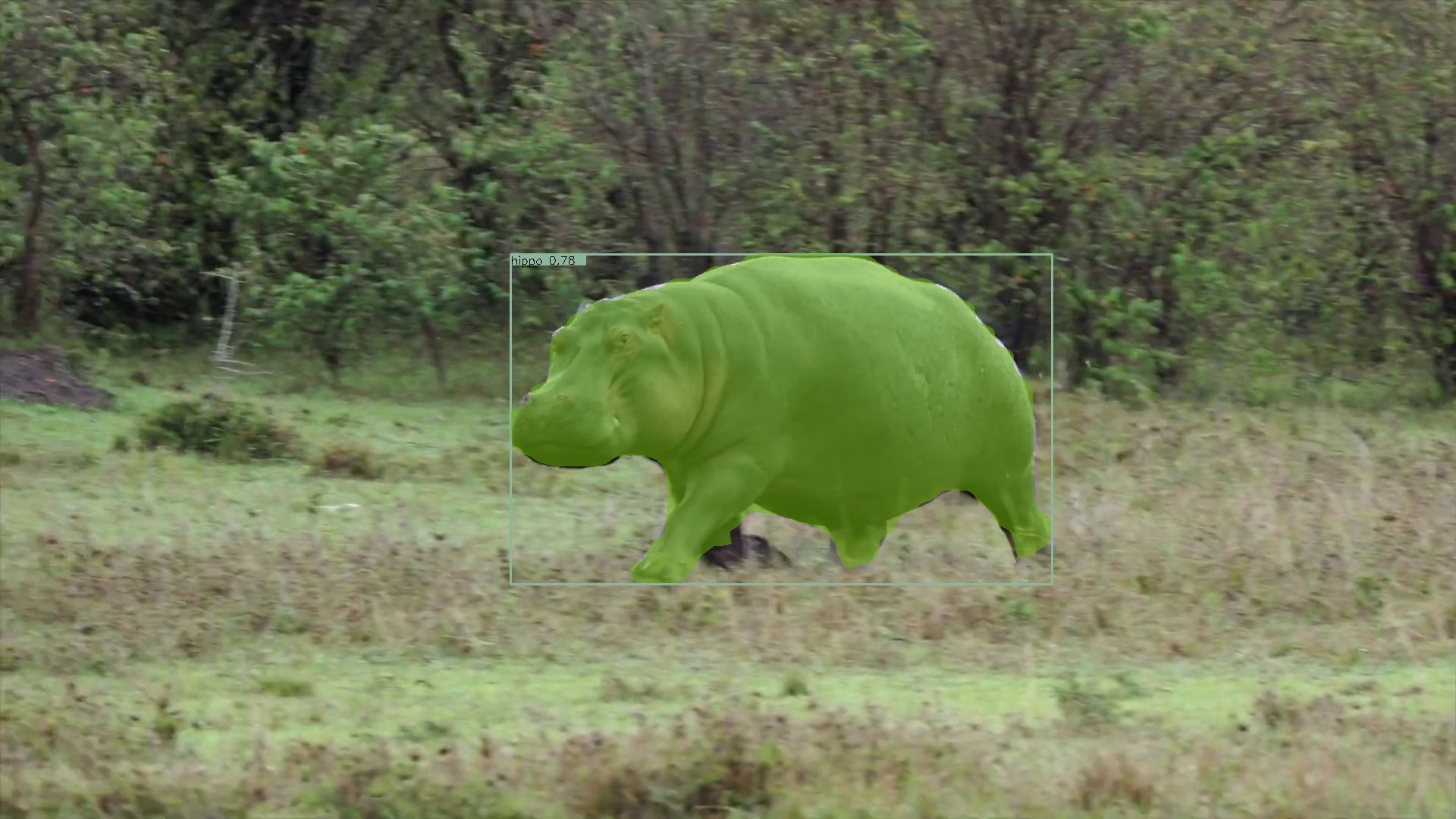} &
        \includegraphics[width=0.2\linewidth]{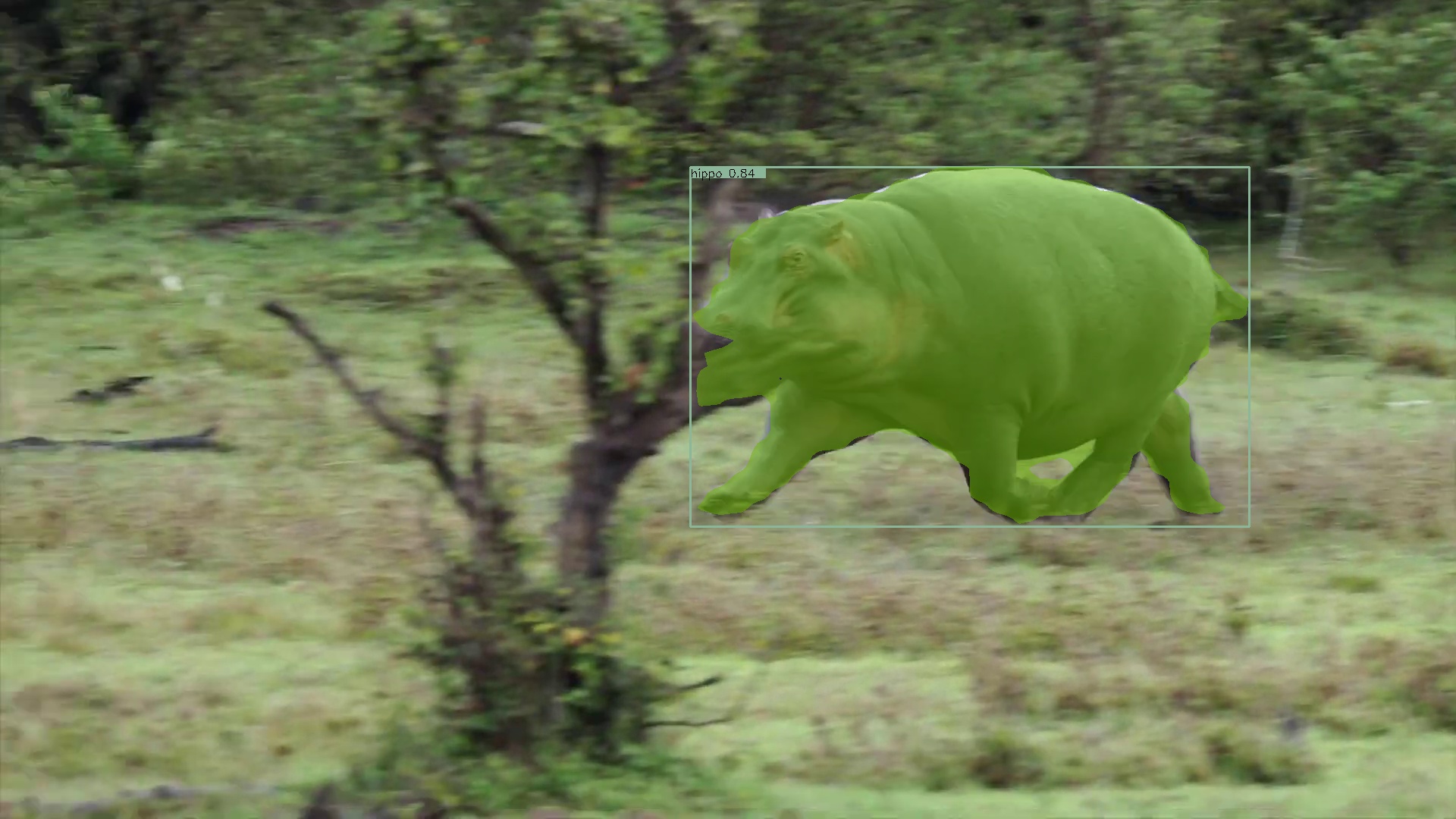} \\

        \bottomrule
    \end{tabular}}
          \caption{\textbf{Open-Vocabulary Tracking.} We condition our Grounding-DINO tracker on text prompts unseen during training and successfully track the corresponding objects in the videos. We use SAM to generate the mask from given the detected boxes. The mask color depicts the object's identity. We choose random internet videos to test our algorithm on diverse real-world scenarios. Best viewed digitally.
    }
    \label{fig:qualitative}
\end{figure*}

We show qualitative results of open-vocabulary tracking in Figure~\ref{fig:qualitative}. 
 We observe that our method does well on tracking, and is able to generalize even to very exotic classes, such as minions. More results can be found in the provided video. 

\subsection{Joint Segment and Track Everything}
\begin{figure*}[!t]
  \centering%
\includegraphics[width=1\linewidth]{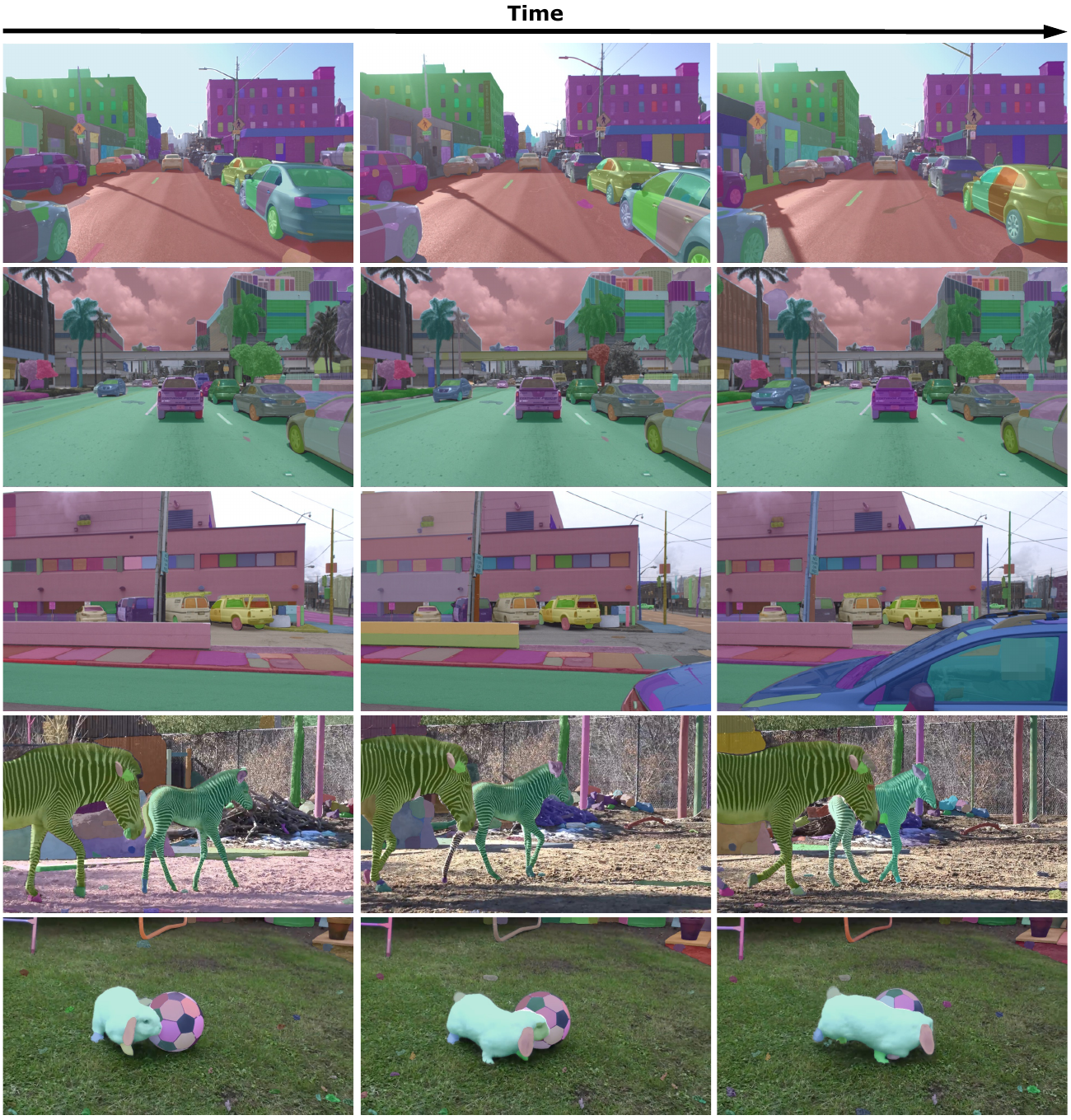}%
  \caption{Qualitative results of unified proposal generation and association. The same colour indicates the same instance. We notice that although we can learn strong associations using MASA, it is still very difficult to generate consistent proposals across frames. For example, we can see the missing segmentation of the building on the left in the second row. This indicates further research efforts are needed on consistent proposal generation in videos.}%
  \label{fig-qualitative-tracking}%
  
\end{figure*}

We provide qualitative results on our joint segmentation and tracking models. Since we learn proposal generation and association in a joint way, it makes our model capable of segmenting and associating anything in videos. ~\figureautorefname~\ref{fig-qualitative-tracking} shows the qualitative association performance using our self-generated proposals. We notice that although we can learn strong associations using MASA, it is very difficult to generate consistent proposals across frames. For example, we can see the missing segmentation for the building on the left in the second row. Those inconsistent detections will lead to severe flickering effects when visualising the results on videos. This indicates we still need further efforts on consistent proposal generation for robust detecting objects in videos.

\section{Implementation Details}
\label{sec:supp-implementation_detail}
We provide more details regarding our model architecture, training, and inference.

\subsection{Architecture Detail}
\noindent\textbf{MASA Adapter}
The MASA Adapter comprises two main parts. The first part involves the construction of a feature pyramid and dynamic feature fusion.  The second part is the FasterRCNN-based detection head for the object prior to distillation and the track head for producing tracking features. 
The construction process for the feature pyramid varies depending on the backbone used. These variations are detailed in the respective sections for each model. The dynamic feature fusion employs standard deformable convolution, as outlined in \cite{deformable_conv}, to aggregate information across spatial locations and feature levels. Additionally, task-aware attention and scale-aware attention from \cite{DyHead} are utilized for SAM-based models. In total, three fusion blocks are established for the feature fusion process.

The FasterRCNN-based detection head includes a region proposal network and a class-agnostic box regression head. The track head comprises four convolutional layers and one fully connected layer, used to generate instance embeddings.

\noindent\textbf{Ours-Detic}
We utilize the pre-trained Detic~\cite{Detic} model with Swin-B~\cite{swin} as the backbone. The pre-trained model adheres to the open-vocabulary object detection setup described in~\cite{VILD}, where rare classes from LVIS are excluded from training. We freeze the Detic Swin-B backbone and employ the standard FPN for constructing the feature pyramid. Specifically, we extract features from the $4^{th}$, $22^{nd}$, and $24^{th}$ blocks of the Swin-B backbone. Subsequently, we integrate the dynamic feature fusion atop the feature pyramid to learn tracking features through detection distillation and instance contrastive learning.

\noindent\textbf{Ours-Grounding-DINO}
We employ the pre-trained Grounding-DINO~\cite{GroundingDINO} model with Swin-B~\cite{swin} as the backbone. The Swin-B backbone is frozen, and we use the standard FPN to construct the feature pyramid. Apart from the differing pre-training and window sizes for the Swin backbone, all learnable components are identical to Ours-Detic.

\noindent\textbf{Ours-SAM-B}
This model is based on SAM, with all original SAM components frozen. To obtain multi-level hierarchical features from the plain ViT backbone of SAM, we extract feature maps from the outputs of the $3^{rd}$, $6^{th}$, $9^{th}$, and $12^{th}$ blocks. Transposed Convolutions are used to upscale the feature map from the $3^{rd}$ block by $4\times$ and from the $6^{th}$ block by $2\times$. We maintain the $9^{th}$ feature map as is, and downscale the feature map from the $12^{th}$ block by $1/2$ using MaxPooling. This approach yields hierarchical features with scale ratios of ${\frac{1}{4}, \frac{1}{8}, \frac{1}{16}, \frac{1}{32}}$. The remainder of the model mirrors the two models mentioned above.

\noindent\textbf{Ours-SAM-H}
The learnable portion is largely similar to Ours-SAM-B. The sole distinction is that we extract features from the outputs of the $8^{th}$, $16^{th}$, $24^{th}$, and $32^{nd}$ blocks to construct the feature pyramid.

\subsection{More Training Details}
For SAM-based models, we turn off MixUp augmentation in the last two epochs. After that, we finetune the track heads of the SAM-based models while freezing the other parts with all augmentations for 6 epochs. 

For training our model with any raw image collection, the following pipeline is utilized. Initially, the 'everything' mode of SAM is employed to generate training data on raw images offline, using the SAM-ViTH model to ensure higher quality. We adhere to the default SAM settings, which involve using 32 sampling points along each side of an image. Additionally, an Intersection over Union (IoU) prediction threshold of 0.88 is applied to filter out low-quality predictions. Subsequently, small disconnected regions and holes in masks are removed. Bounding box Non-Maximum Suppression (NMS) is also used to eliminate overlapping predictions with a threshold of 0.7. In our ablation studies, this pipeline is applied to generate data on raw COCO and BDD100K images.

\subsection{Inference with Given Observations}
 Notably, during testing on UVO, in addition to using proposals generated by our fast-segmenting everything mode, we also incorporate the same per-frame mask observation as employed in ~\cite{deva}. This inclusion aims to minimize the temporal inconsistency in SAM's mask predictions on videos.

\begin{algorithm}[t]
    \caption{Inference pipeline of MASA for associating objects across a video sequence.}
    \label{alg:inference}
    \begin{algorithmic}[1]
        \INPUT frame index $t$, object candidates $r \in P$, confidence $p_r$, detection embeddings $\textbf{q}_r$, and track embeddings $\textbf{q}_\tau$ for all $\tau \in \mathcal{T}$.
        \State \texttt{DuplicateRemoval}($P$)
        \For{$r \in P, \tau \in \mathcal{T}$} \LineComment{compute matching scores}
        \State \textbf{f}$(r, \tau) = $ \texttt{similarity}($\mathbf{q}_r, \mathbf{q}_\tau$)
        \EndFor
        \For{$r \in P$} \LineComment{track management}
        \State $c$ = \texttt{max}$\left(\textbf{f}(r)\right)$ \LineComment{match confidence}
        \State $\tau_{\texttt{match}}$ = \texttt{argmax}$\left(\textbf{f}(r)\right)$ \LineComment{matched track ID}
        \If{$c > \beta$ \textbf{and} $p_i >$ $\beta_{\texttt{obj}}$} \LineComment{object match found}
        \State \texttt{updateTrack}$\left(\tau_{\texttt{match}}, r, \textbf{q}_r, t\right)$  \LineComment{update track}
        \ElsIf{$p_r > \gamma$}
        \State \texttt{createTrack}$\left(r, \textbf{q}_r, t\right)$ \LineComment{create new track}
        \EndIf
        \EndFor
    \end{algorithmic}
\end{algorithm}

\subsection{Inference Details}
\label{sec:supp-infer-details}
Overall, our inference scheme is illustrated in Algorithm~\ref{alg:inference}.
In terms of similarity computation, we provide the formula that we use:

\begin{equation}
    \begin{gathered}
        \textbf{$s_1$}(\tau, r) = \frac{1}{2} \left[\frac{ \text{exp}(\textbf{q}_{r} \cdot \textbf{q}_{\tau})}{\sum_{r' \in P} \text{exp}(\textbf{q}_{r'} \cdot \textbf{q}_{\tau} )} + \frac{\text{exp}(\textbf{q}_{r} \cdot \textbf{q}_{\tau})}{ \sum_{\tau' \in \mathcal{T}} \text{exp}(\textbf{q}_{r} \cdot \textbf{q}_{\tau'} )}\right] \\
        \textbf{$s_2$}(\tau, r)  = \frac{\textbf{q}_r \cdot \textbf{q}_\tau}{\|\textbf{q}_r\| \|\textbf{q}_\tau\|} \\
        s(\tau, r) = \frac{1}{2} (s_1(\tau, r) + s_2(\tau, r))
    \end{gathered}
\end{equation}
where \(\textbf{s}(\tau, r)\) represents the similarity score between a track \(\tau\) and an object candidate \(r\). Here, \(\textbf{q}_r\) denotes the detection embedding of the object candidate \(r\), encapsulating its appearance features, while \(\textbf{q}_\tau\) represents the track embedding for track \(\tau\), capturing the features of the tracked object. The $\textbf{$s_1$}(\tau, r)$ employs an exponential function to compute the dot product of these embeddings, reflecting the degree of similarity between the object candidate and the track. This similarity score is normalized twice: firstly, across all object candidates \(r'\) in the set \(P\) for a given track \(\tau\), and secondly, across all tracks \(\tau'\) in the set \(\mathcal{T}\) for a given object candidate \(r\). This dual normalization ensures a balanced and comprehensive assessment of similarity, facilitating accurate object association in dynamic video sequences. $\textbf{$s_2$}(\tau, r)$ computes the cosine similarity. The final $s(\tau, r)$ score is the average between $\textbf{$s_1$}(\tau, r)$ and $\textbf{$s_2$}(\tau, r)$.

\section{Limitations}
\label{sec:supp-limit}
One key limitation of our approach is handling temporal inconsistencies in detection or segmentation results across video frames. This issue, common in open-world object detection and segmentation models like SAM, is evident when an object detected in one frame is missed in the next, causing flickering effects in video visualization, as seen in our demonstrations. While our MASA adapter excels in learning associations, it cannot rectify foundational models' detection or segmentation errors. The challenge of generating consistent proposals across frames highlights an important area for future research to enhance the robustness and stability of object detection in dynamic video environments.

Another limitation is the lack of a long-term memory system, which is crucial for handling occlusions. We only learn a universal instance appearance model, which can be used directly by different detectors. However, the tracking strategy and memory management are still done by the bi-softmax matching and a queue to store each instance's appearance embeddings. This simple strategy is prone to failure in cases of severe occlusions. Developing a more sophisticated long-term memory system and improved tracking strategies will be essential to address this limitation.

\begin{figure*}[!t]
  \centering%
\includegraphics[width=0.8\linewidth]{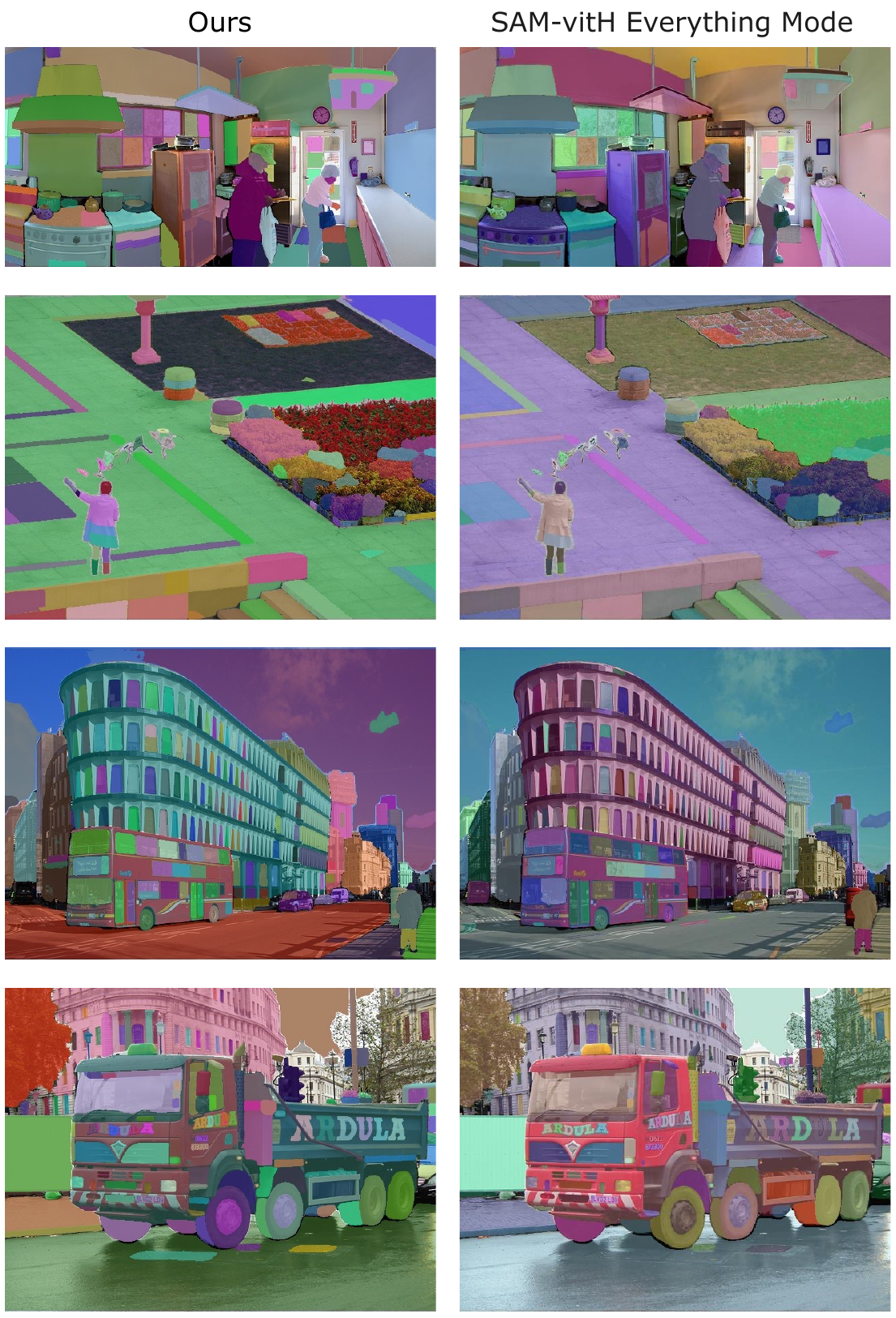}%
  \caption{Qualitative comparison between our fast proposal generation and the original SAM everything mode on images from COCO validation set. The results show that our fast proposal generation can achieve similar segmentation quality to the everything mode of SAM, despite using much less time.}%
  \label{fig-qualitative}%
\end{figure*}

% \clearpage

%%%%%%%%%%%%%%%%%%%%%%%%%%%%%%%%%%%%%%%%%%%%%%%%%%%%%%%%%%%%

\clearpage

\end{document}